\documentclass[conference]{IEEEtran}

\usepackage{comment}
\usepackage{wrapfig}
\usepackage{floatflt}

\newif\ifdraft
\draftfalse

\usepackage{graphicx} % Required for inserting images
\usepackage{xcolor}
\usepackage{{booktabs}} % for tables
\usepackage{tabularx}
\usepackage{amsmath}
\usepackage{amsfonts}

\usepackage{multirow} % multirow in table
\usepackage{colortbl} % color gradient in table

\usepackage{subcaption}

\definecolor{darkblue}{rgb}{0.1,0.1,0.9}
\usepackage[pagebackref,breaklinks,colorlinks,allcolors=darkblue]{hyperref}

\usepackage[capitalize]{cleveref}
\crefname{section}{Sec.}{Secs.}
\Crefname{section}{Section}{Sections}
\Crefname{table}{Table}{Tables}
\crefname{table}{Tab.}{Tabs.}

\title{Shape Bias and Robustness Evaluation via Cue Decomposition for Image Classification and Segmentation}

\author{
\IEEEauthorblockN{1\textsuperscript{st} Edgar Heinert\textsuperscript{\textsection}}
 \IEEEauthorblockA{\textit{Department of  Mathematics} \\
 \textit{University of Wuppertal}\\
 Wuppertal, Germany \\
 heinert@uni-wuppertal.de}
 \and
 \IEEEauthorblockN{1\textsuperscript{st} Thomas Gottwald\textsuperscript{\textsection}}
 \IEEEauthorblockA{\textit{Department of  Mathematics} \\
 \textit{University of Wuppertal}\\
 Wuppertal, Germany \\
 gottwald@uni-wuppertal.de}
 \and
 \IEEEauthorblockN{1\textsuperscript{st} Annika Mütze\textsuperscript{\textsection}}
 \IEEEauthorblockA{\textit{Department of  Mathematics} \\
 \textit{University of Wuppertal}\\
 Wuppertal, Germany \\
 muetze@uni-wuppertal.de}
 \and
 \IEEEauthorblockN{4\textsuperscript{th} Matthias Rottmann}
 \IEEEauthorblockA{\textit{Department of  Mathematics} \\
 \textit{University of Wuppertal}\\
 Wuppertal, Germany \\
 rottmann@uni-wuppertal.de}
 }

\begin{document}
\maketitle
\begingroup\renewcommand\thefootnote{\textsection}
\footnotetext{Equal contribution}
\endgroup
\begin{abstract}
    Previous works studied how deep neural networks (DNNs) perceive image content in terms of their biases towards different image cues, such as texture and shape. Previous methods to measure shape and texture biases are typically style-transfer-based and limited to DNNs for image classification.
    In this work, we provide a new evaluation procedure consisting of 1) a cue-decomposition method that comprises two AI-free data pre-processing methods extracting shape and texture cues, respectively, and 2) a novel cue-decomposition shape bias evaluation metric that leverages the cue-decomposition data. For application purposes we introduce a corresponding cue-decomposition robustness metric that allows for the estimation of the robustness of a DNN w.r.t.\ image corruptions. In our numerical experiments, our findings for biases in image classification DNNs align with those of previous evaluation metrics. However, our cue-decomposition robustness metric shows superior results in terms of estimating the robustness of DNNs. Furthermore, our results for DNNs on the semantic segmentation datasets Cityscapes and ADE20k for the first time shed light into the biases of semantic segmentation DNNs.
\end{abstract}

%%%%%%%%% INTRO 
\section{Introduction}
Deep neural networks (DNNs) have demonstrated exceptional effectiveness in image perception, achieving human-level performance in tasks like image classification, semantic segmentation, and object detection \cite{hirsch2021radiologist, he2015delving, alzubaidi2021review, esteva2017dermatologist}. However, research reveals that the biases they exhibit can differ significantly from human biases \cite{zhang2024can,vakali2024rolling,geirhos2020shortcut, wang2022frequency,keser2024unveiling}, and they often exhibit greater vulnerability to corruptions and adversarial attacks \cite{hendrycks2019benchmarking,kurakin2018adversarial}. Consequently, both understanding and mitigating these biases, as well as enhancing the robustness of DNNs, have become important areas of research \cite{hendrycks2019using, geirhos2021partial, rottmann2023detection}. One such bias, rather observed in convolutional neural networks (CNNs) \cite{lecun1998gradient,krizhevsky2012imagenet} than in transformer-based models \cite{dosovitskiy2020image}, is a tendency to rely on low-level texture cues rather than higher-level shapes \cite{geirhos2021partial,naseer2021intriguing,tripathi2023edges,tuli2021convolutional,Zhang2021Delving}. This bias \cite{zeiler2014visualizing,bau2017quantifying} is particularly relevant since {humans are strongly shape-biased} \cite{jacob2021qualitative,geirhos2018imagenettrained}; for instance, we effortlessly recognize a bear whether it appears in a photography, a sketch, or a sculpture, relying primarily on shape rather than texture. Moreover, DNNs that exhibit more shape bias also tend to be more robust against corruptions and adversarial attacks, making shape bias an important factor in model resilience \cite{xu2020robustandgenaralizable,sun2021canshape, chen2020shape}.

\begin{figure}[tb]
    \centering
    \begin{minipage}[b]{0.245\columnwidth}
        \includegraphics[width=\linewidth]{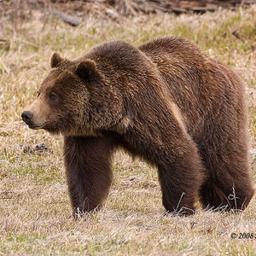}%
    \end{minipage}%
    \hfill 
    \begin{minipage}[b]{0.245\columnwidth}
        \includegraphics[width=\linewidth]{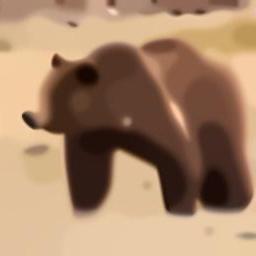}%
    \end{minipage}%
    \hfill
    \begin{minipage}[b]{0.245\columnwidth}
        \includegraphics[width=\linewidth]{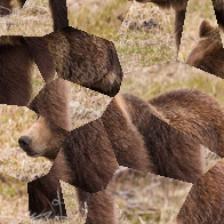}%
    \end{minipage}%
    \hfill
    \begin{minipage}[b]{0.245\columnwidth}
        \includegraphics[width=\linewidth]{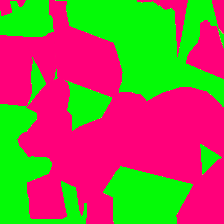}
    \end{minipage}
    \caption{Decomposition of an ImageNet image (left) into its shape cue via Edge Enhancing Diffusion (center left) and texture cue via Voronoi patch shuffling (center right) with the corresponding semantic segmentation mask for the Voronoi-shuffled image (right).}
    \label{fig:bear_decomposition}
\end{figure}

The most widely used measurement of shape and texture biases in image classification was proposed by Geirhos et al.\ \cite{geirhos2018imagenettrained} and relies on measuring performances on \emph{cue-conflict} images. These images are generated via style transfer, where an image showing an object of a given class is combined with a chosen texture from a different class. 
As a result, the final image retains the shape of the original class but adopts the texture of the second class, allowing researchers to see whether a model (or a human) classifies primarily based on shape or texture.
Other approaches rely on evaluating models on datasets with a domain shift that is considered to be more similar to the training domain in terms of shape rather than texture \cite{mishra2020learning}. Examples for such datasets are SketchImageNet and ImageNet-R~\cite{wang2019learning,hendrycks2021many}. Additionally, some methods attempt to remove or reduce specific cues, either by applying regular grid patch shuffling~\cite{islam2021shapeortexture, zhang2019interpretingadversarially, ge2021robust} to reduce shape information or using edge filters to diminish texture information.

In this work, we introduce a new deep-learning-free evaluation procedure for shape and texture biases in DNNs for image classification as well as semantic segmentation.
Instead of using a cue-conflict scheme like \cite{geirhos2018imagenettrained}, we introduce a cue decomposition by constructing a shape and a texture version of a given image. The shape version is obtained by an anisotropic image-diffusion method termed edge-enhancing diffusion (EED) \cite{Weickert1998Anisotropic,heinert2024reducing} that removes texture from a given image while preserving shape. On the other hand, we use a Voronoi diagram \cite{Aurenhammer1991VoronoiDS} of the given image's size and fill its Voronoi cells with random crops of the given image to remove shape information, see also \cref{fig:bear_decomposition} for an illustration.
For a given DNN and a given prediction quality measure, such as accuracy in image classification or mean intersection over union (mIoU) in semantic segmentation, we compute prediction qualities on the texture, shape and original images. Based on those, we define two new evaluation metrics, namely a \emph{cue-decomposition shape bias} for measuring the shape bias in DNNs as well as a \emph{cue-decomposition robustness} that estimates the robustness of a DNN w.r.t.\ image corruptions.

Our experiments, with more than 60 DNNs in total, reveal for image classification that our cue-decomposition shape bias strongly correlates with the cue-conflict metric \cite{geirhos2018imagenettrained}, confirming previous findings, while our cue-decomposition robustness outperforms the cue-conflict metric in terms of predicting robustness to image corruptions. Hereby, our study comprises a wide range of architectures including CNNs, vision transformers (ViTs) and vision language models (VLMs). On top of that, we present a natural extension to semantic segmentation, advancing our understanding of shape and texture biases in semantic segmentation DNNs.
We summarize our contributions as follows:
\begin{itemize}
    \item We introduce a novel evaluation procedure for measuring shape biases. It consists of a cue-decomposition data processing scheme and a novel evaluation metric on top and is available for image classification as well as semantic segmentation.
    \item We provide an additional cue-decomposition robustness metric that allows for DNNs in image classification and semantic segmentation to estimate their robustness w.r.t.\ image corruptions.
    \item In our experiments on ImageNet \cite{deng2009imagenet}, we observe strong correlations to the cue-conflict metric by Geirhos et al.\ while our cue-decomposition robustness clearly outperforms cue-conflict in terms of predicting robustness w.r.t.\ image corruptions. These experiments are complemented with semantic segmentation ones on Cityscapes \cite{Cordts2016Cityscapes} and ADE20k \cite{Zhou2017ADE20k}, providing evidence about texture and shape biases. In semantic segmentation, we observe that the texture cue is less important to predict robustness w.r.t.\ image corruptions.
\end{itemize}
We make our evaluation code as well as a leader board website publicly available with the camera-ready submission.

\section{Related works}

Research on shape and texture biases of DNNs has focused predominantly on image classification, with various methods employed to measure image-cue-related biases. 

\paragraph{Measuring shape bias in image classification}
The already mentioned \emph{cue-conflict} shape bias metric by Geirhos et al.~\cite{geirhos2018imagenettrained} was introduced in a study that revealed that ImageNet pre-trained CNNs tend to be texture biased.
It is a widely used shape bias metric~\cite{geirhos2021partial,Zhang2021Delving,tuli2021convolutional,naseer2021intriguing,jaini2024intriguing,gavrikov2025can}.
However, it is limited by the need for explicit textures that are potentially manually collected.
Another common method to assess texture bias is to consider degraded accuracies on patch-shuffled images,
i.e.\ images where patches following a regular square grid are rearranged for varying grid sizes~\cite{zhang2019interpretingadversarially,luo20defectiveconvolutional,mummadi2021doseenhanced}.
This method is only used for CNNs. It is potentially less meaningful for ViTs as found in \cite{naseer2021intriguing}, where a resilience to this type of image transformation was reported~\cite{naseer2021intriguing}.
The majority of alternative methods include the evaluation on images with preserved shape cues and reduced or altered texture cues.
Such images encompass
style transferred with paintings or textures~\cite{zhang2019interpretingadversarially,mummadi2021doseenhanced,kalischek2023biasbed},
just paintings~\cite{Zhang2021Delving},
sketches~\cite{Zhang2021Delving,kalischek2023biasbed,mishra2020learning},
rapid drawings~\cite{Zhang2021Delving},
edge maps~\cite{kalischek2023biasbed},
silhouettes~\cite{kalischek2023biasbed},
and various levels of saturated images~\cite{zhang2019interpretingadversarially}.

The closest approach to our \emph{cue-decomposition} shape bias metric is the one proposed by Dai et al.~\cite{dai2022rethinking}.
It is based on the estimation of feature contribution for shape and texture cues separately and uses the MeanShiftFilter~\cite{fukunaga1975estimation} for shape cue images and images with blurred edges for texture cues.
However, their focus was on showing that the shape and texture biases of CNNs are task specific. For image classification, we are the first to propose and extensively study an entirely AI-free cue-decomposition-based shape bias evaluation procedure across 43 models. 
Additionally, we introduce a robustness metric to assess DNNs' prediction quality in absence of texture or shape cues.

\paragraph{Measuring shape bias in semantic segmentation}
Estimates of potential shape or texture biases in semantic segmentation are sparsely explored.
The latent-representation-based approach introduced  by Islam et al.~\cite{islam2021shapeortexture} for image classification is potentially but not trivially applicable to semantic segmentation.
In this approach, the number of neurons that encode specific cues is estimated based on the correlation of individual neuron activations on images that share these cues.
For our metric, we evaluate the model in its entirety to facilitate a more comprehensive comparison among different architectures, since we also consider ViTs and VLMs.

Heinert et al.~\cite{heinert2024reducing} employed EED to evaluate and encourage the shape bias of two semantic segmentation models.
The evaluation of shape bias is analogous to the accuracy on texture-reduced images that we use in the classification context. In a similar manner, Theodoridis et al.~\cite{theodoridis2022trapped} utilized degraded performances on style-transferred images to assess shape bias in instance segmentation.
Our shape bias evaluation protocol assesses semantic segmentation models using AI-free shape information extracted via EED, alongside prediction quality measures on Voronoi-shuffled dataset duplicates. We aggregate these measures into the first shape bias and robustness metrics for semantic segmentation, quantifying shape bias and model resilience to the absence of shape or texture cues across 23 models and two datasets.

\paragraph{Analysis of shape bias in image classification}
When it comes to the utilization of shape and texture bias evaluation methods, there are various works, most of which focus on image classification.
Zhang et al.~\cite{zhang2019interpretingadversarially} showed that adversarial training makes CNNs not only more robust against adversarial attacks and image corruptions, but also more shape-biased.
Mummadi et al.~\cite{mummadi2021doseenhanced} have shown for CNNs that shape bias and corruption robustness are to some extent independent.
We find a similar result on wider array of pre-trained DNNs, which also include ViTs and VLMs, where a higher shape bias does not necessarily indicate a higher corruption robustness.

Naseer et al.~\cite{naseer2021intriguing}, Tuli et al.~\cite{tuli2021convolutional} and Zhang et al.~\cite{Zhang2021Delving} compared CNNs to ViT image classifiers and found that ViTs tend to be more shape-biased under the cue-conflict metric.
Naseer et al.\ additionally report a higher corruption robustness of ViTs.
In a large-scale study on out of distribution (OOD) robustness of humans and DNNs Geirhos et al.~\cite{geirhos2021partial} also find an increased shape bias of ViTs over standard CNNs.
For CLIP~\cite{radford2021learning} with a ViT vision encoder, a high shape bias is reported and Gavrikov et al.~\cite{gavrikov2025can} investigate steering of VLM image classification with a focus on shape and texture bias.
They discover that VLMs with ViT vision encoders seem to be more shape biased than their vision encoders alone, which is in line with our results.

We study our novel cue-decomposition shape bias and robustness, evaluating them alongside raw prediction quality measures on cue-specific dataset versions for a diverse range of CNNs, transformers, hybrids, and vision-language models. Additionally, we examine their relationship to the established cue-conflict metric and image corruption robustness, positioning our shape bias metric as highly aligned with prior work while highlighting the superior predictive power of our cue-decomposition robustness metric for general robustness.

\paragraph{Analysis of shape bias in semantic segmentation}
The analysis of shape and texture biases for image segmentation models is underexplored.
In one work by Li et al.~\cite{li2021shapetexture} models are explicitly debiased by training on cue-conflict images and providing the models with both shape and texture labels.
However, they only provide a proof of concept experiment for semantic segmentation.
In a study on OOD texture robustness Theodoridis et al.~\cite{theodoridis2022trapped} found that the task of instance segmentation also leads to a slight shape bias.
Mütze et al.~\cite{mutze2024influence} studied how much segmentation models can still learn when images cues are reduced.
They find that both shape and texture cues are important, which we can confirm in the context of corruption robustness.

For semantic segmentation, we are the first to extensively study any shape-bias metric and our cue-decomposition robustness, offering broad and direct insights into model biases across diverse architectural components, including convolutional, attention-based, and hybrid backbones, as well as various decoder types and object detection fusion.

\section{Cue-decomposition metrics}
To measure the shape bias of trained DNNs, we decompose the image into shape and texture cues and compare the DNN performance on the respective cues. In the following, we describe both cue extraction methods as well as our proposed cue-decomposition metric to measure shape bias.

\paragraph{Shape cue extraction}
To reduce the texture of images but preserve their higher level shape information, we apply a variant of the partial-differential-equation-based (PDE) diffusion process called edge enhancing diffusion (EED) \cite{Weickert1998Anisotropic} as proposed in \cite{heinert2024reducing}. EED iteratively propagates color information along image edges while minimizing color diffusion across them. For a visual example, see \cref{fig:bear_decomposition}.
For simplicity, we introduce the grayscale case: The original image is treated as a function $f$ on a rectangle.  With $f$ as the initial state and zero Neumann boundary conditions, the image is diffused by solving the following PDE iteratively:
\begin{align}
\delta_t u &:= \nabla^T g(\nabla u_\sigma \nabla u_\sigma^T)\nabla u, \\ u(x,0)&\;=f(x) \, . \nonumber
\end{align}
Here $u_\sigma:=  K_\sigma * u$ is smoothed by a Gaussian kernel with standard deviation $\sigma$. Note that, if $g(\nabla u_\sigma \nabla u_\sigma^T) = I $, the diffusion is isotropic since $\delta_t u$ turns into the Laplace operator. The matrix function $g$ is the Charbonnier diffusivity \cite{charbonnier1997deterministic}
$g(s):=1/(\sqrt{1+\frac{s}{\kappa^2}})$ for some contrast parameter $\kappa>0$ and the gradient operator $\nabla$ acts on spatial variables, not time. The diffusion tensor $ g(\nabla u_\sigma \nabla u_\sigma^T)$ is a $2 \times 2$ matrix with eigenvectors parallel and orthogonal to the gradient $\nabla u_\sigma$ and its corresponding eigenvalues are $g(|\nabla u_\sigma|^2)$ and $1$. Along the extension to multichannel images \cite{weickert2006tensor}, for brevity, we leave out the numerical procedure \cite{Weickert1998Anisotropic}, the approach to time and space discretizations \cite[p. 380-391]{weickert20132}, and an explanation of the incorporated spatial kernel smoothing \cite{heinert2024reducing} that we use in this work.

\begin{figure}[tb]
    \centering
    \begin{subfigure}{0.198\linewidth}
        \includegraphics[width=1\linewidth]{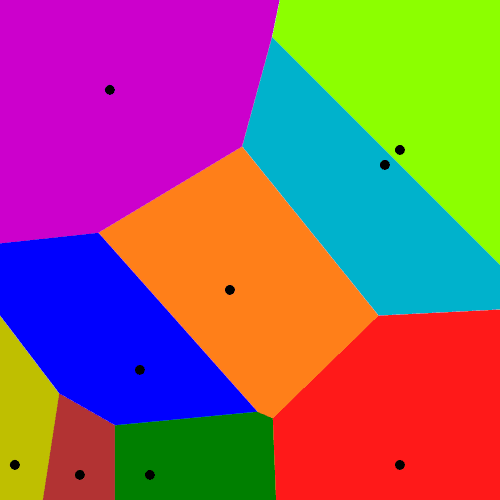}%
    \end{subfigure}%
    \hfill
    \begin{subfigure}{0.198\linewidth}
        \includegraphics[width=1\linewidth]{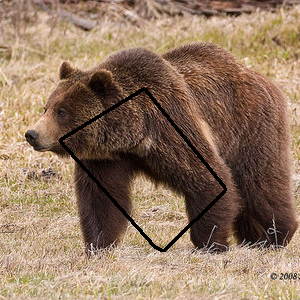}%
    \end{subfigure}%
    \hfill
    \begin{subfigure}{0.198\linewidth}
        \includegraphics[width=1\linewidth]{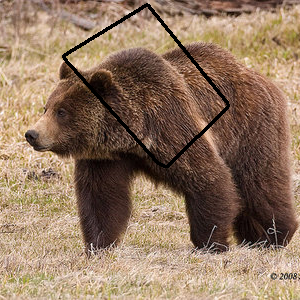}%
    \end{subfigure}%
    \hfill
    \begin{subfigure}{0.198\linewidth}
        \includegraphics[width=1\linewidth]{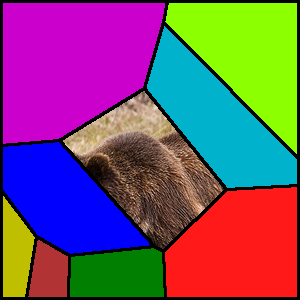}%
    \end{subfigure}%
    \hfill
    \begin{subfigure}{0.198\linewidth}
        \includegraphics[width=1\linewidth]{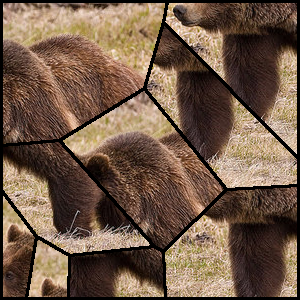}%
    \end{subfigure}%
    \caption{Voronoi shuffling method: The image is decomposed into $N$ Voronoi cells. Each Voronoi cell is filled with random crops of the original image by randomly shifting the cell on the image.}
    \label{fig:voronoi-shuffling}
\end{figure}

\paragraph{Texture cue extraction}
To generate a texture version of a given image, we shuffle the image based on a Voronoi diagram \cite{Aurenhammer1991VoronoiDS}. For this purpose, we firstly sample $N$ random pixel-coordinates from the image to generate a set $\mathcal{S}$ of so-called sites. The corresponding Voronoi diagram $V(\mathcal{S})$ is generated by dividing the image into $N$ cells $V(p), p \in \mathcal{S}$ by assigning each image pixel to the nearest site $p$ w.r.t.\ the Euclidean distance. For each site $p \in \mathcal{S}$, we define its corresponding Voronoi cell
\begin{equation}
        V(p) = \{ x \in \mathbb{R}^2 : \| x - p \|_2 \leq \| x - q \|_2,\ \forall\, q \in \mathcal{S}^- \} \ ,
\end{equation}
with $ \mathcal{S}^- = \mathcal{S} \setminus \{p\}$.
All points with equidistance to multiple sites form the cell boundaries and define the partition by convex polygons.
Finally, we restrict the Voronoi diagram to the original image by cropping it to the original image size.
To generate \emph{Voronoi-shuffled images}, each cell of the Voronoi diagram is filled with random image parts. 
The image content is chosen by randomly shifting the cell on the original image and pasting the resulting image content in the Voronoi cell. This results in a shuffled image and therefore reduces the shape information in the image. This process is visualized in \cref{fig:voronoi-shuffling}

\paragraph{Cue-decomposition metrics}

Given 1) the cue-decomposition consisting of a texture-cue dataset created by EED and a shape-cue dataset of Voronoi-shuffled images, 2) a deep neural network $h$ and 3) a prediction quality measure such as accuracy in image classification or mIoU in semantic segmentation, we  measure prediction qualities $Q_T(h)$ on the texture data, $Q_S(h)$ on the shape data and $Q_O(h)$ on the original data (without any extraction method applied to it). In principle, we could measure the shape bias relative to the texture biases via $Q_S(h)/(Q_S(h) + Q_T(h))$ and compare these quantities over different networks $h$. However, it is not guaranteed that values of $Q_S$ and $Q_T$ are in the same range, which hinders the interpretability of the above quotient. Hence, we normalize the prediction qualities. Let $\mathcal{H}$ be a finite set of DNNs. Defining 
\begin{equation}
    s=\frac{1}{|\mathcal{H}|} \sum_{h \in \mathcal{H}} Q_S(h) \quad \text{and} \quad  t = \frac{1}{|\mathcal{H}|} \sum_{h \in \mathcal{H}} Q_T(h)
    \label{eq:s_and_t}
\end{equation}
we define our cue-decomposition shape bias as
\begin{equation}
    S_\mathrm{cd} = \frac{\frac{1}{s}Q_S}{\frac{1}{s}Q_S + \frac{1}{t}Q_T } \, .
    \label{eq:s_cd}
\end{equation}
This quantity has the advantage that $S_\mathrm{cd}(h)>0.5$ indicates that $h$ is rather shape than texture biased and vice versa. Using classification accuracy or mIoU, this quantity $S_\mathrm{cd}$ can be computed for both tasks.

It has already been shown in \cite{geirhos2018imagenettrained} that DNNs with less texture bias tend to be more robust w.r.t.\ image corruptions. We estimate the robustness of $f$ w.r.t.\ image corruptions by
\begin{equation}
    R_\mathrm{cd} = \frac{Q_S + Q_T}{2 \cdot Q_O} \, .
    \label{eq:r_cd}
\end{equation}
Note that, this quantity is already normalized via $Q_O$ (assuming $Q_S \leq Q_O$ and $Q_T \leq Q_O$) and again available for both image classification and semantic segmentation.

\section{Numerical experiments}
In the following, we study the shape bias of a broad range of vision DNNs on three different datasets and compare our cue-decomposition shape bias procedure with the cue-conflict shape bias metric of Geirhos et al.\ \cite{geirhos2018imagenettrained}: 
\begin{equation}
   \frac{\mathrm{TP}_S}{{\mathrm{TP}}_S + \mathrm{TP}_T}\ ,
   \label{eq:geihros}
\end{equation}
where $\mathrm{TP}_*$ is the number of correct shape ($*=S$) or texture ($*=T$) predictions on a cue-conflict ImageNet version. Furthermore, we investigate the relative robustness performance of these DNNs under image corruption.

\paragraph{Models and datasets}
In our experiments, we evaluate $43$ pre-trained image classification and $23$ semantic segmentation models.
The classification models that we consider comprise $13$ CNNs, $11$ vision transformers, $7$ hybrids of CNN and transformer and $12$ VLMs with zero-shot classification.
All but the VLMs were pre-trained on ImageNet, a well established dataset for pre-training including $1,\!000$ distinct classes.
The semantic segmentation models used in our experiments include $10$ CNNs, $8$ vision transformers and $5$ other models. 
As datasets for the semantic segmentation task, we consider Cityscapes~\cite{Cordts2016Cityscapes} and ADE20k~\cite{Zhou2017ADE20k}. 
Cityscapes is a street scene dataset with $2,\!975$ training and 500 validation images where 19 classes commonly appearing in driving scenarios are semantically labeled. ADE20k comprises more than $20,\!000$ in- and outdoor scenes with $150$ labeled semantic categories.
For better comprehension, we have reduced the number of classification models in this section to include only one representative of each architecture type, e.g.\ we present results for a ResNet101 only and omit other ResNet variants. The full tables including all classification models are provided in the appendix.
All calculated rank correlations are based on the full tables.

\begin{table*}[tb]
    \centering
    \scalebox{0.85}{\begin{tabular}{llrrrrrr|rrrrrr}
\toprule
 \multicolumn{2}{c}{Dataset: \bfseries ImageNet} & \multicolumn{3}{c}{\bfseries Accuracy} & \multicolumn{2}{c}{\bfseries C.-Dec.} & \bfseries C.-Conf. & \multicolumn{6}{c}{\bfseries Relative Robustness} \\
 &  & \bfseries \(\boldsymbol{Q_O}\) & \bfseries \(\boldsymbol{Q_S}\) & \bfseries \(\boldsymbol{Q_T}\) & \bfseries \(\boldsymbol{S_\mathrm{cd}}\) & \bfseries \(\boldsymbol{R_\mathrm{cd}}\) & \bfseries Shape & \bfseries Cont. & \bfseries High & \bfseries Low & \bfseries Noise & \bfseries Phase & \bfseries Mean \\
Arch. & Model &  &  &  &  &  & \bfseries Bias &  & \bfseries Pass & \bfseries Pass &  & \bfseries Noise &  \\
\midrule
\multirow[c]{6}{*}{CNN} & ConvNeXt L \cite{zhuang2022aconvnet} & {\cellcolor[HTML]{FFCD82}} \color[HTML]{000000} \color{black} 0.996 & {\cellcolor[HTML]{BBBBFF}} \color[HTML]{000000} \color{black} 0.838 & {\cellcolor[HTML]{FFCF86}} \color[HTML]{000000} \color{black} 0.969 & {\cellcolor[HTML]{C5C5FF}} \color[HTML]{000000} \color{black} 0.563 & {\cellcolor[HTML]{FFD28E}} \color[HTML]{000000} \color{black} 0.907 & {\cellcolor[HTML]{D8D8FF}} \color[HTML]{000000} \color{black} 0.348 & {\cellcolor[HTML]{FFD28E}} \color[HTML]{000000} \color{black} 0.865 & {\cellcolor[HTML]{B6B6FF}} \color[HTML]{000000} \color{black} 0.800 & {\cellcolor[HTML]{FFD088}} \color[HTML]{000000} \color{black} 0.735 & {\cellcolor[HTML]{B6B6FF}} \color[HTML]{000000} \color{black} 0.962 & {\cellcolor[HTML]{FFD28E}} \color[HTML]{000000} \color{black} 0.558 & {\cellcolor[HTML]{B7B7FF}} \color[HTML]{000000} \color{black} 0.784 \\
\cline{2-2}
 & RegNetY \cite{radosavovic2020designing} & {\cellcolor[HTML]{FFCD82}} \color[HTML]{000000} \color{black} 0.996 & {\cellcolor[HTML]{C0C0FF}} \color[HTML]{000000} \color{black} 0.797 & {\cellcolor[HTML]{FFCC80}} \color[HTML]{000000} \color{black} 0.986 & {\cellcolor[HTML]{C9C9FF}} \color[HTML]{000000} \color{black} 0.546 & {\cellcolor[HTML]{FFD391}} \color[HTML]{000000} \color{black} 0.895 & {\cellcolor[HTML]{C5C5FF}} \color[HTML]{000000} \color{black} 0.456 & {\cellcolor[HTML]{FFE7C3}} \color[HTML]{000000} \color{black} 0.774 & {\cellcolor[HTML]{D0D0FF}} \color[HTML]{000000} \color{black} 0.600 & {\cellcolor[HTML]{FFDAA2}} \color[HTML]{000000} \color{black} 0.694 & {\cellcolor[HTML]{BCBCFF}} \color[HTML]{000000} \color{black} 0.940 & {\cellcolor[HTML]{FFDBA6}} \color[HTML]{000000} \color{black} 0.531 & {\cellcolor[HTML]{CBCBFF}} \color[HTML]{000000} \color{black} 0.708 \\
\cline{2-2}
 & ResNeXt101 \cite{Xie2016} & {\cellcolor[HTML]{FFCF86}} \color[HTML]{000000} \color{black} 0.995 & {\cellcolor[HTML]{CDCDFF}} \color[HTML]{000000} \color{black} 0.678 & {\cellcolor[HTML]{FFDCA7}} \color[HTML]{000000} \color{black} 0.869 & {\cellcolor[HTML]{CBCBFF}} \color[HTML]{000000} \color{black} 0.537 & {\cellcolor[HTML]{FFE0B2}} \color[HTML]{000000} \color{black} 0.777 & {\cellcolor[HTML]{D0D0FF}} \color[HTML]{000000} \color{black} 0.391 & {\cellcolor[HTML]{FFE1B3}} \color[HTML]{000000} \color{black} 0.802 & {\cellcolor[HTML]{DADAFF}} \color[HTML]{000000} \color{black} 0.515 & {\cellcolor[HTML]{FFDAA3}} \color[HTML]{000000} \color{black} 0.693 & {\cellcolor[HTML]{C8C8FF}} \color[HTML]{000000} \color{black} 0.890 & {\cellcolor[HTML]{FFE6C2}} \color[HTML]{000000} \color{black} 0.500 & {\cellcolor[HTML]{D3D3FF}} \color[HTML]{000000} \color{black} 0.680 \\
\cline{2-2}
 & DPN92 \cite{Chen2017} & {\cellcolor[HTML]{FFD28F}} \color[HTML]{000000} \color{black} 0.992 & {\cellcolor[HTML]{EBEBFF}} \color[HTML]{000000} \color{black} 0.408 & {\cellcolor[HTML]{FFEBCE}} \color[HTML]{000000} \color{black} 0.754 & {\cellcolor[HTML]{E0E0FF}} \color[HTML]{000000} \color{black} 0.446 & {\cellcolor[HTML]{FFF6EA}} \color[HTML]{000000} \color{black} 0.585 & {\cellcolor[HTML]{E7E7FF}} \color[HTML]{000000} \color{black} 0.263 & {\cellcolor[HTML]{FFEED3}} \color[HTML]{000000} \color{black} 0.746 & {\cellcolor[HTML]{FAFAFF}} \color[HTML]{000000} \color{black} 0.269 & {\cellcolor[HTML]{FFE9C7}} \color[HTML]{000000} \color{black} 0.637 & {\cellcolor[HTML]{D9D9FF}} \color[HTML]{000000} \color{black} 0.824 & {\cellcolor[HTML]{FFF3E0}} \color[HTML]{000000} \color{black} 0.465 & {\cellcolor[HTML]{EBEBFF}} \color[HTML]{000000} \color{black} 0.588 \\
\cline{2-2}
 & ResNet101 \cite{he2015deepresiduallearningimage} & {\cellcolor[HTML]{FFD698}} \color[HTML]{000000} \color{black} 0.989 & {\cellcolor[HTML]{F4F4FF}} \color[HTML]{000000} \color{black} 0.326 & {\cellcolor[HTML]{FFDFAE}} \color[HTML]{000000} \color{black} 0.848 & {\cellcolor[HTML]{F2F2FF}} \color[HTML]{000000} \color{black} 0.363 & {\cellcolor[HTML]{FFF6E8}} \color[HTML]{000000} \color{black} 0.593 & {\cellcolor[HTML]{F5F5FF}} \color[HTML]{000000} \color{black} 0.181 & {\cellcolor[HTML]{FFDBA6}} \color[HTML]{000000} \color{black} 0.826 & {\cellcolor[HTML]{D8D8FF}} \color[HTML]{000000} \color{black} 0.532 & {\cellcolor[HTML]{FFEED4}} \color[HTML]{000000} \color{black} 0.616 & {\cellcolor[HTML]{CBCBFF}} \color[HTML]{000000} \color{black} 0.877 & {\cellcolor[HTML]{FFEFD8}} \color[HTML]{000000} \color{black} 0.475 & {\cellcolor[HTML]{D6D6FF}} \color[HTML]{000000} \color{black} 0.665 \\
\cline{2-2}
 & VGG19 \cite{Simonyan2014VeryDC} & {\cellcolor[HTML]{FFDDAB}} \color[HTML]{000000} \color{black} 0.983 & {\cellcolor[HTML]{FFFFFF}} \color[HTML]{000000} \color{black} 0.228 & {\cellcolor[HTML]{FFE8C6}} \color[HTML]{000000} \color{black} 0.774 & {\cellcolor[HTML]{FFFFFF}} \color[HTML]{000000} \color{black} 0.304 & {\cellcolor[HTML]{FFFFFF}} \color[HTML]{000000} \color{black} 0.510 & {\cellcolor[HTML]{FFFFFF}} \color[HTML]{000000} \color{black} 0.124 & {\cellcolor[HTML]{FFFFFF}} \color[HTML]{000000} \color{black} 0.670 & {\cellcolor[HTML]{FFFFFF}} \color[HTML]{000000} \color{black} 0.230 & {\cellcolor[HTML]{FFFFFF}} \color[HTML]{000000} \color{black} 0.548 & {\cellcolor[HTML]{FFFFFF}} \color[HTML]{000000} \color{black} 0.670 & {\cellcolor[HTML]{FFFFFF}} \color[HTML]{000000} \color{black} 0.430 & {\cellcolor[HTML]{FFFFFF}} \color[HTML]{000000} \color{black} 0.510 \\
\cline{1-2} \cline{2-2}
\multirow[c]{5}{*}{\shortstack{Vision\\ Transf.}} & EVA02 L \cite{EVA02} & \bfseries {\cellcolor[HTML]{FFCC80}} \color[HTML]{000000} \color{black} 0.997 & \bfseries {\cellcolor[HTML]{B2B2FF}} \color[HTML]{000000} \color{black} 0.921 & \bfseries {\cellcolor[HTML]{FFCC80}} \color[HTML]{000000} \color{black} 0.988 & {\cellcolor[HTML]{C2C2FF}} \color[HTML]{000000} \color{black} 0.581 & \bfseries {\cellcolor[HTML]{FFCC80}} \color[HTML]{000000} \color{black} 0.957 & {\cellcolor[HTML]{B6B6FF}} \color[HTML]{000000} \color{black} 0.542 & \bfseries {\cellcolor[HTML]{FFCC80}} \color[HTML]{000000} \color{black} 0.892 & \bfseries {\cellcolor[HTML]{B2B2FF}} \color[HTML]{000000} \color{black} 0.827 & {\cellcolor[HTML]{FFCD82}} \color[HTML]{000000} \color{black} 0.745 & {\cellcolor[HTML]{B3B3FF}} \color[HTML]{000000} \color{black} 0.975 & \bfseries {\cellcolor[HTML]{FFCC80}} \color[HTML]{000000} \color{black} 0.575 & \bfseries {\cellcolor[HTML]{B2B2FF}} \color[HTML]{000000} \color{black} 0.803 \\
\cline{2-2}
 & BEiT \cite{bao2021beit} & {\cellcolor[HTML]{FFCD82}} \color[HTML]{000000} \color{black} 0.996 & {\cellcolor[HTML]{C2C2FF}} \color[HTML]{000000} \color{black} 0.782 & {\cellcolor[HTML]{FFCE84}} \color[HTML]{000000} \color{black} 0.974 & {\cellcolor[HTML]{CACAFF}} \color[HTML]{000000} \color{black} 0.544 & {\cellcolor[HTML]{FFD595}} \color[HTML]{000000} \color{black} 0.881 & {\cellcolor[HTML]{C0C0FF}} \color[HTML]{000000} \color{black} 0.486 & {\cellcolor[HTML]{FFD698}} \color[HTML]{000000} \color{black} 0.850 & {\cellcolor[HTML]{BBBBFF}} \color[HTML]{000000} \color{black} 0.764 & {\cellcolor[HTML]{FFD494}} \color[HTML]{000000} \color{black} 0.717 & {\cellcolor[HTML]{B5B5FF}} \color[HTML]{000000} \color{black} 0.964 & {\cellcolor[HTML]{FFD08A}} \color[HTML]{000000} \color{black} 0.563 & {\cellcolor[HTML]{BBBBFF}} \color[HTML]{000000} \color{black} 0.772 \\
\cline{2-2}
 & ViT B16 \cite{wu2020visual} & {\cellcolor[HTML]{FFD08A}} \color[HTML]{000000} \color{black} 0.994 & {\cellcolor[HTML]{CBCBFF}} \color[HTML]{000000} \color{black} 0.699 & {\cellcolor[HTML]{FFD392}} \color[HTML]{000000} \color{black} 0.931 & {\cellcolor[HTML]{CECEFF}} \color[HTML]{000000} \color{black} 0.528 & {\cellcolor[HTML]{FFDCA6}} \color[HTML]{000000} \color{black} 0.820 & {\cellcolor[HTML]{C5C5FF}} \color[HTML]{000000} \color{black} 0.455 & {\cellcolor[HTML]{FFD18D}} \color[HTML]{000000} \color{black} 0.869 & {\cellcolor[HTML]{CFCFFF}} \color[HTML]{000000} \color{black} 0.607 & {\cellcolor[HTML]{FFD699}} \color[HTML]{000000} \color{black} 0.709 & {\cellcolor[HTML]{C4C4FF}} \color[HTML]{000000} \color{black} 0.909 & {\cellcolor[HTML]{FFD28F}} \color[HTML]{000000} \color{black} 0.557 & {\cellcolor[HTML]{C5C5FF}} \color[HTML]{000000} \color{black} 0.730 \\
\cline{2-2}
 & Swin B \cite{ze2021swintransformer} & {\cellcolor[HTML]{FFD088}} \color[HTML]{000000} \color{black} 0.994 & {\cellcolor[HTML]{CDCDFF}} \color[HTML]{000000} \color{black} 0.683 & {\cellcolor[HTML]{FFD699}} \color[HTML]{000000} \color{black} 0.911 & {\cellcolor[HTML]{CECEFF}} \color[HTML]{000000} \color{black} 0.527 & {\cellcolor[HTML]{FFDEAC}} \color[HTML]{000000} \color{black} 0.802 & {\cellcolor[HTML]{DDDDFF}} \color[HTML]{000000} \color{black} 0.321 & {\cellcolor[HTML]{FFD79B}} \color[HTML]{000000} \color{black} 0.845 & {\cellcolor[HTML]{D4D4FF}} \color[HTML]{000000} \color{black} 0.564 & {\cellcolor[HTML]{FFD699}} \color[HTML]{000000} \color{black} 0.709 & {\cellcolor[HTML]{B9B9FF}} \color[HTML]{000000} \color{black} 0.950 & {\cellcolor[HTML]{FFDAA2}} \color[HTML]{000000} \color{black} 0.536 & {\cellcolor[HTML]{C8C8FF}} \color[HTML]{000000} \color{black} 0.721 \\
\cline{2-2}
 & Inception v3 \cite{szegedy2015rethinking} & {\cellcolor[HTML]{FFD494}} \color[HTML]{000000} \color{black} 0.990 & {\cellcolor[HTML]{E2E2FF}} \color[HTML]{000000} \color{black} 0.488 & {\cellcolor[HTML]{FFF7EB}} \color[HTML]{000000} \color{black} 0.668 & {\cellcolor[HTML]{CFCFFF}} \color[HTML]{000000} \color{black} 0.521 & {\cellcolor[HTML]{FFF7EA}} \color[HTML]{000000} \color{black} 0.584 & {\cellcolor[HTML]{E3E3FF}} \color[HTML]{000000} \color{black} 0.285 & {\cellcolor[HTML]{FFF1DD}} \color[HTML]{000000} \color{black} 0.730 & {\cellcolor[HTML]{FEFEFF}} \color[HTML]{000000} \color{black} 0.241 & {\cellcolor[HTML]{FFF7EC}} \color[HTML]{000000} \color{black} 0.579 & {\cellcolor[HTML]{D3D3FF}} \color[HTML]{000000} \color{black} 0.849 & {\cellcolor[HTML]{FFF2DE}} \color[HTML]{000000} \color{black} 0.467 & {\cellcolor[HTML]{EEEEFF}} \color[HTML]{000000} \color{black} 0.573 \\
\cline{1-2} \cline{2-2}
\multirow[c]{5}{*}{VLM} & FLAVA-full \cite{singh2022flava} & {\cellcolor[HTML]{FFFFFF}} \color[HTML]{000000} \color{black} 0.955 & {\cellcolor[HTML]{C5C5FF}} \color[HTML]{000000} \color{black} 0.752 & {\cellcolor[HTML]{FFFFFF}} \color[HTML]{000000} \color{black} 0.605 & \bfseries {\cellcolor[HTML]{B2B2FF}} \color[HTML]{000000} \color{black} 0.649 & {\cellcolor[HTML]{FFE8C6}} \color[HTML]{000000} \color{black} 0.711 & {\cellcolor[HTML]{B8B8FF}} \color[HTML]{000000} \color{black} 0.530 & {\cellcolor[HTML]{FFD9A0}} \color[HTML]{000000} \color{black} 0.836 & {\cellcolor[HTML]{D9D9FF}} \color[HTML]{000000} \color{black} 0.525 & {\cellcolor[HTML]{FFDFB0}} \color[HTML]{000000} \color{black} 0.673 & {\cellcolor[HTML]{D4D4FF}} \color[HTML]{000000} \color{black} 0.841 & {\cellcolor[HTML]{FFF5E7}} \color[HTML]{000000} \color{black} 0.458 & {\cellcolor[HTML]{D6D6FF}} \color[HTML]{000000} \color{black} 0.666 \\
\cline{2-2}
 & Align-base \cite{jia2021scaling} & {\cellcolor[HTML]{FFEED6}} \color[HTML]{000000} \color{black} 0.969 & {\cellcolor[HTML]{CECEFF}} \color[HTML]{000000} \color{black} 0.677 & {\cellcolor[HTML]{FFFDFA}} \color[HTML]{000000} \color{black} 0.621 & {\cellcolor[HTML]{B9B9FF}} \color[HTML]{000000} \color{black} 0.618 & {\cellcolor[HTML]{FFEDD2}} \color[HTML]{000000} \color{black} 0.670 & {\cellcolor[HTML]{C0C0FF}} \color[HTML]{000000} \color{black} 0.489 & {\cellcolor[HTML]{FFDAA3}} \color[HTML]{000000} \color{black} 0.829 & {\cellcolor[HTML]{CDCDFF}} \color[HTML]{000000} \color{black} 0.621 & {\cellcolor[HTML]{FFF8EC}} \color[HTML]{000000} \color{black} 0.578 & {\cellcolor[HTML]{DDDDFF}} \color[HTML]{000000} \color{black} 0.806 & {\cellcolor[HTML]{FFE9C8}} \color[HTML]{000000} \color{black} 0.493 & {\cellcolor[HTML]{D6D6FF}} \color[HTML]{000000} \color{black} 0.665 \\
\cline{2-2}
 & CLIP ViT-B32 \cite{radford2021learning} & {\cellcolor[HTML]{FFEBCC}} \color[HTML]{000000} \color{black} 0.972 & {\cellcolor[HTML]{C0C0FF}} \color[HTML]{000000} \color{black} 0.799 & {\cellcolor[HTML]{FFEBCC}} \color[HTML]{000000} \color{black} 0.758 & {\cellcolor[HTML]{BBBBFF}} \color[HTML]{000000} \color{black} 0.611 & {\cellcolor[HTML]{FFDEAC}} \color[HTML]{000000} \color{black} 0.801 & \bfseries {\cellcolor[HTML]{B2B2FF}} \color[HTML]{000000} \color{black} 0.564 & {\cellcolor[HTML]{FFE1B5}} \color[HTML]{000000} \color{black} 0.799 & {\cellcolor[HTML]{DADAFF}} \color[HTML]{000000} \color{black} 0.523 & \bfseries {\cellcolor[HTML]{FFCC80}} \color[HTML]{000000} \color{black} 0.749 & {\cellcolor[HTML]{C6C6FF}} \color[HTML]{000000} \color{black} 0.900 & {\cellcolor[HTML]{FFE3B8}} \color[HTML]{000000} \color{black} 0.510 & {\cellcolor[HTML]{CECEFF}} \color[HTML]{000000} \color{black} 0.696 \\
\cline{2-2}
 & SigLIP-base \cite{zhai2023sigmoid} & {\cellcolor[HTML]{FFE7C2}} \color[HTML]{000000} \color{black} 0.975 & {\cellcolor[HTML]{BCBCFF}} \color[HTML]{000000} \color{black} 0.829 & {\cellcolor[HTML]{FFE3B9}} \color[HTML]{000000} \color{black} 0.816 & {\cellcolor[HTML]{BDBDFF}} \color[HTML]{000000} \color{black} 0.602 & {\cellcolor[HTML]{FFD9A0}} \color[HTML]{000000} \color{black} 0.843 & {\cellcolor[HTML]{C2C2FF}} \color[HTML]{000000} \color{black} 0.478 & {\cellcolor[HTML]{FFDCA8}} \color[HTML]{000000} \color{black} 0.823 & {\cellcolor[HTML]{C2C2FF}} \color[HTML]{000000} \color{black} 0.704 & {\cellcolor[HTML]{FFE1B4}} \color[HTML]{000000} \color{black} 0.667 & {\cellcolor[HTML]{CDCDFF}} \color[HTML]{000000} \color{black} 0.870 & {\cellcolor[HTML]{FFE6C2}} \color[HTML]{000000} \color{black} 0.500 & {\cellcolor[HTML]{CACAFF}} \color[HTML]{000000} \color{black} 0.713 \\
\cline{2-2}
 & CLIP RN101 \cite{radford2021learning} & {\cellcolor[HTML]{FFF1DC}} \color[HTML]{000000} \color{black} 0.967 & {\cellcolor[HTML]{D5D5FF}} \color[HTML]{000000} \color{black} 0.605 & {\cellcolor[HTML]{FFE8C6}} \color[HTML]{000000} \color{black} 0.776 & {\cellcolor[HTML]{CBCBFF}} \color[HTML]{000000} \color{black} 0.537 & {\cellcolor[HTML]{FFE8C5}} \color[HTML]{000000} \color{black} 0.714 & {\cellcolor[HTML]{EAEAFF}} \color[HTML]{000000} \color{black} 0.246 & {\cellcolor[HTML]{FFE7C3}} \color[HTML]{000000} \color{black} 0.774 & {\cellcolor[HTML]{D4D4FF}} \color[HTML]{000000} \color{black} 0.563 & {\cellcolor[HTML]{FFF7EB}} \color[HTML]{000000} \color{black} 0.580 & {\cellcolor[HTML]{E0E0FF}} \color[HTML]{000000} \color{black} 0.797 & {\cellcolor[HTML]{FFFAF3}} \color[HTML]{000000} \color{black} 0.444 & {\cellcolor[HTML]{DFDFFF}} \color[HTML]{000000} \color{black} 0.631 \\
\cline{1-2} \cline{2-2}
\multirow[c]{4}{*}{Hybrid} & SEResNeXt \cite{hu2019squeezeandexcitationnetworks} & {\cellcolor[HTML]{FFCE84}} \color[HTML]{000000} \color{black} 0.996 & {\cellcolor[HTML]{C6C6FF}} \color[HTML]{000000} \color{black} 0.747 & {\cellcolor[HTML]{FFD08B}} \color[HTML]{000000} \color{black} 0.955 & {\cellcolor[HTML]{CBCBFF}} \color[HTML]{000000} \color{black} 0.538 & {\cellcolor[HTML]{FFD89C}} \color[HTML]{000000} \color{black} 0.855 & {\cellcolor[HTML]{C9C9FF}} \color[HTML]{000000} \color{black} 0.436 & {\cellcolor[HTML]{FFDDAA}} \color[HTML]{000000} \color{black} 0.818 & {\cellcolor[HTML]{DADAFF}} \color[HTML]{000000} \color{black} 0.518 & {\cellcolor[HTML]{FFDDAA}} \color[HTML]{000000} \color{black} 0.682 & {\cellcolor[HTML]{B7B7FF}} \color[HTML]{000000} \color{black} 0.957 & {\cellcolor[HTML]{FFDAA2}} \color[HTML]{000000} \color{black} 0.536 & {\cellcolor[HTML]{CDCDFF}} \color[HTML]{000000} \color{black} 0.702 \\
\cline{2-2}
 & CAFormer b36 \cite{yu2022metaformerbaselines} & {\cellcolor[HTML]{FFCD82}} \color[HTML]{000000} \color{black} 0.997 & {\cellcolor[HTML]{C4C4FF}} \color[HTML]{000000} \color{black} 0.758 & {\cellcolor[HTML]{FFCD81}} \color[HTML]{000000} \color{black} 0.983 & {\cellcolor[HTML]{CCCCFF}} \color[HTML]{000000} \color{black} 0.534 & {\cellcolor[HTML]{FFD597}} \color[HTML]{000000} \color{black} 0.873 & {\cellcolor[HTML]{CDCDFF}} \color[HTML]{000000} \color{black} 0.415 & {\cellcolor[HTML]{FFD18D}} \color[HTML]{000000} \color{black} 0.868 & {\cellcolor[HTML]{C3C3FF}} \color[HTML]{000000} \color{black} 0.696 & {\cellcolor[HTML]{FFD392}} \color[HTML]{000000} \color{black} 0.721 & \bfseries {\cellcolor[HTML]{B2B2FF}} \color[HTML]{000000} \color{black} 0.977 & {\cellcolor[HTML]{FFD28F}} \color[HTML]{000000} \color{black} 0.557 & {\cellcolor[HTML]{BCBCFF}} \color[HTML]{000000} \color{black} 0.764 \\
\cline{2-2}
 & SENet154 \cite{hu2019squeezeandexcitationnetworks} & {\cellcolor[HTML]{FFD290}} \color[HTML]{000000} \color{black} 0.992 & {\cellcolor[HTML]{E9E9FF}} \color[HTML]{000000} \color{black} 0.427 & {\cellcolor[HTML]{FFE3BA}} \color[HTML]{000000} \color{black} 0.812 & {\cellcolor[HTML]{E1E1FF}} \color[HTML]{000000} \color{black} 0.439 & {\cellcolor[HTML]{FFF2DE}} \color[HTML]{000000} \color{black} 0.625 & {\cellcolor[HTML]{E7E7FF}} \color[HTML]{000000} \color{black} 0.261 & {\cellcolor[HTML]{FFECD0}} \color[HTML]{000000} \color{black} 0.753 & {\cellcolor[HTML]{FEFEFF}} \color[HTML]{000000} \color{black} 0.238 & {\cellcolor[HTML]{FFE7C2}} \color[HTML]{000000} \color{black} 0.644 & {\cellcolor[HTML]{D7D7FF}} \color[HTML]{000000} \color{black} 0.829 & {\cellcolor[HTML]{FFEED4}} \color[HTML]{000000} \color{black} 0.479 & {\cellcolor[HTML]{EBEBFF}} \color[HTML]{000000} \color{black} 0.588 \\
\cline{2-2}
 & ConvFormer b36 \cite{yu2022metaformerbaselines} & {\cellcolor[HTML]{FFCF88}} \color[HTML]{000000} \color{black} 0.994 & {\cellcolor[HTML]{E3E3FF}} \color[HTML]{000000} \color{black} 0.477 & {\cellcolor[HTML]{FFD08B}} \color[HTML]{000000} \color{black} 0.955 & {\cellcolor[HTML]{E4E4FF}} \color[HTML]{000000} \color{black} 0.426 & {\cellcolor[HTML]{FFE7C3}} \color[HTML]{000000} \color{black} 0.720 & {\cellcolor[HTML]{E5E5FF}} \color[HTML]{000000} \color{black} 0.277 & {\cellcolor[HTML]{FFD698}} \color[HTML]{000000} \color{black} 0.850 & {\cellcolor[HTML]{DEDEFF}} \color[HTML]{000000} \color{black} 0.488 & {\cellcolor[HTML]{FFE6C0}} \color[HTML]{000000} \color{black} 0.648 & {\cellcolor[HTML]{C7C7FF}} \color[HTML]{000000} \color{black} 0.895 & {\cellcolor[HTML]{FFD89E}} \color[HTML]{000000} \color{black} 0.541 & {\cellcolor[HTML]{D1D1FF}} \color[HTML]{000000} \color{black} 0.684 \\
\cline{1-2} \cline{2-2}
\bottomrule
\end{tabular}
}
    \caption{
    (left) Cue-decomposition shape biases and relative robustness performances of classification models with components of the metrics defined in \cref{eq:s_cd} and \cref{eq:r_cd}
    as well as the cue-conflict shape bias metric by Geirhos et al.\ (C.-Conf. Shape Bias) for comparison.
    (right) Relative robustness calculated for different types of image corruptions like contrast changes (Cont.) and averaged over all corruption types in the right-most column (Mean).
    The full table with $43$ models can be found in the appendix.
    }
    \label{tab:classMetricsAndRob}
\end{table*}

\paragraph{Implementation details}
If not stated otherwise, we use $N=32$ sites to generate Voronoi diagrams with $32$ cells. Furthermore, we generate EED images using contrast parameter $\kappa=1/15$, Gaussian kernel size $5$ and standard deviation $\sigma=\sqrt{5}$; if not stated otherwise, EED images are diffused with $16384$ steps for classification experiments and $5792$ for semantic segmentation. All experiments were conducted using mmsegmentation~\cite{mmseg2020}, the timm classification framework by huggingface \cite{rw2019timm, huggingface_github} and the OpenAI repository \cite{clip_github}. Classification results for VLMs are achieved by zero-shot classification. The input text is defined by the class labels of the given dataset in the form ``a photo of a \texttt{class label}''.
For ImageNet we apply the standard pre-processing which consists of a resizing to $256 \times 256$ followed by a quadratic center crop of $224 \times 224$ pixels. ADE20k images have been resized to $512$ on their shorter side while preserving aspect ratio and then center cropped to size 512 $\times 512$ before any further processing.

\subsection{Image classification results}\label{sec:classification}
In this section, we study the cue-decomposition shape bias as well as the cue-decomposition robustness for a choice of different models on the ImageNet dataset. For the sake of comparison, we selected the same subset of ImageNet as used in \cite{geirhos2018imagenettrained}, consisting of $16$ classes and $1,\!200$ images. For those images, we compute the EED-transformed and the Voronoi-shuffled images to compute our cue-decomposition metrics.

\begin{figure}[tb]
    \centering
    \footnotesize
    \begin{minipage}[b]{0.16\columnwidth}
    \centering
        \includegraphics[trim=2.00cm 0cm 0.5cm 0cm, clip, width=\linewidth]{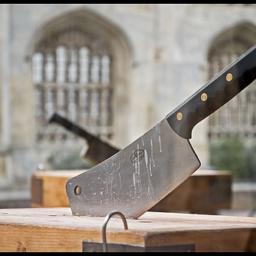}
        original\\ \phantom{original}
    \end{minipage}%
    \hfill 
    \begin{minipage}[b]{0.16\columnwidth}
        \centering
        \includegraphics[trim=2.00cm 0cm 0.5cm 0cm, clip, width=\linewidth]{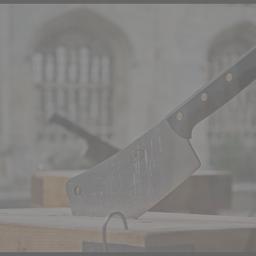}
        contrast\\ \phantom{original}
    \end{minipage}%
    \hfill
        \begin{minipage}[b]{0.16\columnwidth}
        \centering
        \includegraphics[trim=2.00cm 0cm 0.5cm 0cm, clip, width=\linewidth]{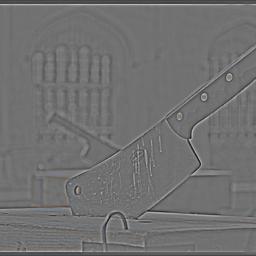}
        high pass filtering%
    \end{minipage}%
    \hfill 
    \begin{minipage}[b]{0.16\columnwidth}
        \centering
        \includegraphics[trim=2.00cm 0cm 0.5cm 0cm, clip, width=\linewidth]{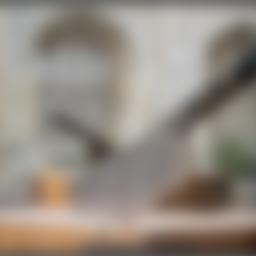}
        low pass filtering%
    \end{minipage}%
    \hfill
    \begin{minipage}[b]{0.16\columnwidth}
        \centering
        \includegraphics[trim=2.00cm 0cm 0.5cm 0cm, clip, width=\linewidth]{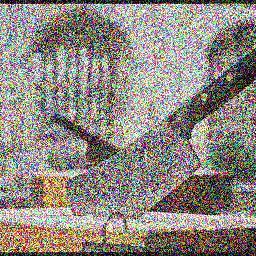}
        Gaussian noise%
    \end{minipage}%
    \hfill
    \begin{minipage}[b]{0.16\columnwidth}
        \centering
        \includegraphics[trim=2.00cm 0cm 0.5cm 0cm, clip, width=\linewidth]{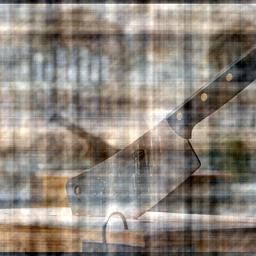}
        phase\\ noise
    \end{minipage}

    \caption{Simple corruptions of an ImageNet image, from left to right: original; contrast level $0.2$; high-pass difference between original and Gaussian blurring with standard deviation 1.5; low-pass Gaussian blurring with standard deviation $8$; Gaussian noise uniformly sampled from $[0,0.6]$; 90 degree phase noise.}
    \label{fig:distorted_knife}
\end{figure}

\paragraph{Cue-decomposition shape biases}

Based on classification accuracies $Q_S$ and $Q_T$ on the EED-transformed and the Voronoi-shuffled images, we computed the cue-decomposition shape bias $S_\mathrm{cd}$ for a number of models. Of those, the results for $20$ models are provided left in \cref{tab:classMetricsAndRob}.
In accordance to the literature, we observe texture biases for CNNs while transformers are less texture biased and almost balanced between their biases w.r.t.\ both cues. VLMs, however, on average show a slight tendency being biased towards shape. 
Comparing the cue-decomposition shape bias with the cue-conflict shape bias \cite{geirhos2018imagenettrained} (cf.\ sixth column of \cref{tab:classMetricsAndRob}), we see that both metrics are strongly correlated. Indeed, the Spearman rank correlation \cite{Spearman1904Proof} of both metrics is about $90.5\%$. 
Additionally, we also show cue-decomposition robustness results which are visually also relatively well-correlated with the two shape-bias metrics. Among each architecture type, there are some models for which relatively high robustness is indicated. We evaluated the quality of the robustness estimates provided by our cue-decomposition robustness in \cref{tab:classMetricsAndRob}. For the sake of comparison, we chose the same image corruptions with varying intensity as used in \cite{geirhos2018imagenettrained}, see also \cref{fig:distorted_knife} for a visual impression. Each relative robustness column of \cref{tab:classMetricsAndRob} shows the classification accuracy that remains after a given input corruption. Additionally, the values reported in the table are averages of the accuracies over multiple corruption intensities. 
Apparently, the model EVA02 large achieves the top rank in terms of mean relative robustness as well as cue-decomposition robustness and the right-most two columns are visually strongly correlated. 
Also here, we computed Spearman rank correlations between the mean robustness and the cue-conflict shape bias as well as mean robustness and the cue-decomposition shape bias. While the former rank correlation reaches a value of $79.3\%$, our method (the latter) achieves a rank correlation of $95.1\%$, clearly demonstrating the practical usefulness of our cue-decomposition evaluation procedure.

\paragraph{Ablation study of cue-decomposition and cue-conflict candidate schemes}

\begin{figure}[tb]
    \centering
    \begin{minipage}[b]{0.16\columnwidth}
        \includegraphics[trim=0.25cm 0cm 2.25cm 0cm, clip, width=\linewidth]{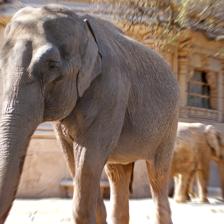}
    \end{minipage}%
    \hfill%
    \begin{minipage}[b]{0.16\columnwidth}
        \includegraphics[trim=0.25cm 0cm 2.25cm 0cm, clip, width=\linewidth]{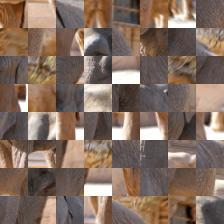}
    \end{minipage}%
    \hfill%
    \begin{minipage}[b]{0.16\columnwidth}
        \includegraphics[trim=0.25cm 0cm 2.25cm 0cm, clip, width=\linewidth]{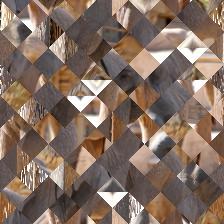}
    \end{minipage}%
    \hfill%
    \begin{minipage}[b]{0.16\columnwidth}
        \includegraphics[trim=0.25cm 0cm 2.25cm 0cm, clip, width=\linewidth]{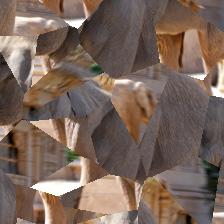}
    \end{minipage}%
    \hfill%
    \begin{minipage}[b]{0.16\columnwidth}
        \includegraphics[trim=0.25cm 0cm 2.25cm 0cm, clip, width=\linewidth]{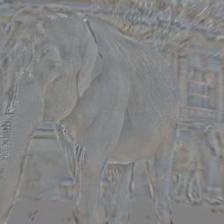}
    \end{minipage}%
    \hfill%
    \begin{minipage}[b]{0.16\columnwidth}
        \includegraphics[trim=0.25cm 0cm 2.25cm 0cm, clip, width=\linewidth]{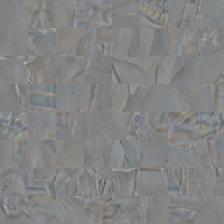}
    \end{minipage}
    \caption{Studied AI-free texture cue extraction candidates of an ImageNet image, from left to right: original; patch shuffling; diamond shuffling; Voronoi shuffling; difference of original and EED; the former patch shuffled.}
    \label{fig:texture_decomp_cand}
\end{figure}

The cue-decomposition evaluation procedure proposed in this work, while motivated by the visual disentanglement of image cues, is the result of a broad and thorough numerical study on which texture extraction scheme best complements the EED method for shape extraction. The shape extraction capabilities of EED itself have already been studied in \cite{heinert2024reducing}. Specifically, we looked into different shuffling methods as well as into a difference image of the original image and EED image to extract the texture cue.
Hereby, we consider regular patch shuffling (short `patch'), proposed as a texture indicator in \cite{zhang2019interpretingadversarially,mummadi2021doseenhanced} as well as shuffling diamond patches.
For patch-shuffling, the image is decomposed along a regular grid into quadratic patches which are randomly shuffled.
We decided to use a decomposition into \(8 \times 8\) patches to account for the trade-off between destroying most of the shape information, while preserving most of the texture information.
We additionally propose diamond shuffling, to apply a regular grid shuffling that allows to preserve all pixels during the shuffling but follows a different type of grid than that used by CNNs or the input patches of vision transformers.
The diamond shuffling (short: `diamond') follows the same concept but the grid is rotated by \(45\) degrees and wraps around the image borders.
We choose the diamond construction parameters such that their size is comparable to the \(8 \times 8\) axis-aligned patch grid.
Considering texture and shape as complementary cues and EED as shape extraction method, taking the pixel-wise difference image between the original image and the EED image leads to an additional texture candidate, which we call `tex.EED' in the tables.

\begin{table}
    \centering
    \scalebox{0.85}{\begin{tabular}{llrr|rr}
\toprule
\multicolumn{2}{c}{Dataset: \bfseries ImageNet} & \multicolumn{2}{c}{\bfseries Rank Corr. to} & \multicolumn{2}{c}{\bfseries Mean Tex.} \\
 &  & \bfseries Rel. & \bfseries C.-Conf. & \multicolumn{2}{c}{\bfseries Quality} \\
 &  & \bfseries Rob. & \bfseries Shape & \bfseries \(\boldsymbol{t}\) & \bfseries \(\boldsymbol{\overline{\mathrm{TP}}_T}\) \\
Metric & Data & \bfseries Mean & \bfseries Bias &  &  \\
\midrule
C.-Conf. & (Geirhos et al.\ \cite{geirhos2018imagenettrained}) & {\cellcolor[HTML]{C2C2FF}} \color[HTML]{000000} \color{black} 0.793 & {\cellcolor[HTML]{B2B2FF}} \color[HTML]{000000} \color{black} 1.000 & {\cellcolor[HTML]{FFFFFF}} \color[HTML]{000000} \color{black} - & {\cellcolor[HTML]{CFCFFF}} \color[HTML]{000000} \color{black} 0.491 \\
\cline{1-2} \cline{2-2}
Accuracy & EED & {\cellcolor[HTML]{BBBBFF}} \color[HTML]{000000} \color{black} 0.887 & {\cellcolor[HTML]{B8B8FF}} \color[HTML]{000000} \color{black} 0.925 & {\cellcolor[HTML]{FFFFFF}} \color[HTML]{000000} \color{black} - & {\cellcolor[HTML]{FFFFFF}} \color[HTML]{000000} \color{black} - \\
\cline{1-2} \cline{2-2}
\(R_\mathrm{cd}\) & EED, Voronoi & \bfseries {\cellcolor[HTML]{B6B6FF}} \color[HTML]{000000} \color{black} 0.951 & {\cellcolor[HTML]{BFBFFF}} \color[HTML]{000000} \color{black} 0.830 & \bfseries {\cellcolor[HTML]{FFCC80}} \color[HTML]{000000} \color{black} 0.854 & {\cellcolor[HTML]{FFFFFF}} \color[HTML]{000000} \color{black} - \\
\cline{1-2} \cline{2-2}
\multirow[c]{5}{*}{\(S_\mathrm{cd}\)} & EED, Voronoi & {\cellcolor[HTML]{C9C9FF}} \color[HTML]{000000} \color{black} 0.707 & {\cellcolor[HTML]{BABAFF}} \color[HTML]{000000} \color{black} 0.905 & \bfseries {\cellcolor[HTML]{FFCC80}} \color[HTML]{000000} \color{black} 0.854 & {\cellcolor[HTML]{FFFFFF}} \color[HTML]{000000} \color{black} - \\
\cline{2-2}
 & EED, tex.EED & {\cellcolor[HTML]{D4D4FF}} \color[HTML]{000000} \color{black} 0.556 & {\cellcolor[HTML]{BFBFFF}} \color[HTML]{000000} \color{black} 0.835 & {\cellcolor[HTML]{FFD390}} \color[HTML]{000000} \color{black} 0.758 & {\cellcolor[HTML]{FFFFFF}} \color[HTML]{000000} \color{black} - \\
\cline{2-2}
 & EED, tex.EEDpat. & {\cellcolor[HTML]{EFEFFF}} \color[HTML]{000000} \color{black} 0.206 & {\cellcolor[HTML]{D9D9FF}} \color[HTML]{000000} \color{black} 0.497 & {\cellcolor[HTML]{FFFFFF}} \color[HTML]{000000} \color{black} 0.145 & {\cellcolor[HTML]{FFFFFF}} \color[HTML]{000000} \color{black} - \\
\cline{2-2}
 & EED, patch & {\cellcolor[HTML]{FFF3F3}} \color[HTML]{000000} \color{black} -0.113 & {\cellcolor[HTML]{EAEAFF}} \color[HTML]{000000} \color{black} 0.268 & {\cellcolor[HTML]{FFD89D}} \color[HTML]{000000} \color{black} 0.688 & {\cellcolor[HTML]{FFFFFF}} \color[HTML]{000000} \color{black} - \\
\cline{2-2}
 & EED, diamond & {\cellcolor[HTML]{FFCACA}} \color[HTML]{000000} \color{black} -0.517 & {\cellcolor[HTML]{FFE1E1}} \color[HTML]{000000} \color{black} -0.296 & {\cellcolor[HTML]{FFF1DC}} \color[HTML]{000000} \color{black} 0.343 & {\cellcolor[HTML]{FFFFFF}} \color[HTML]{000000} \color{black} - \\
\cline{1-2} \cline{2-2}
\multirow[c]{10}{*}{\shortstack{other\\ C.-Conf.}} & EED, tex.EEDpat. & {\cellcolor[HTML]{DBDBFF}} \color[HTML]{000000} \color{black} 0.463 & {\cellcolor[HTML]{DBDBFF}} \color[HTML]{000000} \color{black} 0.470 & {\cellcolor[HTML]{FFFFFF}} \color[HTML]{000000} \color{black} - & {\cellcolor[HTML]{FFFFFF}} \color[HTML]{000000} \color{black} 0.126 \\
\cline{2-2}
 & EED(0.5), Voronoi(0.5) & {\cellcolor[HTML]{E7E7FF}} \color[HTML]{000000} \color{black} 0.320 & {\cellcolor[HTML]{DEDEFF}} \color[HTML]{000000} \color{black} 0.425 & {\cellcolor[HTML]{FFFFFF}} \color[HTML]{000000} \color{black} - & {\cellcolor[HTML]{CACAFF}} \color[HTML]{000000} \color{black} 0.523 \\
\cline{2-2}
 & EED, tex.EEDpat.(2) & {\cellcolor[HTML]{E8E8FF}} \color[HTML]{000000} \color{black} 0.292 & {\cellcolor[HTML]{E6E6FF}} \color[HTML]{000000} \color{black} 0.332 & {\cellcolor[HTML]{FFFFFF}} \color[HTML]{000000} \color{black} - & {\cellcolor[HTML]{F3F3FF}} \color[HTML]{000000} \color{black} 0.218 \\
\cline{2-2}
 & EED(0.7), Voronoi(0.3) & {\cellcolor[HTML]{F1F1FF}} \color[HTML]{000000} \color{black} 0.186 & {\cellcolor[HTML]{DADAFF}} \color[HTML]{000000} \color{black} 0.488 & {\cellcolor[HTML]{FFFFFF}} \color[HTML]{000000} \color{black} - & {\cellcolor[HTML]{EAEAFF}} \color[HTML]{000000} \color{black} 0.285 \\
\cline{2-2}
 & EED(0.7), patch(0.3) & {\cellcolor[HTML]{F6F6FF}} \color[HTML]{000000} \color{black} 0.114 & {\cellcolor[HTML]{F0F0FF}} \color[HTML]{000000} \color{black} 0.203 & {\cellcolor[HTML]{FFFFFF}} \color[HTML]{000000} \color{black} - & {\cellcolor[HTML]{FFFFFF}} \color[HTML]{000000} \color{black} 0.123 \\
\cline{2-2}
 & EED, tex.EED & {\cellcolor[HTML]{F7F7FF}} \color[HTML]{000000} \color{black} 0.106 & {\cellcolor[HTML]{F4F4FF}} \color[HTML]{000000} \color{black} 0.133 & {\cellcolor[HTML]{FFFFFF}} \color[HTML]{000000} \color{black} - & {\cellcolor[HTML]{D4D4FF}} \color[HTML]{000000} \color{black} 0.448 \\
\cline{2-2}
 & EED(0.3), Voronoi(0.7) & {\cellcolor[HTML]{FAFAFF}} \color[HTML]{000000} \color{black} 0.066 & {\cellcolor[HTML]{F3F3FF}} \color[HTML]{000000} \color{black} 0.150 & {\cellcolor[HTML]{FFFFFF}} \color[HTML]{000000} \color{black} - & \bfseries {\cellcolor[HTML]{B2B2FF}} \color[HTML]{000000} \color{black} 0.705 \\
\cline{2-2}
 & EED(0.8), tex.EEDpat.(3) & {\cellcolor[HTML]{FFFFFF}} \color[HTML]{000000} \color{black} 0.001 & {\cellcolor[HTML]{F2F2FF}} \color[HTML]{000000} \color{black} 0.169 & {\cellcolor[HTML]{FFFFFF}} \color[HTML]{000000} \color{black} - & {\cellcolor[HTML]{EAEAFF}} \color[HTML]{000000} \color{black} 0.286 \\
\cline{2-2}
 & EED(0.5), patch(0.5) & {\cellcolor[HTML]{FFEFEF}} \color[HTML]{000000} \color{black} -0.152 & {\cellcolor[HTML]{FFF2F2}} \color[HTML]{000000} \color{black} -0.127 & {\cellcolor[HTML]{FFFFFF}} \color[HTML]{000000} \color{black} - & {\cellcolor[HTML]{E9E9FF}} \color[HTML]{000000} \color{black} 0.292 \\
\cline{2-2}
 & EED(0.3), patch(0.7) & {\cellcolor[HTML]{FFD3D3}} \color[HTML]{000000} \color{black} -0.435 & {\cellcolor[HTML]{FFD5D5}} \color[HTML]{000000} \color{black} -0.410 & {\cellcolor[HTML]{FFFFFF}} \color[HTML]{000000} \color{black} - & {\cellcolor[HTML]{D3D3FF}} \color[HTML]{000000} \color{black} 0.460 \\
\cline{1-2} \cline{2-2}
\bottomrule
\end{tabular}
}
    \caption{(left) Rank correlations for the classification task of different cue-conflict (C.-Conf.) and cue-decomposition \(S_\mathrm{cd}\) shape bias metrics candidates as well as the robustness \(R_\mathrm{cd}\) metric to relative corruption robustness and to  cue-conflict shape bias  \cite{geirhos2018imagenettrained}.
    (right) The influence of texture is measured by the mean texture performance for cue-decomposition candidates by \cref{eq:s_and_t} and for cue-conflict candidates by $\overline{\mathrm{TP}}_T = \frac{1}{|\mathcal{H}|} \sum_{h \in \mathcal{H}} \mathrm{TP}_T(h) $ for the model set $\mathcal{H}$.
    Additionally, correlations of accuracies on EED images and cue-conflict shape bias included.
    Additional detailed results are reported in the appendix.}
    \label{tab:classCorr}
\end{table}

\begin{figure}[tb]
    \centering
    \begin{minipage}[b]{0.24\columnwidth}
        \includegraphics[width=\linewidth]{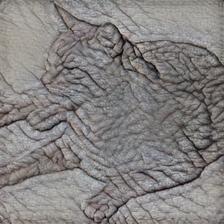}
    \end{minipage}
    \begin{minipage}[b]{0.24\columnwidth}
        \includegraphics[width=\linewidth]{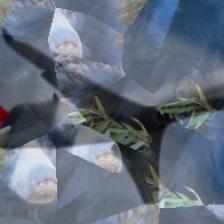}
    \end{minipage}
    \begin{minipage}[b]{0.24\columnwidth}
        \includegraphics[width=\linewidth]{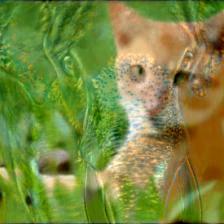}
    \end{minipage}
    \begin{minipage}[b]{0.24\columnwidth}
        \includegraphics[width=\linewidth]{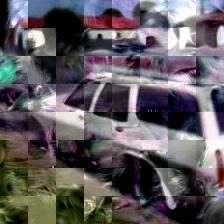}
    \end{minipage}
    \caption{Examples for studied cue-conflict candidates for ImageNet, from left to right: Style transfer of a cat with an elephant texture; blend of an EED airplane image with a Voronoi-shuffled image of a bear; sum of an EED cat image with the difference of an original image of an elephant and its EED version; sum of an EED image of a car ($\gamma_S = 1$) with the patch-shuffled difference of an original dog image and its EED version with $\gamma_T = 2$.}
    \label{fig:texture_candidates}
    \hspace{5pt}
\end{figure}

\begin{table*}[t]
    \centering
    \scalebox{0.85}{\begin{tabular}{llrrrrr|rrrrr}
\toprule
 &  & \multicolumn{5}{c}{\bfseries Cityscapes} & \multicolumn{5}{c}{\bfseries ADE20k} \\
 &  & \multicolumn{3}{c}{\bfseries mIoU} & \multicolumn{2}{c}{\bfseries C.-Dec.} & \multicolumn{3}{c}{\bfseries mIoU} & \multicolumn{2}{c}{\bfseries C.-Dec.} \\
 &  & \bfseries \(\boldsymbol{Q_O}\) & \bfseries \(\boldsymbol{Q_S}\) & \bfseries \(\boldsymbol{Q_T}\) & \bfseries \(\boldsymbol{S_\mathrm{cd}}\) & \bfseries \(\boldsymbol{R_\mathrm{cd}}\) & \bfseries \(\boldsymbol{Q_O}\) & \bfseries \(\boldsymbol{Q_S}\) & \bfseries \(\boldsymbol{Q_T}\) & \bfseries \(\boldsymbol{S_\mathrm{cd}}\) & \bfseries \(\boldsymbol{R_\mathrm{cd}}\) \\
Arch. & Model &  &  &  &  &  &  &  &  &  &  \\
\midrule
\multirow[c]{4}{*}{CNN} & Mask2Former RN101 \cite{cheng2021mask2former} & {\cellcolor[HTML]{FFDDAA}} \color[HTML]{000000} \color{black} 0.808 & {\cellcolor[HTML]{EDEDFF}} \color[HTML]{000000} \color{black} 0.211 & {\cellcolor[HTML]{FFF1DB}} \color[HTML]{000000} \color{black} 0.321 & {\cellcolor[HTML]{E2E2FF}} \color[HTML]{000000} \color{black} 0.480 & {\cellcolor[HTML]{FFF5E7}} \color[HTML]{000000} \color{black} 0.329 & {\cellcolor[HTML]{D5D5FF}} \color[HTML]{000000} \color{black} 0.467 & {\cellcolor[HTML]{FFF1DC}} \color[HTML]{000000} \color{black} 0.207 & {\cellcolor[HTML]{E4E4FF}} \color[HTML]{000000} \color{black} 0.100 & {\cellcolor[HTML]{FFE7C3}} \color[HTML]{000000} \color{black} 0.515 & {\cellcolor[HTML]{FFFFFF}} \color[HTML]{000000} \color{black} 0.329 \\
\cline{2-2}
 & FCN RN101 \cite{shelhamer2016fullyconvolutionalnetworkssemantic} & {\cellcolor[HTML]{FFFFFF}} \color[HTML]{000000} \color{black} 0.752 & {\cellcolor[HTML]{EFEFFF}} \color[HTML]{000000} \color{black} 0.205 & {\cellcolor[HTML]{FFDAA3}} \color[HTML]{000000} \color{black} 0.381 & {\cellcolor[HTML]{EFEFFF}} \color[HTML]{000000} \color{black} 0.430 & {\cellcolor[HTML]{FFE6C0}} \color[HTML]{000000} \color{black} 0.390 & {\cellcolor[HTML]{FEFEFF}} \color[HTML]{000000} \color{black} 0.390 & {\cellcolor[HTML]{FFFFFF}} \color[HTML]{000000} \color{black} 0.169 & {\cellcolor[HTML]{E9E9FF}} \color[HTML]{000000} \color{black} 0.094 & {\cellcolor[HTML]{FFF3E1}} \color[HTML]{000000} \color{black} 0.481 & {\cellcolor[HTML]{FAFAFF}} \color[HTML]{000000} \color{black} 0.337 \\
\cline{2-2}
 & DeepLabV3Plus RN101 \cite{chen2018encoderdecoderatrousseparableconvolution} & {\cellcolor[HTML]{FFE1B3}} \color[HTML]{000000} \color{black} 0.802 & {\cellcolor[HTML]{F7F7FF}} \color[HTML]{000000} \color{black} 0.179 & {\cellcolor[HTML]{FFE7C3}} \color[HTML]{000000} \color{black} 0.346 & {\cellcolor[HTML]{F2F2FF}} \color[HTML]{000000} \color{black} 0.421 & {\cellcolor[HTML]{FFF6E8}} \color[HTML]{000000} \color{black} 0.328 & {\cellcolor[HTML]{E0E0FF}} \color[HTML]{000000} \color{black} 0.446 & {\cellcolor[HTML]{FFF4E4}} \color[HTML]{000000} \color{black} 0.198 & {\cellcolor[HTML]{E0E0FF}} \color[HTML]{000000} \color{black} 0.105 & {\cellcolor[HTML]{FFEFD7}} \color[HTML]{000000} \color{black} 0.492 & {\cellcolor[HTML]{F8F8FF}} \color[HTML]{000000} \color{black} 0.340 \\
\cline{2-2}
 & PSPNet RN50 \cite{zhao2017pyramidsceneparsingnetwork} & {\cellcolor[HTML]{FFEFD6}} \color[HTML]{000000} \color{black} 0.779 & {\cellcolor[HTML]{FFFFFF}} \color[HTML]{000000} \color{black} 0.154 & {\cellcolor[HTML]{FFE0B0}} \color[HTML]{000000} \color{black} 0.366 & {\cellcolor[HTML]{FFFFFF}} \color[HTML]{000000} \color{black} 0.371 & {\cellcolor[HTML]{FFF4E4}} \color[HTML]{000000} \color{black} 0.334 & {\cellcolor[HTML]{F4F4FF}} \color[HTML]{000000} \color{black} 0.410 & {\cellcolor[HTML]{FFF8EE}} \color[HTML]{000000} \color{black} 0.188 & {\cellcolor[HTML]{D3D3FF}} \color[HTML]{000000} \color{black} 0.119 & {\cellcolor[HTML]{FFFFFF}} \color[HTML]{000000} \color{black} 0.447 & {\cellcolor[HTML]{E4E4FF}} \color[HTML]{000000} \color{black} 0.374 \\
\cline{1-2} \cline{2-2}
\multirow[c]{3}{*}{\shortstack{Vision\\ Transf.}} & SETR ViT-L\_mla \cite{zheng2021rethinkingsemanticsegmentationsequencetosequence} & {\cellcolor[HTML]{FFF9F0}} \color[HTML]{000000} \color{black} 0.762 & {\cellcolor[HTML]{B4B4FF}} \color[HTML]{000000} \color{black} 0.391 & {\cellcolor[HTML]{FFFCF7}} \color[HTML]{000000} \color{black} 0.291 & \bfseries {\cellcolor[HTML]{B2B2FF}} \color[HTML]{000000} \color{black} 0.653 & {\cellcolor[HTML]{FFD79A}} \color[HTML]{000000} \color{black} 0.448 & {\cellcolor[HTML]{D7D7FF}} \color[HTML]{000000} \color{black} 0.463 & {\cellcolor[HTML]{FFD089}} \color[HTML]{000000} \color{black} 0.295 & {\cellcolor[HTML]{CCCCFF}} \color[HTML]{000000} \color{black} 0.127 & {\cellcolor[HTML]{FFDCA8}} \color[HTML]{000000} \color{black} 0.544 & \bfseries {\cellcolor[HTML]{B2B2FF}} \color[HTML]{000000} \color{black} 0.456 \\
\cline{2-2}
 & SegFormer MiT-b4 \cite{xie2021segformersimpleefficientdesign} & {\cellcolor[HTML]{FFD698}} \color[HTML]{000000} \color{black} 0.819 & {\cellcolor[HTML]{B4B4FF}} \color[HTML]{000000} \color{black} 0.392 & {\cellcolor[HTML]{FFD391}} \color[HTML]{000000} \color{black} 0.400 & {\cellcolor[HTML]{C7C7FF}} \color[HTML]{000000} \color{black} 0.579 & {\cellcolor[HTML]{FFCD82}} \color[HTML]{000000} \color{black} 0.483 & {\cellcolor[HTML]{CDCDFF}} \color[HTML]{000000} \color{black} 0.480 & {\cellcolor[HTML]{FFE1B4}} \color[HTML]{000000} \color{black} 0.249 & {\cellcolor[HTML]{D5D5FF}} \color[HTML]{000000} \color{black} 0.117 & {\cellcolor[HTML]{FFE4BC}} \color[HTML]{000000} \color{black} 0.523 & {\cellcolor[HTML]{E0E0FF}} \color[HTML]{000000} \color{black} 0.381 \\
\cline{2-2}
 & Mask2Former Swin-B \cite{cheng2021mask2former} & \bfseries {\cellcolor[HTML]{FFCC80}} \color[HTML]{000000} \color{black} 0.835 & \bfseries {\cellcolor[HTML]{B2B2FF}} \color[HTML]{000000} \color{black} 0.397 & \bfseries {\cellcolor[HTML]{FFCC80}} \color[HTML]{000000} \color{black} 0.419 & {\cellcolor[HTML]{C9C9FF}} \color[HTML]{000000} \color{black} 0.571 & \bfseries {\cellcolor[HTML]{FFCC80}} \color[HTML]{000000} \color{black} 0.489 & \bfseries {\cellcolor[HTML]{B2B2FF}} \color[HTML]{000000} \color{black} 0.530 & \bfseries {\cellcolor[HTML]{FFCC80}} \color[HTML]{000000} \color{black} 0.306 & \bfseries {\cellcolor[HTML]{B2B2FF}} \color[HTML]{000000} \color{black} 0.156 & {\cellcolor[HTML]{FFEBCE}} \color[HTML]{000000} \color{black} 0.502 & {\cellcolor[HTML]{BFBFFF}} \color[HTML]{000000} \color{black} 0.435 \\
\cline{1-2} \cline{2-2}
\multirow[c]{2}{*}{other} & OCRNet hr18 \cite{yuan2021segmentationtransformerobjectcontextualrepresentations} & {\cellcolor[HTML]{FFEFD6}} \color[HTML]{000000} \color{black} 0.779 & {\cellcolor[HTML]{FAFAFF}} \color[HTML]{000000} \color{black} 0.172 & {\cellcolor[HTML]{FFFFFF}} \color[HTML]{000000} \color{black} 0.283 & {\cellcolor[HTML]{E7E7FF}} \color[HTML]{000000} \color{black} 0.461 & {\cellcolor[HTML]{FFFFFF}} \color[HTML]{000000} \color{black} 0.292 & {\cellcolor[HTML]{FFFFFF}} \color[HTML]{000000} \color{black} 0.389 & {\cellcolor[HTML]{FFF6E8}} \color[HTML]{000000} \color{black} 0.195 & {\cellcolor[HTML]{FFFFFF}} \color[HTML]{000000} \color{black} 0.070 & \bfseries {\cellcolor[HTML]{FFCC80}} \color[HTML]{000000} \color{black} 0.591 & {\cellcolor[HTML]{F8F8FF}} \color[HTML]{000000} \color{black} 0.340 \\
\cline{2-2}
 & DNL RN101 \cite{yin2020disentangled} & {\cellcolor[HTML]{FFECD0}} \color[HTML]{000000} \color{black} 0.783 & {\cellcolor[HTML]{EEEEFF}} \color[HTML]{000000} \color{black} 0.209 & {\cellcolor[HTML]{FFDBA6}} \color[HTML]{000000} \color{black} 0.377 & {\cellcolor[HTML]{EDEDFF}} \color[HTML]{000000} \color{black} 0.437 & {\cellcolor[HTML]{FFEACA}} \color[HTML]{000000} \color{black} 0.374 & {\cellcolor[HTML]{EAEAFF}} \color[HTML]{000000} \color{black} 0.428 & {\cellcolor[HTML]{FFEFD6}} \color[HTML]{000000} \color{black} 0.213 & {\cellcolor[HTML]{D6D6FF}} \color[HTML]{000000} \color{black} 0.116 & {\cellcolor[HTML]{FFF1DD}} \color[HTML]{000000} \color{black} 0.486 & {\cellcolor[HTML]{DDDDFF}} \color[HTML]{000000} \color{black} 0.385 \\
\cline{1-2} \cline{2-2}
\bottomrule
\end{tabular}
}
    \caption{
    Cue-decomposition shape biases and robustness metrics of semantic segmentation models including their metric components as defined in \cref{eq:s_cd} and \cref{eq:r_cd}.
    On the left for the Cityscapes dataset and on the right for ADE20k.
    The full table with $23$ models can be found in the appendix.
    }
    \label{tab:semsegMetrics}
\end{table*}

In addition, we considered a patch-shuffled version (short `tex.EEDpat') of the tex.EED images with \(8 \times 8\) patches to further distort shape information, as from visual inspection we could still recognize the shape of most of the depicted objects.
\Cref{fig:texture_decomp_cand} shows examples generated by the respective methods.
Besides cue-decomposition, we also introduce and study different cue-conflict methods. By overlaying EED images with texture cue images, we generate images with conflicting cues. Therefore, we combine an EED image of class $C_i$ with a texture image of class $C_j$, $i \neq j$, by either blending an EED image with a texture image (e.g.\ `Voronoi, diamond, patch, tex.EED, tex.EEDpat.')
or by summing the EED image and the texture image pixel-wise up and clipping values exceeding the maximal value of $255$.
In both settings, before blending or summing up, the texture and the shape cue image are weighted with $\gamma_S \in \mathbb{R}_{\geq0}$ and $\gamma_T \in \mathbb{R}_{\geq0}$ for the shape and the texture cue, respectively. See \cref{fig:texture_candidates} for visual examples.

To evaluate these additional cue-conflict images we compute the cue-conflict shape biases according to \cref{eq:geihros}.
Our results shown in \cref{tab:classCorr} demonstrate that decomposing the image cues with the help of Voronoi shuffling and EED yields higher rank correlation to the cue-conflict shape bias metric of Geirhos et al.\ (cf.\ last column in \cref{tab:classCorr}) than the other cue-decomposition settings as well as the cue-conflict alternatives. Especially the diamond shuffling and cue-conflict of EED with highly weighted patch-shuffling leads to bad rank correlations. We found that classification based on diamond-shuffled images is a very hard task for all model architectures and therefore texture performance dropped even for networks where a distinct texture bias was already shown in literature. 
For the sake of completeness, we also compare against the cue-conflict shape bias metric and the performance on EED images alone (first and second row). Even though the Spearman rank correlations between EED accuracies and the cue-conflict metric is slightly higher than our proposed procedure ($S_\mathrm{cd}$ EED, Voronoi) and second highest when comparing rank correlations with respect to the relative corruption robustness mean, this metric ignores texture influence. 
In contrast, our cue-decomposition metric with EED and Voronoi-shuffled images ($S_\mathrm{cd}$ EED, Voronoi) takes shape as well as texture performance into account. 
Indeed, the third column of  \cref{tab:classCorr} signals that DNNs can extract information from Voronoi-shuffled data comparatively well.
In addition, when computing the robustness $R_\mathrm{cd}$ (\cref{eq:r_cd}) based on EED and Voronoi-shuffled images we see the highest rank correlation with $95.1\%$ w.r.t.\ the mean relative robustness, outperforming the cue-conflict shape bias metric of Geirhos et al.\ ($79.3\%$) and the EED accuracy metric ($88.7\%$).

\begin{table*}
    \centering
    \scalebox{0.83}{\begin{tabular}{llrrrrrrr|rrrrrrr}
\toprule
 &  & \multicolumn{7}{c}{\bfseries Cityscapes} & \multicolumn{7}{c}{\bfseries ADE20k} \\
 &  & \multicolumn{6}{c}{\bfseries Relative Robustness} & \bfseries C.-Dec. & \multicolumn{6}{c}{\bfseries Relative Robustness} & \bfseries C.-Dec. \\
 &  & \bfseries Cont. & \bfseries High & \bfseries Low & \bfseries Noise & \bfseries Phase & \bfseries Mean & \bfseries \(\boldsymbol{R_\mathrm{cd}}\) & \bfseries Cont. & \bfseries High & \bfseries Low & \bfseries Noise & \bfseries Phase & \bfseries Mean & \bfseries \(\boldsymbol{R_\mathrm{cd}}\) \\
Arch. & Model &  & \bfseries Pass & \bfseries Pass &  & \bfseries Noise &  &  &  & \bfseries Pass & \bfseries Pass &  & \bfseries Noise &  &  \\
\midrule
\multirow[c]{4}{*}{CNN} & DeepLabV3Plus RN101 \cite{chen2018encoderdecoderatrousseparableconvolution} & {\cellcolor[HTML]{FFF6E9}} \color[HTML]{000000} \color{black} 0.741 & {\cellcolor[HTML]{BEBEFF}} \color[HTML]{000000} \color{black} 0.337 & {\cellcolor[HTML]{FFFAF3}} \color[HTML]{000000} \color{black} 0.445 & {\cellcolor[HTML]{F4F4FF}} \color[HTML]{000000} \color{black} 0.373 & {\cellcolor[HTML]{FFF5E6}} \color[HTML]{000000} \color{black} 0.257 & {\cellcolor[HTML]{F2F2FF}} \color[HTML]{000000} \color{black} 0.431 & {\cellcolor[HTML]{FFF6E8}} \color[HTML]{000000} \color{black} 0.328 & {\cellcolor[HTML]{ECECFF}} \color[HTML]{000000} \color{black} 0.679 & {\cellcolor[HTML]{FFF6E8}} \color[HTML]{000000} \color{black} 0.162 & {\cellcolor[HTML]{F7F7FF}} \color[HTML]{000000} \color{black} 0.404 & {\cellcolor[HTML]{FFFBF6}} \color[HTML]{000000} \color{black} 0.534 & {\cellcolor[HTML]{FFFFFF}} \color[HTML]{000000} \color{black} 0.348 & {\cellcolor[HTML]{FFF8EE}} \color[HTML]{000000} \color{black} 0.425 & {\cellcolor[HTML]{F8F8FF}} \color[HTML]{000000} \color{black} 0.340 \\
\cline{2-2}
 & FCN RN101 \cite{shelhamer2016fullyconvolutionalnetworkssemantic} & {\cellcolor[HTML]{FFF3E2}} \color[HTML]{000000} \color{black} 0.752 & {\cellcolor[HTML]{D2D2FF}} \color[HTML]{000000} \color{black} 0.279 & {\cellcolor[HTML]{FFFAF2}} \color[HTML]{000000} \color{black} 0.446 & {\cellcolor[HTML]{F7F7FF}} \color[HTML]{000000} \color{black} 0.357 & {\cellcolor[HTML]{FFF2DF}} \color[HTML]{000000} \color{black} 0.265 & {\cellcolor[HTML]{F8F8FF}} \color[HTML]{000000} \color{black} 0.420 & {\cellcolor[HTML]{FFE6C0}} \color[HTML]{000000} \color{black} 0.390 & {\cellcolor[HTML]{FFFFFF}} \color[HTML]{000000} \color{black} 0.642 & {\cellcolor[HTML]{FFFFFF}} \color[HTML]{000000} \color{black} 0.129 & {\cellcolor[HTML]{FFFFFF}} \color[HTML]{000000} \color{black} 0.389 & {\cellcolor[HTML]{FFFFFF}} \color[HTML]{000000} \color{black} 0.514 & {\cellcolor[HTML]{FFFFFF}} \color[HTML]{000000} \color{black} 0.347 & {\cellcolor[HTML]{FFFFFF}} \color[HTML]{000000} \color{black} 0.404 & {\cellcolor[HTML]{FAFAFF}} \color[HTML]{000000} \color{black} 0.337 \\
\cline{2-2}
 & Mask2Former RN101 \cite{cheng2021mask2former} & {\cellcolor[HTML]{FFEDD3}} \color[HTML]{000000} \color{black} 0.775 & {\cellcolor[HTML]{E5E5FF}} \color[HTML]{000000} \color{black} 0.225 & {\cellcolor[HTML]{FFF1DC}} \color[HTML]{000000} \color{black} 0.488 & {\cellcolor[HTML]{FDFDFF}} \color[HTML]{000000} \color{black} 0.330 & {\cellcolor[HTML]{FFF6EA}} \color[HTML]{000000} \color{black} 0.253 & {\cellcolor[HTML]{FAFAFF}} \color[HTML]{000000} \color{black} 0.414 & {\cellcolor[HTML]{FFF5E7}} \color[HTML]{000000} \color{black} 0.329 & {\cellcolor[HTML]{E6E6FF}} \color[HTML]{000000} \color{black} 0.690 & {\cellcolor[HTML]{FFFBF6}} \color[HTML]{000000} \color{black} 0.142 & {\cellcolor[HTML]{E6E6FF}} \color[HTML]{000000} \color{black} 0.435 & {\cellcolor[HTML]{FFF7EC}} \color[HTML]{000000} \color{black} 0.555 & {\cellcolor[HTML]{FDFDFF}} \color[HTML]{000000} \color{black} 0.350 & {\cellcolor[HTML]{FFF5E6}} \color[HTML]{000000} \color{black} 0.434 & {\cellcolor[HTML]{FFFFFF}} \color[HTML]{000000} \color{black} 0.329 \\
\cline{2-2}
 & PSPNet RN50 \cite{zhao2017pyramidsceneparsingnetwork} & {\cellcolor[HTML]{FFFAF2}} \color[HTML]{000000} \color{black} 0.728 & {\cellcolor[HTML]{D0D0FF}} \color[HTML]{000000} \color{black} 0.285 & {\cellcolor[HTML]{FFFFFF}} \color[HTML]{000000} \color{black} 0.423 & {\cellcolor[HTML]{FAFAFF}} \color[HTML]{000000} \color{black} 0.345 & {\cellcolor[HTML]{FFFCF6}} \color[HTML]{000000} \color{black} 0.237 & {\cellcolor[HTML]{FFFFFF}} \color[HTML]{000000} \color{black} 0.403 & {\cellcolor[HTML]{FFF4E4}} \color[HTML]{000000} \color{black} 0.334 & {\cellcolor[HTML]{FCFCFF}} \color[HTML]{000000} \color{black} 0.647 & {\cellcolor[HTML]{FFFCF8}} \color[HTML]{000000} \color{black} 0.139 & {\cellcolor[HTML]{F7F7FF}} \color[HTML]{000000} \color{black} 0.403 & {\cellcolor[HTML]{FFFEFD}} \color[HTML]{000000} \color{black} 0.518 & {\cellcolor[HTML]{FFFFFF}} \color[HTML]{000000} \color{black} 0.347 & {\cellcolor[HTML]{FFFDFA}} \color[HTML]{000000} \color{black} 0.411 & {\cellcolor[HTML]{E4E4FF}} \color[HTML]{000000} \color{black} 0.374 \\
\cline{1-2} \cline{2-2}
\multirow[c]{3}{*}{\shortstack{Vision\\ Transf.}} & Mask2Former Swin-B \cite{cheng2021mask2former} & \bfseries {\cellcolor[HTML]{FFCC80}} \color[HTML]{000000} \color{black} 0.908 & {\cellcolor[HTML]{CDCDFF}} \color[HTML]{000000} \color{black} 0.292 & {\cellcolor[HTML]{FFD28F}} \color[HTML]{000000} \color{black} 0.627 & \bfseries {\cellcolor[HTML]{B2B2FF}} \color[HTML]{000000} \color{black} 0.666 & {\cellcolor[HTML]{FFD89E}} \color[HTML]{000000} \color{black} 0.344 & \bfseries {\cellcolor[HTML]{B2B2FF}} \color[HTML]{000000} \color{black} 0.567 & \bfseries {\cellcolor[HTML]{FFCC80}} \color[HTML]{000000} \color{black} 0.489 & {\cellcolor[HTML]{C1C1FF}} \color[HTML]{000000} \color{black} 0.761 & \bfseries {\cellcolor[HTML]{FFCC80}} \color[HTML]{000000} \color{black} 0.309 & {\cellcolor[HTML]{C1C1FF}} \color[HTML]{000000} \color{black} 0.504 & \bfseries {\cellcolor[HTML]{FFCC80}} \color[HTML]{000000} \color{black} 0.786 & \bfseries {\cellcolor[HTML]{B2B2FF}} \color[HTML]{000000} \color{black} 0.450 & \bfseries {\cellcolor[HTML]{FFCC80}} \color[HTML]{000000} \color{black} 0.562 & {\cellcolor[HTML]{BFBFFF}} \color[HTML]{000000} \color{black} 0.435 \\
\cline{2-2}
 & SegFormer MiT-b4 \cite{xie2021segformersimpleefficientdesign} & {\cellcolor[HTML]{FFCD82}} \color[HTML]{000000} \color{black} 0.905 & \bfseries {\cellcolor[HTML]{B2B2FF}} \color[HTML]{000000} \color{black} 0.368 & {\cellcolor[HTML]{FFE2B8}} \color[HTML]{000000} \color{black} 0.553 & {\cellcolor[HTML]{BEBEFF}} \color[HTML]{000000} \color{black} 0.611 & {\cellcolor[HTML]{FFCE85}} \color[HTML]{000000} \color{black} 0.373 & {\cellcolor[HTML]{B5B5FF}} \color[HTML]{000000} \color{black} 0.562 & {\cellcolor[HTML]{FFCD82}} \color[HTML]{000000} \color{black} 0.483 & \bfseries {\cellcolor[HTML]{B2B2FF}} \color[HTML]{000000} \color{black} 0.787 & {\cellcolor[HTML]{FFE4BC}} \color[HTML]{000000} \color{black} 0.224 & {\cellcolor[HTML]{D1D1FF}} \color[HTML]{000000} \color{black} 0.474 & {\cellcolor[HTML]{FFDBA4}} \color[HTML]{000000} \color{black} 0.708 & {\cellcolor[HTML]{C2C2FF}} \color[HTML]{000000} \color{black} 0.429 & {\cellcolor[HTML]{FFD89E}} \color[HTML]{000000} \color{black} 0.524 & {\cellcolor[HTML]{E0E0FF}} \color[HTML]{000000} \color{black} 0.381 \\
\cline{2-2}
 & SETR ViT-L\_mla \cite{zheng2021rethinkingsemanticsegmentationsequencetosequence} & {\cellcolor[HTML]{FFE5BF}} \color[HTML]{000000} \color{black} 0.808 & {\cellcolor[HTML]{FFFFFF}} \color[HTML]{000000} \color{black} 0.151 & \bfseries {\cellcolor[HTML]{FFCC80}} \color[HTML]{000000} \color{black} 0.656 & {\cellcolor[HTML]{B8B8FF}} \color[HTML]{000000} \color{black} 0.643 & \bfseries {\cellcolor[HTML]{FFCC80}} \color[HTML]{000000} \color{black} 0.380 & {\cellcolor[HTML]{C5C5FF}} \color[HTML]{000000} \color{black} 0.527 & {\cellcolor[HTML]{FFD79A}} \color[HTML]{000000} \color{black} 0.448 & {\cellcolor[HTML]{EEEEFF}} \color[HTML]{000000} \color{black} 0.673 & {\cellcolor[HTML]{FFEBCE}} \color[HTML]{000000} \color{black} 0.200 & \bfseries {\cellcolor[HTML]{B2B2FF}} \color[HTML]{000000} \color{black} 0.531 & {\cellcolor[HTML]{FFDBA4}} \color[HTML]{000000} \color{black} 0.707 & {\cellcolor[HTML]{BBBBFF}} \color[HTML]{000000} \color{black} 0.437 & {\cellcolor[HTML]{FFDDAA}} \color[HTML]{000000} \color{black} 0.510 & \bfseries {\cellcolor[HTML]{B2B2FF}} \color[HTML]{000000} \color{black} 0.456 \\
\cline{1-2} \cline{2-2}
\multirow[c]{2}{*}{other} & OCRNet hr18 \cite{yuan2021segmentationtransformerobjectcontextualrepresentations} & {\cellcolor[HTML]{FFFFFF}} \color[HTML]{000000} \color{black} 0.706 & {\cellcolor[HTML]{D8D8FF}} \color[HTML]{000000} \color{black} 0.261 & {\cellcolor[HTML]{FFFCF8}} \color[HTML]{000000} \color{black} 0.437 & {\cellcolor[HTML]{ECECFF}} \color[HTML]{000000} \color{black} 0.407 & {\cellcolor[HTML]{FFFDFA}} \color[HTML]{000000} \color{black} 0.233 & {\cellcolor[HTML]{FDFDFF}} \color[HTML]{000000} \color{black} 0.409 & {\cellcolor[HTML]{FFFFFF}} \color[HTML]{000000} \color{black} 0.292 & {\cellcolor[HTML]{EBEBFF}} \color[HTML]{000000} \color{black} 0.680 & {\cellcolor[HTML]{FFDFAF}} \color[HTML]{000000} \color{black} 0.242 & {\cellcolor[HTML]{EDEDFF}} \color[HTML]{000000} \color{black} 0.421 & {\cellcolor[HTML]{FFF4E4}} \color[HTML]{000000} \color{black} 0.571 & {\cellcolor[HTML]{FDFDFF}} \color[HTML]{000000} \color{black} 0.350 & {\cellcolor[HTML]{FFEFD8}} \color[HTML]{000000} \color{black} 0.453 & {\cellcolor[HTML]{F8F8FF}} \color[HTML]{000000} \color{black} 0.340 \\
\cline{2-2}
 & DNL RN101 \cite{yin2020disentangled} & {\cellcolor[HTML]{FFF1DC}} \color[HTML]{000000} \color{black} 0.761 & {\cellcolor[HTML]{CFCFFF}} \color[HTML]{000000} \color{black} 0.287 & {\cellcolor[HTML]{FFFAF2}} \color[HTML]{000000} \color{black} 0.447 & {\cellcolor[HTML]{FFFFFF}} \color[HTML]{000000} \color{black} 0.321 & {\cellcolor[HTML]{FFFFFF}} \color[HTML]{000000} \color{black} 0.227 & {\cellcolor[HTML]{FDFDFF}} \color[HTML]{000000} \color{black} 0.408 & {\cellcolor[HTML]{FFEACA}} \color[HTML]{000000} \color{black} 0.374 & {\cellcolor[HTML]{F4F4FF}} \color[HTML]{000000} \color{black} 0.663 & {\cellcolor[HTML]{FFFDFA}} \color[HTML]{000000} \color{black} 0.138 & {\cellcolor[HTML]{F6F6FF}} \color[HTML]{000000} \color{black} 0.405 & {\cellcolor[HTML]{FFFDFA}} \color[HTML]{000000} \color{black} 0.526 & {\cellcolor[HTML]{F9F9FF}} \color[HTML]{000000} \color{black} 0.356 & {\cellcolor[HTML]{FFFBF4}} \color[HTML]{000000} \color{black} 0.418 & {\cellcolor[HTML]{DDDDFF}} \color[HTML]{000000} \color{black} 0.385 \\
\cline{1-2} \cline{2-2}
\bottomrule
\end{tabular}
}
    \caption{
    Relative robustness calculated for different types of image corruptions like contrast changes (Cont.) and averaged over all corruption types in the last column (Mean).
    On the left for the Cityscapes dataset and on the right for ADE20k.
    The full table with $23$ models can be found in the appendix.
    }
    \label{tab:semsegRob}
\end{table*}

\begin{table}
    \centering
    \scalebox{0.85}{\begin{tabular}{llrrrrrr}
\toprule
\multicolumn{2}{c}{Dataset: \bfseries Cityscapes} & \multicolumn{6}{c}{\bfseries Rank Correlation to Relative Robustness} \\
 &  & \bfseries Cont. & \bfseries High & \bfseries Low & \bfseries Noise & \bfseries Phase & \bfseries Mean \\
Metric & Data &  & \bfseries Pass & \bfseries Pass &  & \bfseries Noise &  \\
\midrule
\multirow[c]{5}{*}{mIoU} & EED & \bfseries {\cellcolor[HTML]{BDBDFF}} \color[HTML]{000000} \color{black} 0.859 & {\cellcolor[HTML]{FFF2F2}} \color[HTML]{000000} \color{black} -0.133 & \bfseries {\cellcolor[HTML]{BABAFF}} \color[HTML]{000000} \color{black} 0.898 & {\cellcolor[HTML]{CBCBFF}} \color[HTML]{000000} \color{black} 0.676 & \bfseries {\cellcolor[HTML]{C0C0FF}} \color[HTML]{000000} \color{black} 0.813 & {\cellcolor[HTML]{C4C4FF}} \color[HTML]{000000} \color{black} 0.762 \\
\cline{2-2}
 & Vor. 32 & {\cellcolor[HTML]{CACAFF}} \color[HTML]{000000} \color{black} 0.681 & {\cellcolor[HTML]{E0E0FF}} \color[HTML]{000000} \color{black} 0.409 & {\cellcolor[HTML]{D1D1FF}} \color[HTML]{000000} \color{black} 0.598 & {\cellcolor[HTML]{DCDCFF}} \color[HTML]{000000} \color{black} 0.449 & {\cellcolor[HTML]{CECEFF}} \color[HTML]{000000} \color{black} 0.637 & {\cellcolor[HTML]{D6D6FF}} \color[HTML]{000000} \color{black} 0.529 \\
\cline{2-2}
 & Cityscapes & {\cellcolor[HTML]{D6D6FF}} \color[HTML]{000000} \color{black} 0.532 & \bfseries {\cellcolor[HTML]{D3D3FF}} \color[HTML]{000000} \color{black} 0.570 & {\cellcolor[HTML]{E6E6FF}} \color[HTML]{000000} \color{black} 0.330 & {\cellcolor[HTML]{E8E8FF}} \color[HTML]{000000} \color{black} 0.294 & {\cellcolor[HTML]{F9F9FF}} \color[HTML]{000000} \color{black} 0.082 & {\cellcolor[HTML]{D8D8FF}} \color[HTML]{000000} \color{black} 0.504 \\
\cline{2-2}
 & Vor. 64 & {\cellcolor[HTML]{D0D0FF}} \color[HTML]{000000} \color{black} 0.610 & {\cellcolor[HTML]{E4E4FF}} \color[HTML]{000000} \color{black} 0.346 & {\cellcolor[HTML]{D5D5FF}} \color[HTML]{000000} \color{black} 0.551 & {\cellcolor[HTML]{DCDCFF}} \color[HTML]{000000} \color{black} 0.446 & {\cellcolor[HTML]{CDCDFF}} \color[HTML]{000000} \color{black} 0.650 & {\cellcolor[HTML]{DCDCFF}} \color[HTML]{000000} \color{black} 0.461 \\
\cline{2-2}
 & Vor. 128 & {\cellcolor[HTML]{E2E2FF}} \color[HTML]{000000} \color{black} 0.381 & {\cellcolor[HTML]{E5E5FF}} \color[HTML]{000000} \color{black} 0.337 & {\cellcolor[HTML]{F3F3FF}} \color[HTML]{000000} \color{black} 0.155 & {\cellcolor[HTML]{F7F7FF}} \color[HTML]{000000} \color{black} 0.106 & {\cellcolor[HTML]{EFEFFF}} \color[HTML]{000000} \color{black} 0.207 & {\cellcolor[HTML]{F0F0FF}} \color[HTML]{000000} \color{black} 0.196 \\
\cline{1-2} \cline{2-2}
\multirow[c]{3}{*}{\(S_\mathrm{cd}\)} & EED, Vor. 128 & {\cellcolor[HTML]{C4C4FF}} \color[HTML]{000000} \color{black} 0.773 & {\cellcolor[HTML]{FFF1F1}} \color[HTML]{000000} \color{black} -0.137 & {\cellcolor[HTML]{BBBBFF}} \color[HTML]{000000} \color{black} 0.885 & \bfseries {\cellcolor[HTML]{C3C3FF}} \color[HTML]{000000} \color{black} 0.774 & {\cellcolor[HTML]{C2C2FF}} \color[HTML]{000000} \color{black} 0.800 & \bfseries {\cellcolor[HTML]{C1C1FF}} \color[HTML]{000000} \color{black} 0.809 \\
\cline{2-2}
 & EED, Vor. 64 & {\cellcolor[HTML]{C9C9FF}} \color[HTML]{000000} \color{black} 0.711 & {\cellcolor[HTML]{F3F3FF}} \color[HTML]{000000} \color{black} 0.162 & {\cellcolor[HTML]{C2C2FF}} \color[HTML]{000000} \color{black} 0.797 & {\cellcolor[HTML]{CECEFF}} \color[HTML]{000000} \color{black} 0.640 & {\cellcolor[HTML]{CBCBFF}} \color[HTML]{000000} \color{black} 0.676 & {\cellcolor[HTML]{CACAFF}} \color[HTML]{000000} \color{black} 0.686 \\
\cline{2-2}
 & EED, Vor. 32 & {\cellcolor[HTML]{C7C7FF}} \color[HTML]{000000} \color{black} 0.733 & {\cellcolor[HTML]{F5F5FF}} \color[HTML]{000000} \color{black} 0.125 & {\cellcolor[HTML]{C0C0FF}} \color[HTML]{000000} \color{black} 0.814 & {\cellcolor[HTML]{D0D0FF}} \color[HTML]{000000} \color{black} 0.613 & {\cellcolor[HTML]{C9C9FF}} \color[HTML]{000000} \color{black} 0.706 & {\cellcolor[HTML]{CCCCFF}} \color[HTML]{000000} \color{black} 0.672 \\
\cline{1-2} \cline{2-2}
\multirow[c]{3}{*}{\(R_\mathrm{cd}\)} & EED, Vor. 128 & {\cellcolor[HTML]{CBCBFF}} \color[HTML]{000000} \color{black} 0.678 & {\cellcolor[HTML]{FFEFEF}} \color[HTML]{000000} \color{black} -0.149 & {\cellcolor[HTML]{CBCBFF}} \color[HTML]{000000} \color{black} 0.677 & {\cellcolor[HTML]{D1D1FF}} \color[HTML]{000000} \color{black} 0.597 & {\cellcolor[HTML]{C3C3FF}} \color[HTML]{000000} \color{black} 0.775 & {\cellcolor[HTML]{D3D3FF}} \color[HTML]{000000} \color{black} 0.570 \\
\cline{2-2}
 & EED, Vor. 32 & {\cellcolor[HTML]{CECEFF}} \color[HTML]{000000} \color{black} 0.637 & {\cellcolor[HTML]{F1F1FF}} \color[HTML]{000000} \color{black} 0.186 & {\cellcolor[HTML]{CFCFFF}} \color[HTML]{000000} \color{black} 0.627 & {\cellcolor[HTML]{DDDDFF}} \color[HTML]{000000} \color{black} 0.444 & {\cellcolor[HTML]{C2C2FF}} \color[HTML]{000000} \color{black} 0.789 & {\cellcolor[HTML]{DBDBFF}} \color[HTML]{000000} \color{black} 0.473 \\
\cline{2-2}
 & EED, Vor. 64 & {\cellcolor[HTML]{D1D1FF}} \color[HTML]{000000} \color{black} 0.598 & {\cellcolor[HTML]{EDEDFF}} \color[HTML]{000000} \color{black} 0.230 & {\cellcolor[HTML]{D2D2FF}} \color[HTML]{000000} \color{black} 0.586 & {\cellcolor[HTML]{DCDCFF}} \color[HTML]{000000} \color{black} 0.461 & {\cellcolor[HTML]{C4C4FF}} \color[HTML]{000000} \color{black} 0.772 & {\cellcolor[HTML]{DBDBFF}} \color[HTML]{000000} \color{black} 0.466 \\
\cline{1-2} \cline{2-2}
\bottomrule
\end{tabular}
}
    \caption{
    Rank Correlation of different mIoU, cue-decomposition shape bias \(S_\mathrm{cd}\) and robustness \(R_\mathrm{cd}\) metrics to relative corruption robustness for semantic segmentation on Cityscapes.
    }
    \label{tab:semsegCityCorr}
\end{table}

\subsection{Semantic segmentation results}

In this section, we discuss our numerical results for cue-decomposition shape biases and robustness values on Cityscapes and ADE20K. This is the first extensive study on image-cue biases in semantic segmentation DNNs.

\paragraph{Cue-decomposition metrics}

\Cref{tab:semsegMetrics} provides our cue-decomposition metrics and its constituting prediction quality metrics. 
Across both datasets, we observe similar tendencies as in image classification. Vision transformers tend to be more shape biased than other architectures. In particular, vision transformers are also capable of extracting way more information from the EED-transformed data as other architecture types (cf.\ $Q_S$). Although less pronounced, this holds true for the Voronoi-shuffled data as well (cf.\ $Q_T$). This already indicates some robustness to the absence of image cues.
Indeed, \cref{tab:semsegRob} shows that vision transformers are more robust also w.r.t.\ image corruptions than other architecture types. Visually, there appears to be some correlation of mean relative robustness and $R_\mathrm{cd}$. It is also interesting to see that different DNNs of the same architecture type can respond quite differently to a given image corruption, e.g.\ transformers to high pass filtering.

\paragraph{Robustness to image corruptions}

We again consider the estimation of robustness w.r.t.\ images corruptions. \Cref{tab:semsegCityCorr,tab:semsegade20kCorr} show a behavior different from the image classification case. 
Our cue-decomposition robustness  
is now outperformed by the cue-decomposition shape bias in terms of rank correlation consistently across both datasets. Hence, the visual correlation of $R_\mathrm{cd}$ and mean relative robustness turns out to be less strong in terms of rank correlation.
Furthermore, two observations are interesting. First, the mIoU on the original dataset is a strong indicator of robustness for semantic segmentation DNNs. Generally, the stronger the mIoU performance, the more robust is the DNN.
Second, the mIoU on EED-transformed data itself also provides a strong estimator, ranked 2nd on Cityscapes and ranked 1st on ADE20k. Hence, the more robust a given DNN is to the absence of texture, the more robust it is w.r.t.\ image corruptions.

In summary, the relationship between cue-dependence and robustness seems to be different to the image classification case. While it is helpful to know how much information a DNN for image classification extracts from the texture-cue, this seems of less importance to semantic segmentation DNNs. It rather matters how much information they extract from the shape cue.

\begin{table}
    \centering
    \scalebox{0.85}{\begin{tabular}{llrrrrrr}
\toprule
 \multicolumn{2}{c}{Dataset: \bfseries ADE20k} & \multicolumn{6}{c}{\bfseries Rank Correlation to Relative Robustness} \\
 &  & \bfseries Cont. & \bfseries High & \bfseries Low & \bfseries Noise & \bfseries Phase & \bfseries Mean \\
Metric & Data &  & \bfseries Pass & \bfseries Pass &  & \bfseries Noise &  \\
\midrule
\multirow[c]{6}{*}{mIoU} & EED & {\cellcolor[HTML]{E7E7FF}} \color[HTML]{000000} \color{black} 0.309 & {\cellcolor[HTML]{DFDFFF}} \color[HTML]{000000} \color{black} 0.421 & \bfseries {\cellcolor[HTML]{BABAFF}} \color[HTML]{000000} \color{black} 0.902 & \bfseries {\cellcolor[HTML]{BDBDFF}} \color[HTML]{000000} \color{black} 0.867 & \bfseries {\cellcolor[HTML]{B9B9FF}} \color[HTML]{000000} \color{black} 0.914 & \bfseries {\cellcolor[HTML]{BEBEFF}} \color[HTML]{000000} \color{black} 0.846 \\
\cline{2-2}
 & ADE20k & \bfseries {\cellcolor[HTML]{D8D8FF}} \color[HTML]{000000} \color{black} 0.507 & \bfseries {\cellcolor[HTML]{D9D9FF}} \color[HTML]{000000} \color{black} 0.493 & {\cellcolor[HTML]{C4C4FF}} \color[HTML]{000000} \color{black} 0.761 & {\cellcolor[HTML]{C9C9FF}} \color[HTML]{000000} \color{black} 0.705 & {\cellcolor[HTML]{C7C7FF}} \color[HTML]{000000} \color{black} 0.732 & {\cellcolor[HTML]{C8C8FF}} \color[HTML]{000000} \color{black} 0.719 \\
\cline{2-2}
 & Vor. 8 & {\cellcolor[HTML]{E9E9FF}} \color[HTML]{000000} \color{black} 0.285 & {\cellcolor[HTML]{EAEAFF}} \color[HTML]{000000} \color{black} 0.281 & {\cellcolor[HTML]{E3E3FF}} \color[HTML]{000000} \color{black} 0.360 & {\cellcolor[HTML]{E5E5FF}} \color[HTML]{000000} \color{black} 0.342 & {\cellcolor[HTML]{DBDBFF}} \color[HTML]{000000} \color{black} 0.470 & {\cellcolor[HTML]{E4E4FF}} \color[HTML]{000000} \color{black} 0.355 \\
\cline{2-2}
 & Vor. 64 & {\cellcolor[HTML]{FBFBFF}} \color[HTML]{000000} \color{black} 0.048 & {\cellcolor[HTML]{FBFBFF}} \color[HTML]{000000} \color{black} 0.053 & {\cellcolor[HTML]{E1E1FF}} \color[HTML]{000000} \color{black} 0.384 & {\cellcolor[HTML]{E3E3FF}} \color[HTML]{000000} \color{black} 0.360 & {\cellcolor[HTML]{E1E1FF}} \color[HTML]{000000} \color{black} 0.388 & {\cellcolor[HTML]{E7E7FF}} \color[HTML]{000000} \color{black} 0.314 \\
\cline{2-2}
 & Vor. 32 & {\cellcolor[HTML]{F6F6FF}} \color[HTML]{000000} \color{black} 0.112 & {\cellcolor[HTML]{F3F3FF}} \color[HTML]{000000} \color{black} 0.151 & {\cellcolor[HTML]{E9E9FF}} \color[HTML]{000000} \color{black} 0.286 & {\cellcolor[HTML]{EAEAFF}} \color[HTML]{000000} \color{black} 0.275 & {\cellcolor[HTML]{E2E2FF}} \color[HTML]{000000} \color{black} 0.381 & {\cellcolor[HTML]{EBEBFF}} \color[HTML]{000000} \color{black} 0.261 \\
\cline{2-2}
 & Vor. 16 & {\cellcolor[HTML]{EEEEFF}} \color[HTML]{000000} \color{black} 0.218 & {\cellcolor[HTML]{EBEBFF}} \color[HTML]{000000} \color{black} 0.258 & {\cellcolor[HTML]{ECECFF}} \color[HTML]{000000} \color{black} 0.247 & {\cellcolor[HTML]{EFEFFF}} \color[HTML]{000000} \color{black} 0.204 & {\cellcolor[HTML]{E3E3FF}} \color[HTML]{000000} \color{black} 0.365 & {\cellcolor[HTML]{EDEDFF}} \color[HTML]{000000} \color{black} 0.235 \\
\cline{1-2} \cline{2-2}
\multirow[c]{4}{*}{\(S_\mathrm{cd}\)} & EED, Vor. 16 & {\cellcolor[HTML]{F0F0FF}} \color[HTML]{000000} \color{black} 0.198 & {\cellcolor[HTML]{E5E5FF}} \color[HTML]{000000} \color{black} 0.339 & {\cellcolor[HTML]{C3C3FF}} \color[HTML]{000000} \color{black} 0.786 & {\cellcolor[HTML]{C3C3FF}} \color[HTML]{000000} \color{black} 0.781 & {\cellcolor[HTML]{C6C6FF}} \color[HTML]{000000} \color{black} 0.735 & {\cellcolor[HTML]{C5C5FF}} \color[HTML]{000000} \color{black} 0.751 \\
\cline{2-2}
 & EED, Vor. 8 & {\cellcolor[HTML]{F4F4FF}} \color[HTML]{000000} \color{black} 0.139 & {\cellcolor[HTML]{E8E8FF}} \color[HTML]{000000} \color{black} 0.296 & {\cellcolor[HTML]{C0C0FF}} \color[HTML]{000000} \color{black} 0.818 & {\cellcolor[HTML]{C3C3FF}} \color[HTML]{000000} \color{black} 0.784 & {\cellcolor[HTML]{C7C7FF}} \color[HTML]{000000} \color{black} 0.728 & {\cellcolor[HTML]{C6C6FF}} \color[HTML]{000000} \color{black} 0.737 \\
\cline{2-2}
 & EED, Vor. 32 & {\cellcolor[HTML]{F0F0FF}} \color[HTML]{000000} \color{black} 0.188 & {\cellcolor[HTML]{E6E6FF}} \color[HTML]{000000} \color{black} 0.332 & {\cellcolor[HTML]{CBCBFF}} \color[HTML]{000000} \color{black} 0.675 & {\cellcolor[HTML]{CCCCFF}} \color[HTML]{000000} \color{black} 0.661 & {\cellcolor[HTML]{CFCFFF}} \color[HTML]{000000} \color{black} 0.632 & {\cellcolor[HTML]{CDCDFF}} \color[HTML]{000000} \color{black} 0.649 \\
\cline{2-2}
 & EED, Vor. 64 & {\cellcolor[HTML]{F4F4FF}} \color[HTML]{000000} \color{black} 0.133 & {\cellcolor[HTML]{E9E9FF}} \color[HTML]{000000} \color{black} 0.282 & {\cellcolor[HTML]{E0E0FF}} \color[HTML]{000000} \color{black} 0.411 & {\cellcolor[HTML]{E1E1FF}} \color[HTML]{000000} \color{black} 0.396 & {\cellcolor[HTML]{E2E2FF}} \color[HTML]{000000} \color{black} 0.374 & {\cellcolor[HTML]{E0E0FF}} \color[HTML]{000000} \color{black} 0.404 \\
\cline{1-2} \cline{2-2}
\multirow[c]{4}{*}{\(R_\mathrm{cd}\)} & EED, Vor. 64 & {\cellcolor[HTML]{FEFEFF}} \color[HTML]{000000} \color{black} 0.011 & {\cellcolor[HTML]{F7F7FF}} \color[HTML]{000000} \color{black} 0.107 & {\cellcolor[HTML]{CECEFF}} \color[HTML]{000000} \color{black} 0.644 & {\cellcolor[HTML]{CECEFF}} \color[HTML]{000000} \color{black} 0.633 & {\cellcolor[HTML]{CBCBFF}} \color[HTML]{000000} \color{black} 0.672 & {\cellcolor[HTML]{D3D3FF}} \color[HTML]{000000} \color{black} 0.570 \\
\cline{2-2}
 & EED, Vor. 32 & {\cellcolor[HTML]{FFF7F7}} \color[HTML]{000000} \color{black} -0.072 & {\cellcolor[HTML]{FCFCFF}} \color[HTML]{000000} \color{black} 0.040 & {\cellcolor[HTML]{D3D3FF}} \color[HTML]{000000} \color{black} 0.563 & {\cellcolor[HTML]{D4D4FF}} \color[HTML]{000000} \color{black} 0.558 & {\cellcolor[HTML]{CFCFFF}} \color[HTML]{000000} \color{black} 0.625 & {\cellcolor[HTML]{DADAFF}} \color[HTML]{000000} \color{black} 0.479 \\
\cline{2-2}
 & EED, Vor. 16 & {\cellcolor[HTML]{FFFAFA}} \color[HTML]{000000} \color{black} -0.053 & {\cellcolor[HTML]{F9F9FF}} \color[HTML]{000000} \color{black} 0.075 & {\cellcolor[HTML]{D8D8FF}} \color[HTML]{000000} \color{black} 0.504 & {\cellcolor[HTML]{D8D8FF}} \color[HTML]{000000} \color{black} 0.504 & {\cellcolor[HTML]{D2D2FF}} \color[HTML]{000000} \color{black} 0.589 & {\cellcolor[HTML]{DDDDFF}} \color[HTML]{000000} \color{black} 0.439 \\
\cline{2-2}
 & EED, Vor. 8 & {\cellcolor[HTML]{FFF3F3}} \color[HTML]{000000} \color{black} -0.118 & {\cellcolor[HTML]{FFFFFF}} \color[HTML]{000000} \color{black} 0.002 & {\cellcolor[HTML]{DCDCFF}} \color[HTML]{000000} \color{black} 0.456 & {\cellcolor[HTML]{DCDCFF}} \color[HTML]{000000} \color{black} 0.454 & {\cellcolor[HTML]{D2D2FF}} \color[HTML]{000000} \color{black} 0.584 & {\cellcolor[HTML]{E2E2FF}} \color[HTML]{000000} \color{black} 0.381 \\
\cline{1-2} \cline{2-2}
\bottomrule
\end{tabular}
}
    \caption{
    Rank Correlation of different mIoU, cue-decomposition shape bias \(S_\mathrm{cd}\) and robustness \(R_\mathrm{cd}\) metrics to relative corruption robustness for semantic segmentation on ADE20k.
    }
    \label{tab:semsegade20kCorr}
\end{table}

\section{Conclusion}

We presented a new AI-free evaluation procedure for image cue biases. Based on two data-processing schemes, we decomposed datasets into a texture and a shape version. Those are used for the definition of two evaluation metrics, the cue-decomposition shape bias and the cue-decomposition robustness. 
We evaluate the shape bias and robustness of more than $60$ DNNs, allowing to compare different architectures types like CNNs, transformer and VLMs including different backbones side-by-side.
While our results for image classification are strongly aligned with findings of previous works, we outperform them in estimating the robustness of DNNs. Furthermore, our procedure allows to measure shape biases for semantic segmentation and we provide the first broader study of semantic segmentation DNNs on two datasets, showing that their biases are similar to those of image classification DNNs, however, the implications on robustness are different from those in image classification. Additionally, we find alternative ways provided by our cue-decomposition to effectively estimate robustness. We hope that our work paves the way to further insights into the biases of DNNs, in particular on complex vision tasks.

{
    \small
    \bibliographystyle{IEEEtran}
    \bibliography{main}
}

\newpage
\appendix

In this document, we provide extended results of most of the experiments presented in the main manuscript. In particular, we utilize additional models. In some experiments we show more fine-grained evaluation metrics. Additionally, we provide a qualitative example and interpret it.

\section{Extended image classification results}

In this section, we present the full image classification evaluations for all $43$ pre-trained models and additional $4$ self-trained models. To facilitate reproducibility, we provide the full model names as provided in the corresponding software repository in \cref{tab-app:classModels}. Additionally, to the models in the main manuscript, we consider variations of a given model with different backbones. 
Another major add-on are the four models trained by us, using different image augmentation / manipulation schemes to either enhance the shape or the texture bias. These models serve as extreme examples and additional sanity check of our cue-decomposition evaluation procedure. The four models are trained as follows:
\begin{itemize}
    \item \textbf{ResNet101 patch:} A ResNet101 trained on patch shuffled ImageNet with patches of size $28 \times 28$ to maximize the texture bias.
    \item \textbf{ResNet101 style:} A ResNet101 trained on stylized ImageNet images, stylized with paintings analogously to the work by Geirhos et al.~\cite{geirhos2018imagenettrained} in order to maximize shape bias.
    \item \textbf{ViT B16 patch:} A ViT B16 analogous to ResNet101 patch.
    \item \textbf{VIT B16 style:} A ViT B16 analogous to ResNet101 style.
\end{itemize}
We trained all four networks for $300$ epochs on $1.28$ million images of $1,\!000$ classes of the ImageNet training data. Note that, the augmentations applied to the training data differ from the transformations that we apply to the data we use for our bias evaluations.
During training, we applied data augmentation, i.e., color jitter with chance 30\% and intensity 0.4, 50\% chance for horizontal flip, randomized aspect ratio 0.75--1.33 and randomized scale 0.8--1.0.
\begin{table}[t]
    \centering
    \scalebox{0.73}{
\begin{tabular}{lll}
\toprule
 &  & \bfseries Long Model Name \\
Arch. & Model &  \\
\midrule
\multirow[c]{13}{*}{CNN} & ConvNeXt L \cite{zhuang2022aconvnet} & convnext\_large\_mlp.clip\_laion2b\_soup\_ft\_in12k\_in1k\_384 \\
\cline{2-3}
 & DPN92 \cite{Chen2017} & dpn92 \\
\cline{2-3}
 & RegNetY \cite{radosavovic2020designing} & regnety\_1280.swag\_ft\_in1k \\
\cline{2-3}
 & ResNeXt101 \cite{Xie2016} & resnext101\_32x32d.fb\_wsl\_ig1b\_ft\_in1k \\
\cline{2-3}
 & ResNet101 \cite{he2015deepresiduallearningimage} & resnet101 \\
\cline{2-3}
 & ResNet152 \cite{he2015deepresiduallearningimage} & resnet152 \\
\bottomrule
\end{tabular}
}
    \caption{
    All models were pre-trained except the ones clustered as `Trained' and all but the VLMs are classifiers for the $1,\!000$ classes of ImageNet.
    The `Trained' models are excluded for the normalization of the cue-decomposition shape bias and the rank correlations.
    }
    \label{tab-app:classModels}
\end{table}
\begin{table*}[t]
    \centering
    \scalebox{0.94}{\begin{tabular}{llrrrrrr|rrrrrr}
\toprule
 \multicolumn{2}{c}{Dataset: \bfseries ImageNet} & \multicolumn{3}{c}{\bfseries Accuracy} & \multicolumn{2}{c}{\bfseries C.-Dec.} & \bfseries C.-Conf. & \multicolumn{6}{c}{\bfseries Relative Robustness} \\
 &  & \bfseries \(\boldsymbol{Q_O}\) & \bfseries \(\boldsymbol{Q_S}\) & \bfseries \(\boldsymbol{Q_T}\) & \bfseries \(\boldsymbol{S_\mathrm{cd}}\) & \bfseries \(\boldsymbol{R_\mathrm{cd}}\) & \bfseries Shape & \bfseries Cont. & \bfseries High & \bfseries Low & \bfseries Noise & \bfseries Phase & \bfseries Mean \\
Arch. & Model &  &  &  &  &  & \bfseries Bias &  & \bfseries Pass & \bfseries Pass &  & \bfseries Noise &  \\
\midrule
\multirow[c]{13}{*}{CNN} & ConvNeXt L \cite{zhuang2022aconvnet} & {\cellcolor[HTML]{FFCC80}} \color[HTML]{000000} \color{black} 0.996 & {\cellcolor[HTML]{BABAFF}} \color[HTML]{000000} \color{black} 0.838 & {\cellcolor[HTML]{FFCD83}} \color[HTML]{000000} \color{black} 0.969 & {\cellcolor[HTML]{CACAFF}} \color[HTML]{000000} \color{black} 0.563 & {\cellcolor[HTML]{FFD08B}} \color[HTML]{000000} \color{black} 0.907 & {\cellcolor[HTML]{E6E6FF}} \color[HTML]{000000} \color{black} 0.348 & {\cellcolor[HTML]{FFD18B}} \color[HTML]{000000} \color{black} 0.865 & {\cellcolor[HTML]{B5B5FF}} \color[HTML]{000000} \color{black} 0.800 & {\cellcolor[HTML]{FFD18C}} \color[HTML]{000000} \color{black} 0.735 & {\cellcolor[HTML]{B5B5FF}} \color[HTML]{000000} \color{black} 0.962 & {\cellcolor[HTML]{FFD08A}} \color[HTML]{000000} \color{black} 0.558 & {\cellcolor[HTML]{B6B6FF}} \color[HTML]{000000} \color{black} 0.784 \\
\bottomrule
\end{tabular}
}
    \caption{
    (left) Cue-decomposition shape biases and relative robustness metric of classification models with components of the metrics
    as well as the cue-conflict shape bias metric by Geirhos et al.\ (C.-Conf. Shape Bias) for comparison.
    Note that the `Trained' models are not included for the normalization of the cue-decomposition shape bias.
    (right) Relative robustness calculated for different types of image corruptions like contrast modifications (Contr.) and averaged over all corruption types in the last column (Mean).
    }
    \label{tab-app:classMetricsRob}
\end{table*}
\begin{table*}[t]
    \centering
    \scalebox{0.79}{
\begin{tabular}{llrr|rrr|rrr|rrr|rrr}
\toprule
 \multicolumn{2}{c}{Dataset: \bfseries ImageNet} & \multicolumn{2}{c}{\bfseries } & \multicolumn{3}{c}{\bfseries patch} & \multicolumn{3}{c}{\bfseries diamond} & \multicolumn{3}{c}{\bfseries tex.EED} & \multicolumn{3}{c}{\bfseries tex.EEDpat.} \\
 &  & \multicolumn{2}{c}{\bfseries Accuracy} & \bfseries Accuracy & \multicolumn{2}{c}{\bfseries C.-Dec.} & \bfseries Accuracy & \multicolumn{2}{c}{\bfseries C.-Dec.} & \bfseries Accuracy & \multicolumn{2}{c}{\bfseries C.-Dec.} & \bfseries Accuracy & \multicolumn{2}{c}{\bfseries C.-Dec.} \\
 &  & \bfseries \(\boldsymbol{Q_O}\) & \bfseries \(\boldsymbol{Q_S}\) & \bfseries \(\boldsymbol{Q_T}\) & \bfseries \(\boldsymbol{S_\mathrm{cd}}\) & \bfseries \(\boldsymbol{R_\mathrm{cd}}\) & \bfseries \(\boldsymbol{Q_T}\) & \bfseries \(\boldsymbol{S_\mathrm{cd}}\) & \bfseries \(\boldsymbol{R_\mathrm{cd}}\) & \bfseries \(\boldsymbol{Q_T}\) & \bfseries \(\boldsymbol{S_\mathrm{cd}}\) & \bfseries \(\boldsymbol{R_\mathrm{cd}}\) & \bfseries \(\boldsymbol{Q_T}\) & \bfseries \(\boldsymbol{S_\mathrm{cd}}\) & \bfseries \(\boldsymbol{R_\mathrm{cd}}\) \\
Arch. & Model &  &  &  &  &  &  &  &  &  &  &  &  &  &  \\
\bottomrule
\end{tabular}
}
    \caption{
    Different candidates for cue-decomposition shape bias and robustness metrics using different images that contain mainly shape information.
    Note that the trained models are not included for the normalization of the cue-decomposition shape bias.
    }
    \label{tab-app:classMetricCandidates}
\end{table*}
\begin{table*}[t]
    \centering
    \scalebox{0.84}{\begin{tabular}{llrrrrrrr}
\toprule
 \multicolumn{2}{c}{Dataset: \bfseries ImageNet} & \multicolumn{7}{c}{\bfseries Rank Correlation to} \\
 &  & \multicolumn{6}{c}{\bfseries Relative Robustness} & \bfseries C.-Conf. \\
 &  & \bfseries Cont. & \bfseries High & \bfseries Low & \bfseries Noise & \bfseries Phase & \bfseries Mean & \bfseries Shape \\
Metric & Data &  & \bfseries Pass & \bfseries Pass &  & \bfseries Noise &  & \bfseries Bias \\
\midrule
C.-Conf. & (Geirhos et al.\ \cite{geirhos2018imagenettrained}) & {\cellcolor[HTML]{D5D5FF}} \color[HTML]{000000} \color{black} 0.542 & {\cellcolor[HTML]{C9C9FF}} \color[HTML]{000000} \color{black} 0.703 & {\cellcolor[HTML]{BFBFFF}} \color[HTML]{000000} \color{black} 0.832 & {\cellcolor[HTML]{CACAFF}} \color[HTML]{000000} \color{black} 0.693 & {\cellcolor[HTML]{CECEFF}} \color[HTML]{000000} \color{black} 0.637 & {\cellcolor[HTML]{C2C2FF}} \color[HTML]{000000} \color{black} 0.793 & {\cellcolor[HTML]{B2B2FF}} \color[HTML]{000000} \color{black} 1.000 \\
\cline{1-2} \cline{2-2}
\multirow[c]{7}{*}{Accuracy} & EED & {\cellcolor[HTML]{D3D3FF}} \color[HTML]{000000} \color{black} 0.577 & {\cellcolor[HTML]{BEBEFF}} \color[HTML]{000000} \color{black} 0.842 & {\cellcolor[HTML]{BFBFFF}} \color[HTML]{000000} \color{black} 0.833 & {\cellcolor[HTML]{C6C6FF}} \color[HTML]{000000} \color{black} 0.746 & {\cellcolor[HTML]{CACAFF}} \color[HTML]{000000} \color{black} 0.694 & {\cellcolor[HTML]{BBBBFF}} \color[HTML]{000000} \color{black} 0.887 & {\cellcolor[HTML]{B8B8FF}} \color[HTML]{000000} \color{black} 0.925 \\
\bottomrule
\end{tabular}
}
    \caption{
    Rank correlations for the classification task of different cue-conflict (C.-Conf.), cue-decomposition \(S_\mathrm{cd}\) shape bias metrics and cue-decomposition robustness \(R_\mathrm{cd}\) metric candidates to relative corruption robustness and to the cue-conflict shape bias of Geirhos et al.~\cite{geirhos2018imagenettrained}.
    Additionally are the correlations of the accuracy components of the cue-decomposition metrics and the Geirhos et al.\ cue-conflict shape bias included.
    }
    \label{tab-app:classCorr}
\end{table*}
\begin{table*}[t]
    \centering
    \scalebox{0.85}{\begin{tabular}{lll}
\toprule
\multicolumn{2}{c}{Dataset: \bfseries Cityscapes} & \bfseries Long Model Name \\
Arch. & Model &  \\
\midrule
\multirow[c]{9}{*}{CNN} & DeepLabV3Plus RN101 \cite{chen2018encoderdecoderatrousseparableconvolution} & deeplabv3plus\_r50b-d8\_4xb2-80k\_cityscapes-512x1024 \\
\cline{2-3}
 & DeepLabV3Plus RN18 \cite{chen2018encoderdecoderatrousseparableconvolution} & dnl\_r101-d8\_4xb2-40k\_cityscapes-512x1024 \\
\cline{2-3}
 & DeepLabV3Plus RN50 \cite{chen2018encoderdecoderatrousseparableconvolution} & dnl\_r50-d8\_4xb2-40k\_cityscapes-512x1024 \\
\cline{2-3}
 & FCN RN101 \cite{shelhamer2016fullyconvolutionalnetworkssemantic} & fcn-d6\_r50b-d16\_4xb2-80k\_cityscapes-512x1024 \\
\cline{2-3}
 & FCN RN18 \cite{shelhamer2016fullyconvolutionalnetworkssemantic} & fcn\_r101-d8\_4xb2-40k\_cityscapes-512x1024 \\
\cline{2-3}
 & FCN-d6 RN50 \cite{shelhamer2016fullyconvolutionalnetworkssemantic} & fcn\_r18b-d8\_4xb2-80k\_cityscapes-512x1024 \\
\cline{2-3}
 & Mask2Former RN101 \cite{cheng2021mask2former} & mask2former\_r101\_8xb2-90k\_cityscapes-512x1024 \\
\cline{2-3}
 & Mask2Former RN50 \cite{cheng2021mask2former} & mask2former\_r50\_8xb2-90k\_cityscapes-512x1024 \\
\cline{2-3}
 & PSPNet RN50 \cite{zhao2017pyramidsceneparsingnetwork} & pspnet\_r50-d8\_4xb2-40k\_cityscapes-512x1024 \\
\cline{1-3} \cline{2-3}
\multirow[c]{8}{*}{\shortstack{Vision\\ Transf.}} & Mask2Former Swin-B \cite{cheng2021mask2former} & mask2former\_swin-b-in22k-384x384-pre\_8xb2-90k\_cityscapes-512x1024 \\
\cline{2-3}
 & Mask2Former Swin-L \cite{cheng2021mask2former} & mask2former\_swin-l-in22k-384x384-pre\_8xb2-90k\_cityscapes-512x1024 \\
\cline{2-3}
 & SETR ViT-L\_mla \cite{zheng2021rethinkingsemanticsegmentationsequencetosequence} & segformer\_mit-b0\_8xb1-160k\_cityscapes-1024x1024 \\
\cline{2-3}
 & SETR ViT-L\_naive \cite{zheng2021rethinkingsemanticsegmentationsequencetosequence} & segformer\_mit-b2\_8xb1-160k\_cityscapes-1024x1024 \\
\cline{2-3}
 & SETR ViT-L\_pup \cite{zheng2021rethinkingsemanticsegmentationsequencetosequence} & segformer\_mit-b4\_8xb1-160k\_cityscapes-1024x1024 \\
\cline{2-3}
 & SegFormer MiT-b0 \cite{xie2021segformersimpleefficientdesign} & setr\_vit-l\_mla\_8xb1-80k\_cityscapes-768x768 \\
\cline{2-3}
 & SegFormer MiT-b2 \cite{xie2021segformersimpleefficientdesign} & setr\_vit-l\_naive\_8xb1-80k\_cityscapes-768x768 \\
\cline{2-3}
 & SegFormer MiT-b4 \cite{xie2021segformersimpleefficientdesign} & setr\_vit-l\_pup\_8xb1-80k\_cityscapes-768x768 \\
\cline{1-3} \cline{2-3}
\multirow[c]{5}{*}{other} & DNL RN101 \cite{yin2020disentangled} & deeplabv3plus\_r101b-d8\_4xb2-80k\_cityscapes-512x1024 \\
\cline{2-3}
 & DNL RN50 \cite{yin2020disentangled} & deeplabv3plus\_r18-d8\_4xb2-80k\_cityscapes-512x1024 \\
\cline{2-3}
 & OCRNet RN101 \cite{yuan2021segmentationtransformerobjectcontextualrepresentations} & ocrnet\_hr18\_4xb2-40k\_cityscapes-512x1024 \\
\cline{2-3}
 & OCRNet hr18 \cite{yuan2021segmentationtransformerobjectcontextualrepresentations} & ocrnet\_hr48\_4xb2-40k\_cityscapes-512x1024 \\
\cline{2-3}
 & OCRNet hr48 \cite{yuan2021segmentationtransformerobjectcontextualrepresentations} & ocrnet\_r101-d8\_4xb2-40k\_cityscapes-512x1024 \\
\cline{1-3} \cline{2-3}
\bottomrule
\end{tabular}
}
    \caption{Pre-trained Cityscapes semantic segmentation models used for evaluation.}
    \label{tab-app:classModelsCity}
\end{table*}
\paragraph{Cue-decomposition metrics}
The results of our extended choice of models and the ones trained by us are provided in \cref{tab-app:classMetricsRob}. Indeed, the four models trained by us achieve, depending on patching or stylizing, extreme texture or shape biases in both metrics, our cue-decomposition shape bias as well as the cue-conflict shape bias. It is also noteworthy that a given DNN trained on stylized images is more robust than when trained on patch-shuffled data. On the other hand, the DNNs trained on patch-shuffled data achieve higher accuracy on the original data than the DNNs trained on stylized data, which shows that accuracy and robustness are not necessarily tied together.

The different choices of backbones in \cref{tab-app:classMetricsRob} additionally reveal some nuances. VLMs with transformer encoder tend to be more shape biased than VLMs with CNN encoder. We observe also that networks with strong accuracy on the original data ($Q_O$) tend to be more shape biased. This finding appears also reasonable to us since local texture cues should be easily extractable, also by DNNs with smaller capacity and lower performance, while shape cues, being less local, are harder to extract.

\paragraph{Ablation study on cue-extraction procedures}
We provide a more fine-grained view on our ablation study from the main manuscript in \cref{tab-app:classMetricCandidates}, providing multiple metrics per cue extraction procedure over the full range of networks considered in our experiments. It turns out that many models perform well on patch-shuffled and texture EED (difference of original image and EED image) images. However, when combining both, i.e., when patch-shuffling the texture EED image (cf.\ tex.EEDpat. in the table), we see a clear drop in classification accuracy $Q_T$. This observation, among other reasons mentioned in the main manuscript, disqualified the patch-shuffled difference images for further considerations. For the patch-shuffled images, most models perform well in terms of $Q_T$, except for Inception models, SENet, VGGs and DPN92. This also led us to the conclusion that the classification accuracy on patch-shuffled data polarizes strongly across network architectures, being very challenging for some networks and almost trivial for other networks, and thus not a good choice as a texture extraction procedure. Diamond shuffling appeared very challenging for most of the networks, showing low classification accuracy on a wide range of DNNs. For the interested reader, we also provide the results of our trained models, also for the sake of completeness.

\paragraph{A robustness view on the ablation study of different cue-extraction procedures}
In \cref{tab-app:classCorr}, we extend the ablation study by a robustness view. Additionally, we provide a per-image-corruption breakdown of the rank correlations, where the latter are considered as function of the respective cue-decomposition robustness estimate and different cue extraction procedures. The extraction procedures range over different cue-decomposition and cue-conflict procedures.

The mean relative robustness (2nd column from the right) is strong for all $R_\mathrm{cd}$ metrics and for standalone EED as well. This underlines the validity of the general notion behind our $R_\mathrm{cd}$ metric. Compared to the EED-Voronoi combination, most of the other $R_\mathrm{cd}$ metrics and EED accuracy exhibit a strong rank correlation to the cue-conflict metric by Geirhos et al. Looking into the different types of image corruptions, we see that the correlations are relatively similar across the different cue extraction procedures. Our EED-based cue-conflict schemes (bottom section of the table) are unable to estimate the robustness w.r.t.\ high pass filtering, while the cue-decomposition robustness-based $R_\mathrm{cd}$ schemes can deal with high pass filtering very well. Contrast manipulation (Cont., left-most column) represents a challenge for robustness estimation across the board.

\section{Additional semantic segmentation results}
\begin{table*}[t]
    \centering
    \scalebox{0.85}{
\begin{tabular}{lll}
\toprule
\multicolumn{2}{c}{Dataset: \bfseries ADE20k} & \bfseries Long Model Name \\
Arch. & Model &  \\
\midrule
\multirow[c]{7}{*}{CNN} & DeepLabV3Plus RN101 \cite{chen2018encoderdecoderatrousseparableconvolution} & deeplabv3plus\_r101-d8\_4xb4-160k\_ade20k-512x512 \\
\cline{2-3}
 & DeepLabV3Plus RN50 \cite{chen2018encoderdecoderatrousseparableconvolution} & deeplabv3plus\_r50-d8\_4xb4-80k\_ade20k-512x512 \\
\cline{2-3}
 & FCN RN101 \cite{shelhamer2016fullyconvolutionalnetworkssemantic} & fcn\_r101-d8\_4xb4-80k\_ade20k-512x512 \\
\cline{2-3}
 & FCN RN50 \cite{shelhamer2016fullyconvolutionalnetworkssemantic} & fcn\_r50-d8\_4xb4-80k\_ade20k-512x512 \\
\cline{2-3}
 & Mask2Former RN101 \cite{cheng2021mask2former} & mask2former\_r101\_8xb2-160k\_ade20k-512x512 \\
\cline{2-3}
 & Mask2Former RN50 \cite{cheng2021mask2former} & mask2former\_r50\_8xb2-160k\_ade20k-512x512 \\
\bottomrule
\end{tabular}
}
    \caption{Pre-trained ADE20k semantic segmentation models used for evaluation.}
    \label{tab-app:classModelsADE20k}
\end{table*}

\begin{table*}[t]
    \centering
    \scalebox{0.85}{\begin{tabular}{llrrrrr|rrrrr}
\toprule
 &  & \multicolumn{5}{c}{\bfseries Cityscapes} & \multicolumn{5}{c}{\bfseries ADE20k} \\
 &  & \multicolumn{3}{c}{\bfseries mIoU} & \multicolumn{2}{c}{\bfseries C.-Dec.} & \multicolumn{3}{c}{\bfseries mIoU} & \multicolumn{2}{c}{\bfseries C.-Dec.} \\
 &  & \bfseries \(\boldsymbol{Q_O}\) & \bfseries \(\boldsymbol{Q_S}\) & \bfseries \(\boldsymbol{Q_T}\) & \bfseries \(\boldsymbol{S_\mathrm{cd}}\) & \bfseries \(\boldsymbol{R_\mathrm{cd}}\) & \bfseries \(\boldsymbol{Q_O}\) & \bfseries \(\boldsymbol{Q_S}\) & \bfseries \(\boldsymbol{Q_T}\) & \bfseries \(\boldsymbol{S_\mathrm{cd}}\) & \bfseries \(\boldsymbol{R_\mathrm{cd}}\) \\
Arch. & Model &  &  &  &  &  &  &  &  &  &  \\
\midrule
\multirow[c]{10}{*}{CNN} & Mask2Former RN50 \cite{cheng2021mask2former} & {\cellcolor[HTML]{FFD89E}} \color[HTML]{000000} \color{black} 0.804 & {\cellcolor[HTML]{F3F3FF}} \color[HTML]{000000} \color{black} 0.206 & {\cellcolor[HTML]{FFF9F0}} \color[HTML]{000000} \color{black} 0.303 & {\cellcolor[HTML]{DCDCFF}} \color[HTML]{000000} \color{black} 0.488 & {\cellcolor[HTML]{FFF9F0}} \color[HTML]{000000} \color{black} 0.316 & {\cellcolor[HTML]{D2D2FF}} \color[HTML]{000000} \color{black} 0.466 & {\cellcolor[HTML]{FFECCE}} \color[HTML]{000000} \color{black} 0.208 & {\cellcolor[HTML]{E6E6FF}} \color[HTML]{000000} \color{black} 0.102 & {\cellcolor[HTML]{FFE1B4}} \color[HTML]{000000} \color{black} 0.513 & {\cellcolor[HTML]{FDFDFF}} \color[HTML]{000000} \color{black} 0.333 \\
\bottomrule
\end{tabular}
}
    \caption{
    Cue-decomposition shape biases and robustness metrics of semantic segmentation models with components of the metrics.
    On the left for the Cityscapes dataset and on the right for ADE20k.
    }
    \label{tab-app:semsegMetrics}
\end{table*}

\begin{table*}[t]
    \centering
    \scalebox{0.83}{\begin{tabular}{llrrrrrrr|rrrrrrr}
\toprule
 &  & \multicolumn{7}{c}{\bfseries Cityscapes} & \multicolumn{7}{c}{\bfseries ADE20k} \\
 &  & \multicolumn{6}{c}{\bfseries Relative Robustness} & \bfseries C.-Dec. & \multicolumn{6}{c}{\bfseries Relative Robustness} & \bfseries C.-Dec. \\
 &  & \bfseries Cont. & \bfseries High & \bfseries Low & \bfseries Noise & \bfseries Phase & \bfseries Mean & \bfseries \(\boldsymbol{R_\mathrm{cd}}\) & \bfseries Cont. & \bfseries High & \bfseries Low & \bfseries Noise & \bfseries Phase & \bfseries Mean & \bfseries \(\boldsymbol{R_\mathrm{cd}}\) \\
Arch. & Model &  & \bfseries Pass & \bfseries Pass &  & \bfseries Noise &  &  &  & \bfseries Pass & \bfseries Pass &  & \bfseries Noise &  &  \\
\midrule
\multirow[c]{10}{*}{CNN} & DeepLabV3Plus RN101 \cite{chen2018encoderdecoderatrousseparableconvolution} & {\cellcolor[HTML]{FFF0DA}} \color[HTML]{000000} \color{black} 0.741 & {\cellcolor[HTML]{C3C3FF}} \color[HTML]{000000} \color{black} 0.337 & {\cellcolor[HTML]{FFF7EA}} \color[HTML]{000000} \color{black} 0.445 & {\cellcolor[HTML]{F2F2FF}} \color[HTML]{000000} \color{black} 0.373 & {\cellcolor[HTML]{FFF0DA}} \color[HTML]{000000} \color{black} 0.257 & {\cellcolor[HTML]{ECECFF}} \color[HTML]{000000} \color{black} 0.431 & {\cellcolor[HTML]{FFF6EA}} \color[HTML]{000000} \color{black} 0.328 & {\cellcolor[HTML]{D2D2FF}} \color[HTML]{000000} \color{black} 0.679 & {\cellcolor[HTML]{FFF0DA}} \color[HTML]{000000} \color{black} 0.162 & {\cellcolor[HTML]{F4F4FF}} \color[HTML]{000000} \color{black} 0.404 & {\cellcolor[HTML]{FFFAF4}} \color[HTML]{000000} \color{black} 0.534 & {\cellcolor[HTML]{F4F4FF}} \color[HTML]{000000} \color{black} 0.348 & {\cellcolor[HTML]{FFF6EA}} \color[HTML]{000000} \color{black} 0.425 & {\cellcolor[HTML]{F8F8FF}} \color[HTML]{000000} \color{black} 0.340 \\
\bottomrule
\end{tabular}
}
    \caption{
    Relative robustness calculated for different types of image corruptions like contrast modifications (Contr.) and averaged over all corruption types in the column `Mean'. This is compared to our cue-decomposition robustness metric (last column)
    On the left for the Cityscapes dataset and on the right for ADE20k.
    }
    \label{tab-app:semsegRob}
\end{table*}

In this section, we provide results from additional semantic segmentation networks. The full list of the evaluated semantic segmentation models and their corresponding model names as used in the software frameworks is listed in \cref{tab-app:classModelsCity} for Cityscapes-trained models and in \cref{tab-app:classModelsADE20k} for ADE20k-trained models.
In addition to the main paper, the backbones architectures are varied or different decoders are used in the case of SETR.

\paragraph{Cue-decomposition shape bias \& robustness}
In the main paper we already showed that the choice of architecture type, e.g. CNN or transformer, has the most impact on the metrics. This can also be seen in the enlarged \cref{tab-app:semsegMetrics}.
This finding also generalizes to the choice of decoder for SETR and the choice of backbone type for OCR-Net.
In comparison, changing the backbone model, e.g.\ replacing ResNet50 with ResNet101, has less influence on the shape bias and the cue extraction capabilities.
However, the depth of the models often increases the shape cue performance $Q_S$ (cf.\ Segformer) whereas the texture cue performance $Q_T$ seems not to be correlated to the model depth. This is reasonable, since more complex patterns and spacial hierarchies can be learned with deeper models.

In \cref{tab-app:semsegRob} we provide the extended results of the relative robustness results for Cityscapes as well as ADE20k.
We see only minor differences when exchanging the backbone with smaller or larger variants.
When comparing \cref{tab-app:semsegMetrics} with \cref{tab-app:semsegRob}, we observe that segmentation models with high texture extraction capability ($Q_T$) like Mask2Former is more robust to high-pass and noise corruptions whereas models with higher shape extraction capability like SETR are more robust to low pass corruptions that perturb the texture of an image.
Furthermore, our evaluation procedure allows to understand the differences in model types like SETR-mla and SETR-pup.
For Cityscapes, the performance of SETR-pup w.r.t.\ $Q_S$ and $Q_T$ exceeds the respective performances of SETR-mla by several percentage points. On ADE20k, the situation is vice-versa although not as pronounced as on Cityscapes. Taking a closer look at the decoder architectures and the datasets, our impression is that the pup (progressive upsampling) decoder, which operates on a single scale at the target resolution, is very suitable for Cityscapes. Images from Cityscapes have the property that objects in a given image location occur at very similar scales. This is not the case in ADE20k where the camera location in a scene and the context are way less static. Hence, on ADE20k, the mla (multi-level feature aggregation) decoder appears to be a more suitable choice, as it is potentially able to handle multiple object scales at the given target resolution.

\begin{figure*}[h]
    \centering
    \begin{tabular}{|p{0.11\textwidth}|>{\centering\arraybackslash}m{0.25\textwidth}|>{\centering\arraybackslash}m{0.25\textwidth}|>{\centering\arraybackslash}m{0.25\textwidth}|}
        \hline
        \textbf{} & \textbf{Original} & \textbf{EED} & \textbf{Voronoi} \\
        \hline
        \vspace{-32pt} Input & 
        \includegraphics[width=0.25\textwidth]{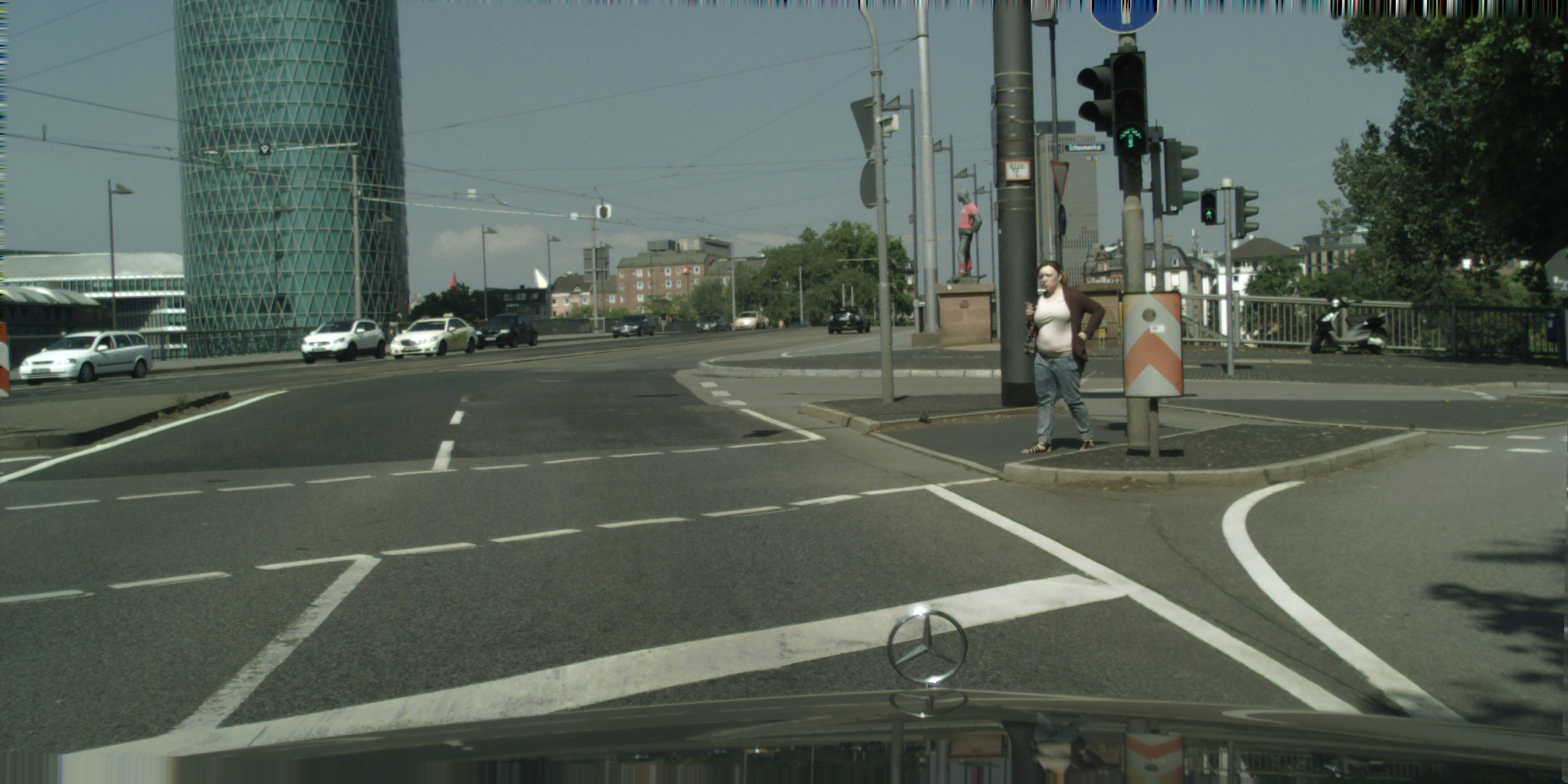} &
        \includegraphics[width=0.25\textwidth]{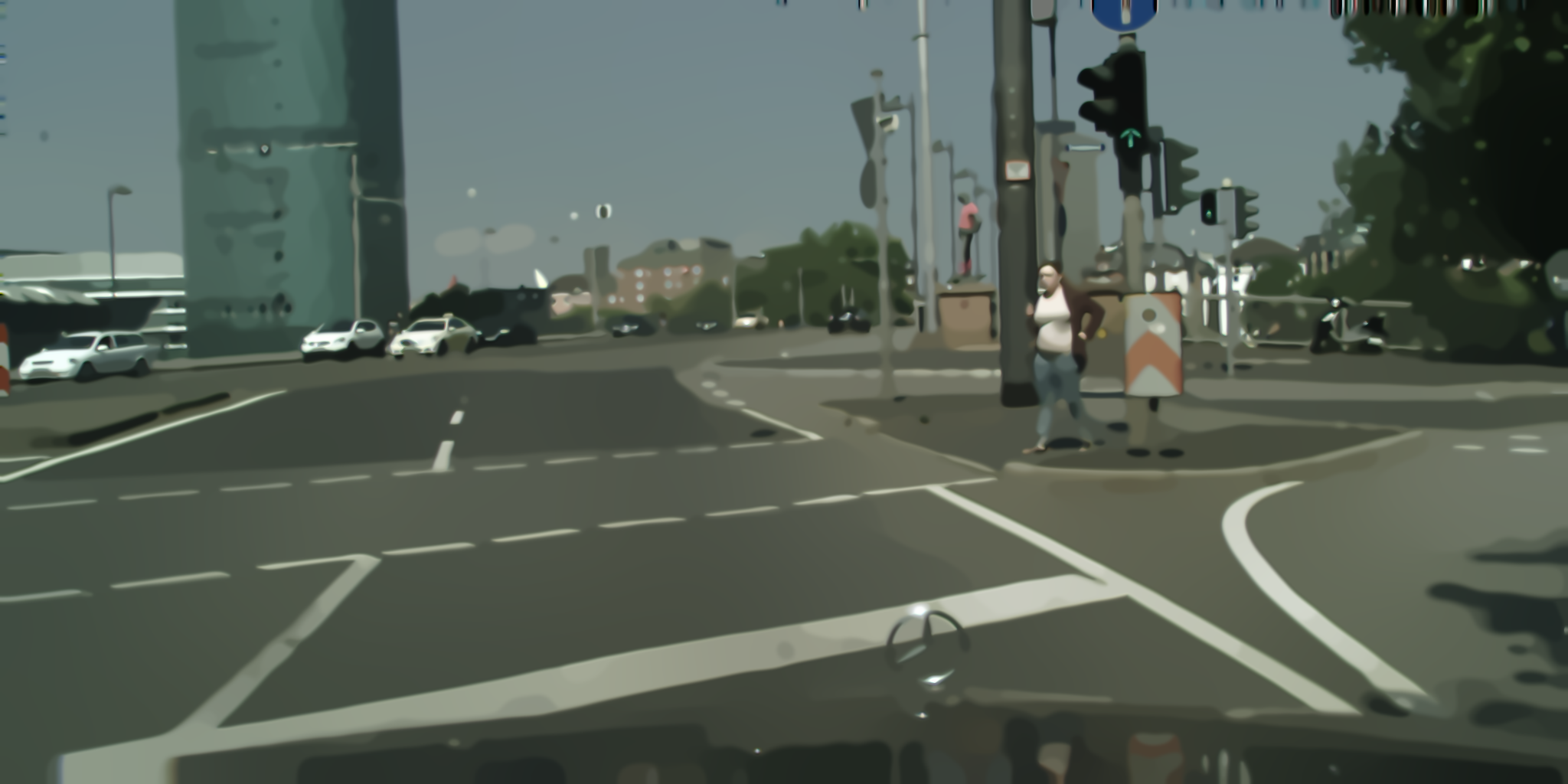} & 
        \includegraphics[width=0.25\textwidth]{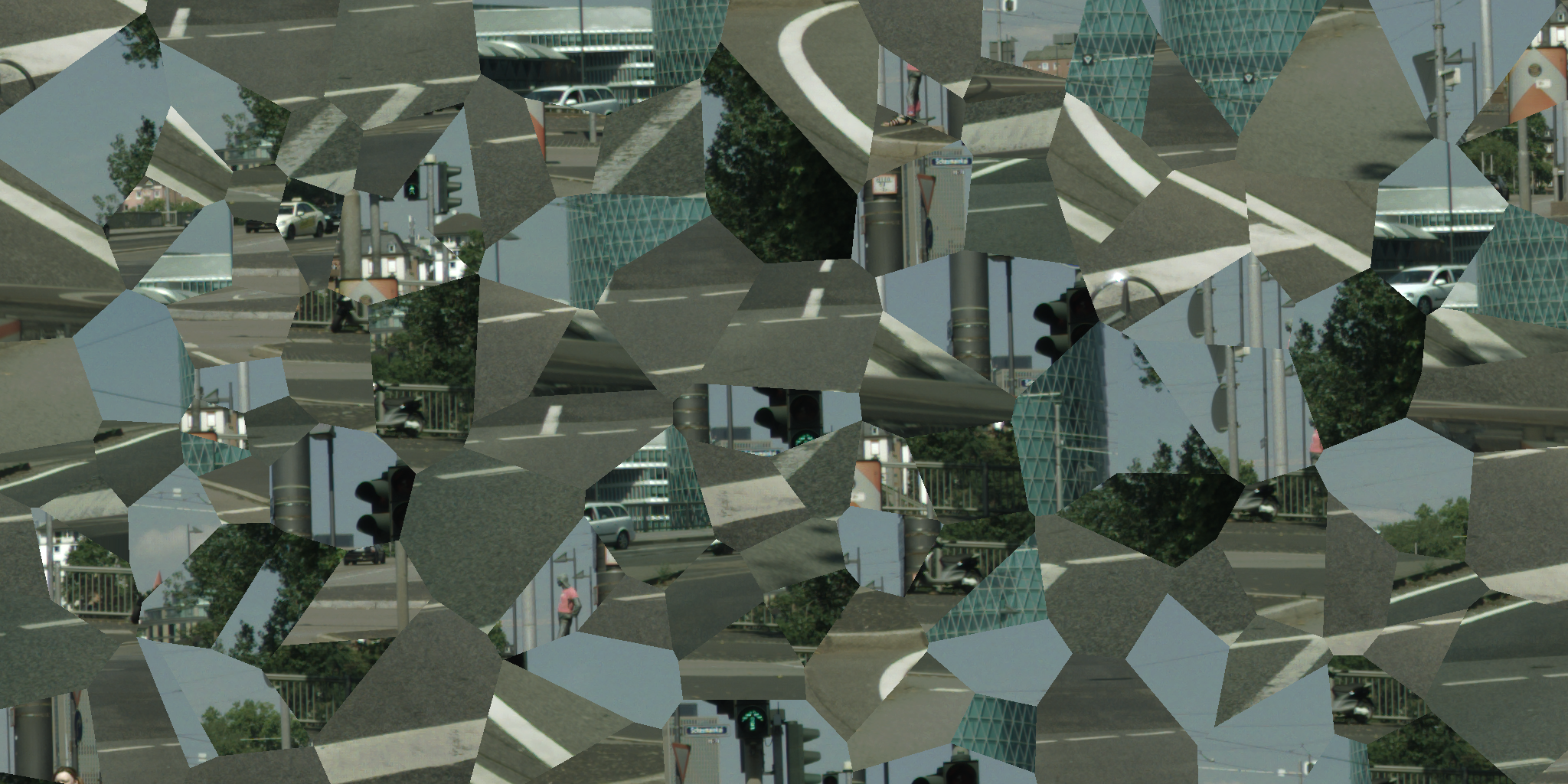} \\
        \hline
        \vspace{-32pt} Ground truth & 
        \includegraphics[width=0.25\textwidth]{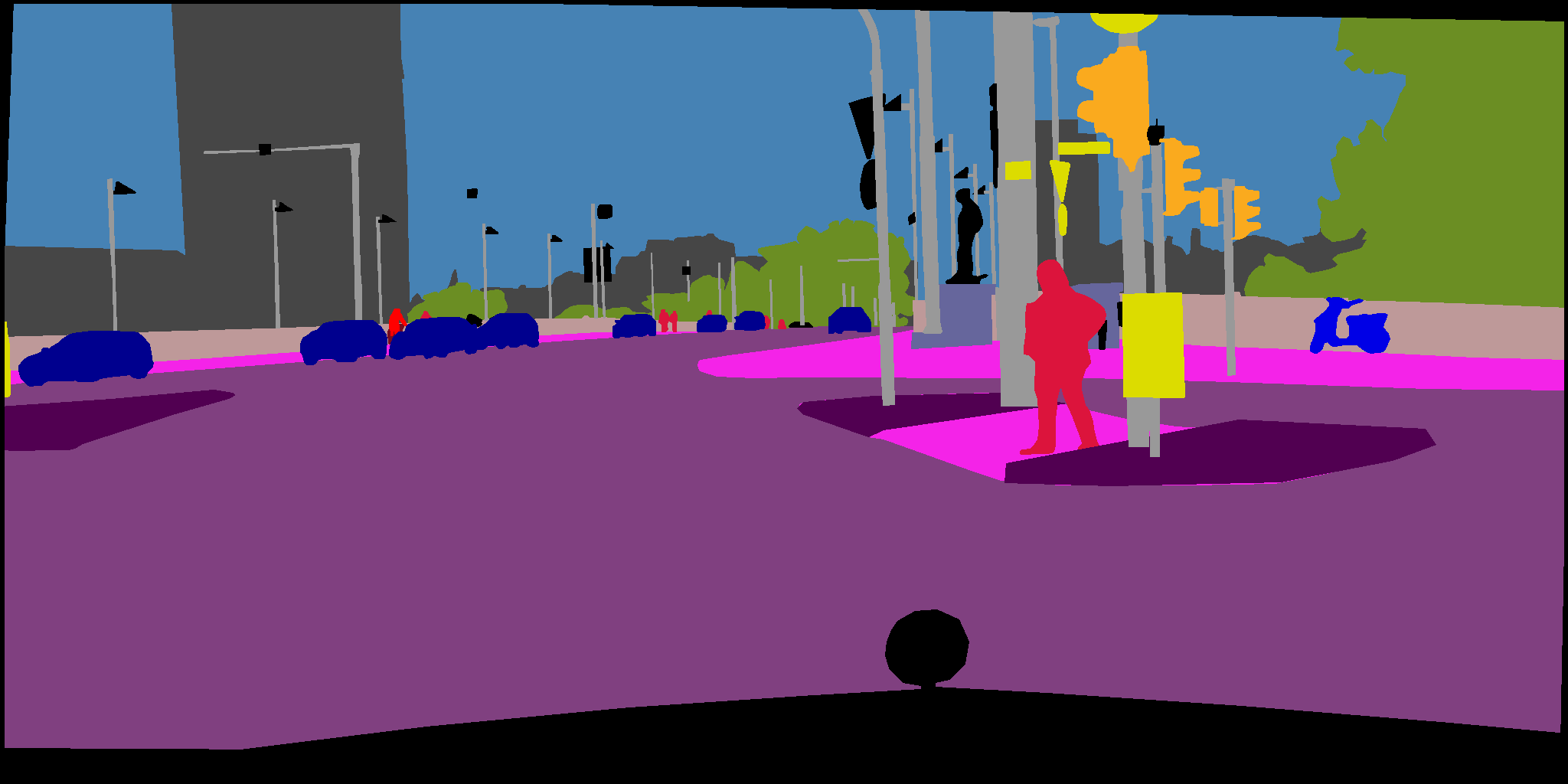} &
        \includegraphics[width=0.25\textwidth]{images/semseg_ex2/frankfurt_000000_009561_gtFine_color.png} & 
        \includegraphics[width=0.25\textwidth]{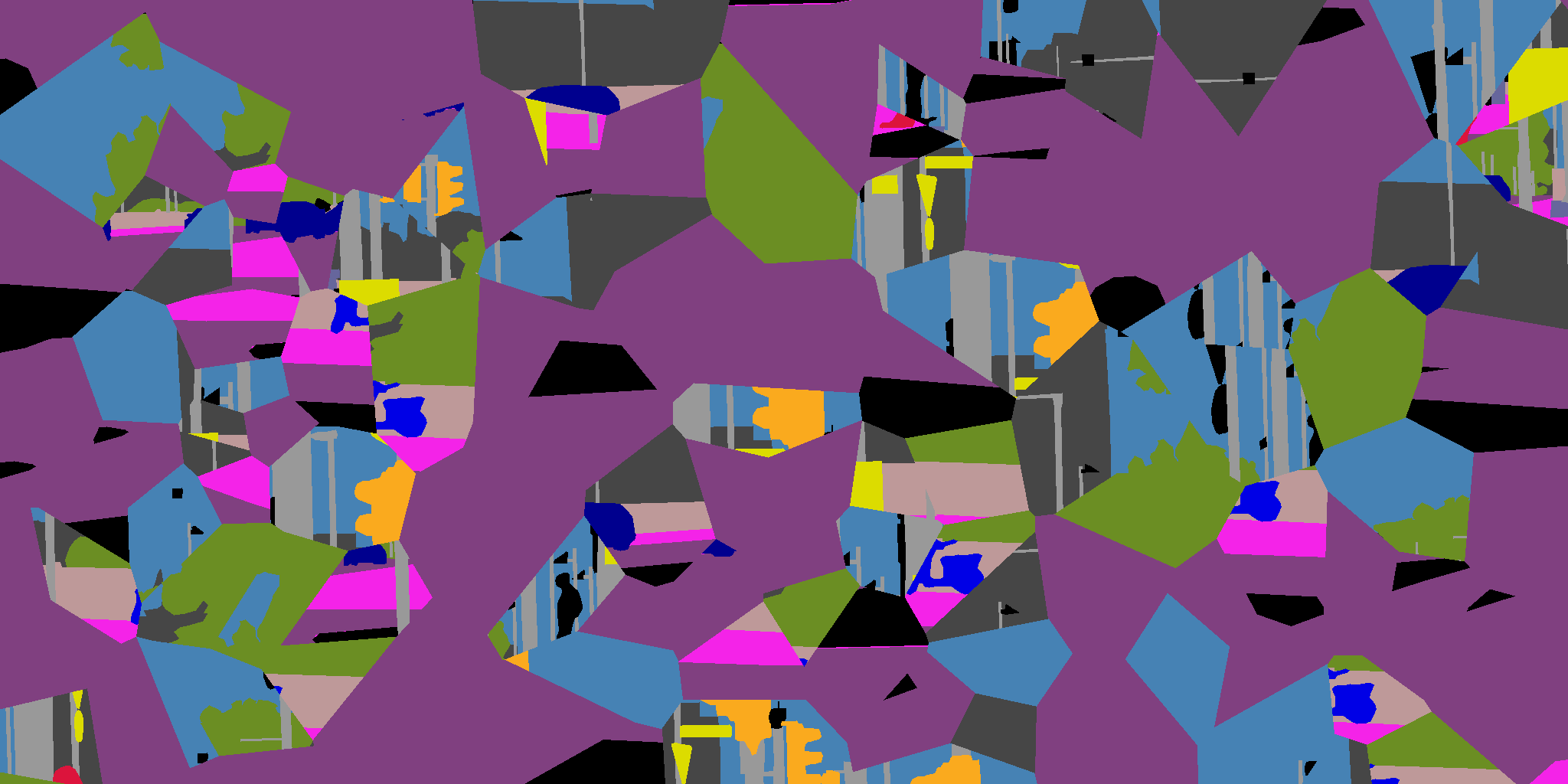} \\
        \hline
        \vspace{-32pt} DeepLab V3Plus RN101 & 
        \includegraphics[width=0.25\textwidth]{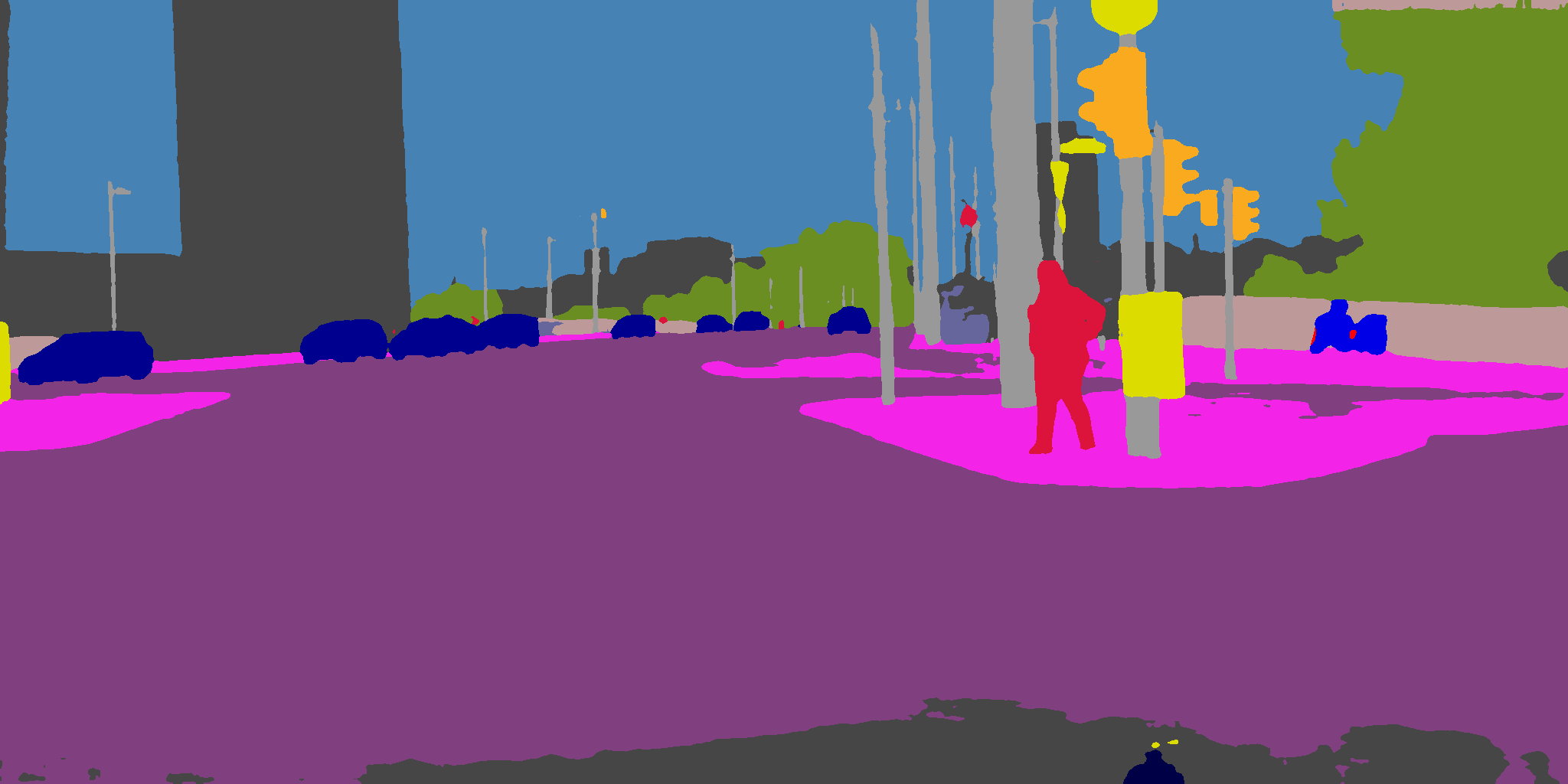} & 
        \includegraphics[width=0.25\textwidth]{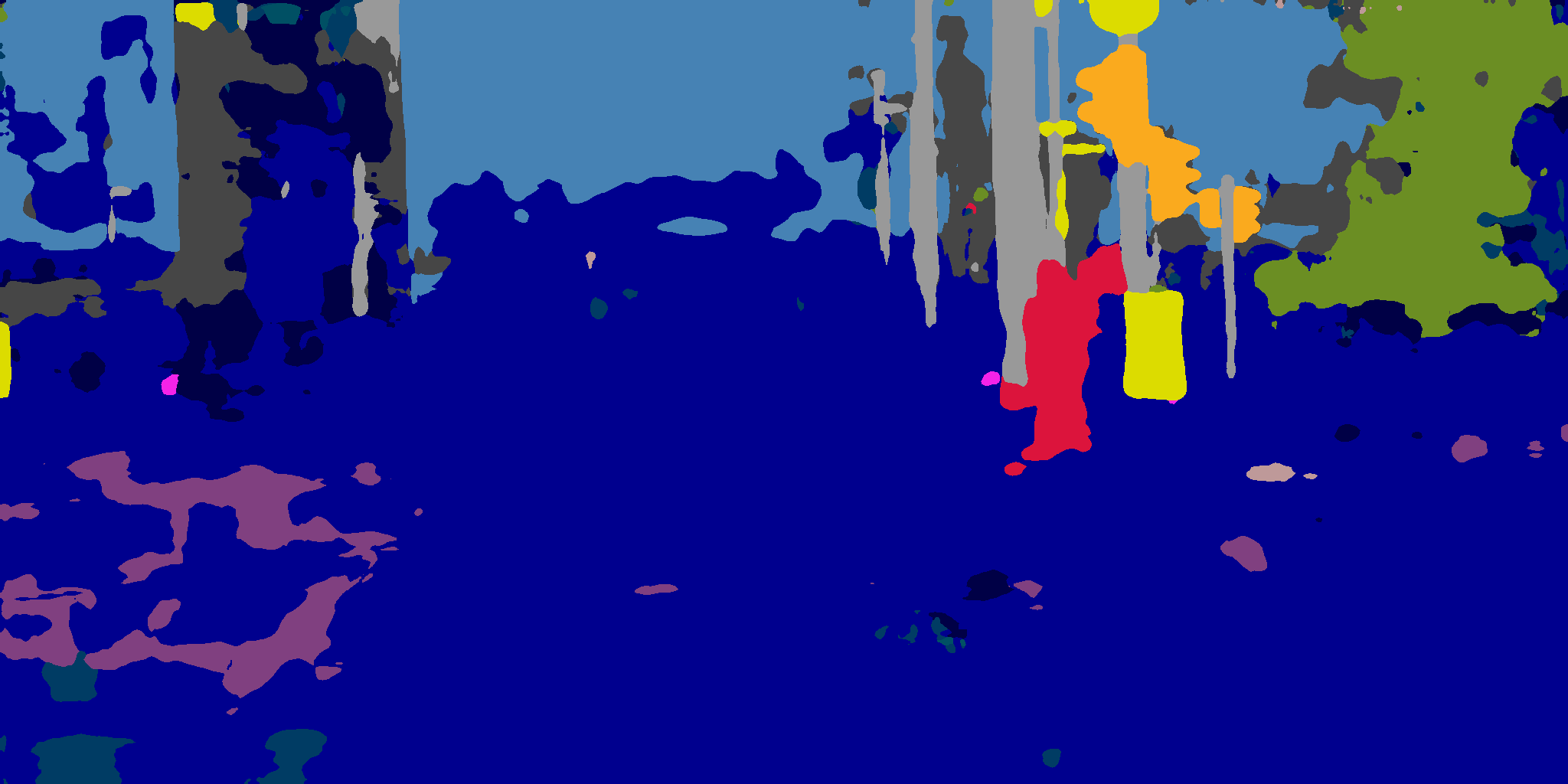} & 
        \includegraphics[width=0.25\textwidth]{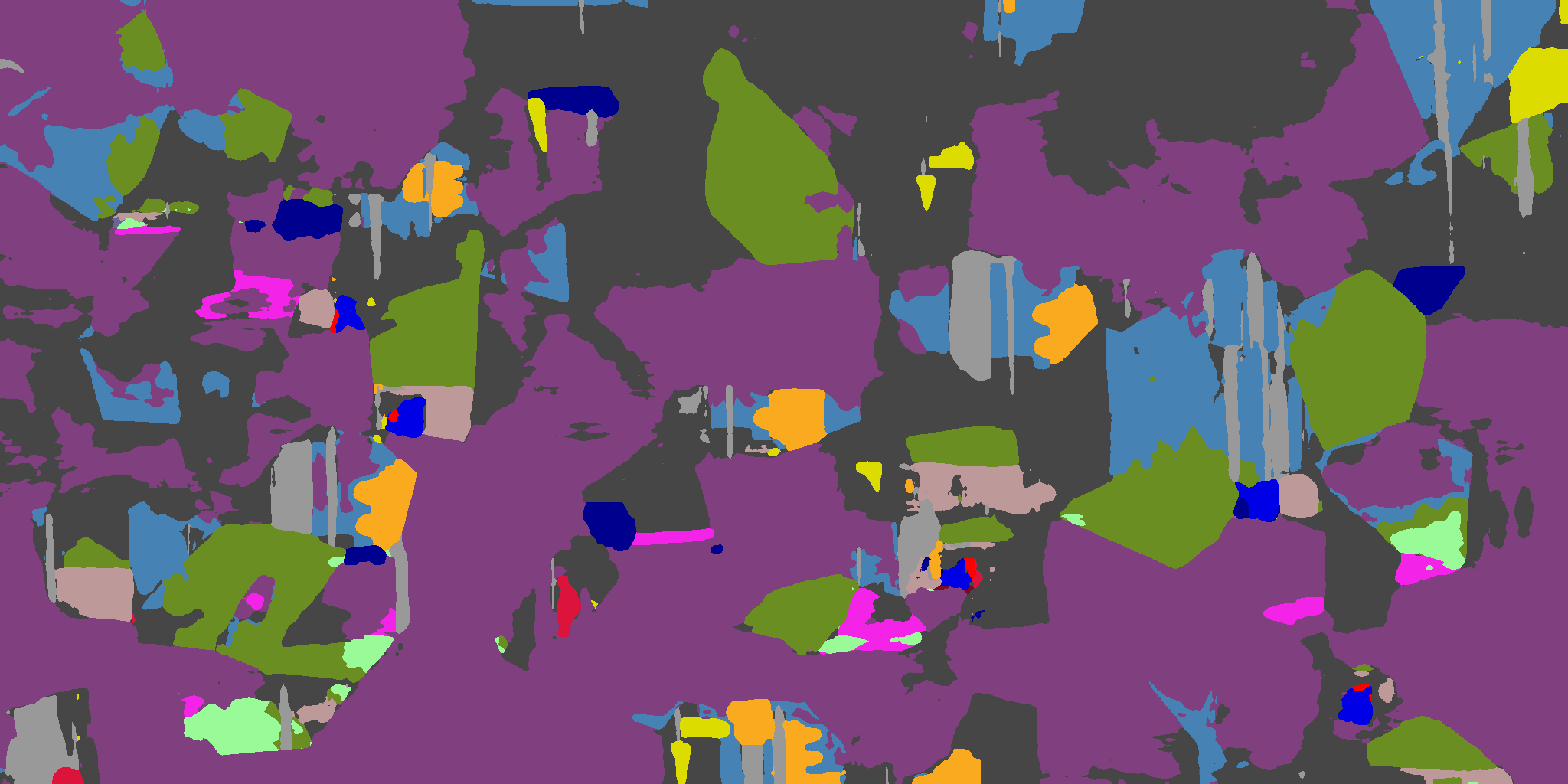} \\
        $S_\mathrm{cd} = 0.421$ & \(Q_O=0.802\) & \(Q_S=0.180\) & \(Q_T=0.346\) \\
        \hline
        \vspace{-32pt} Mask2 \mbox{Former} Swin-L & 
        \includegraphics[width=0.25\textwidth]{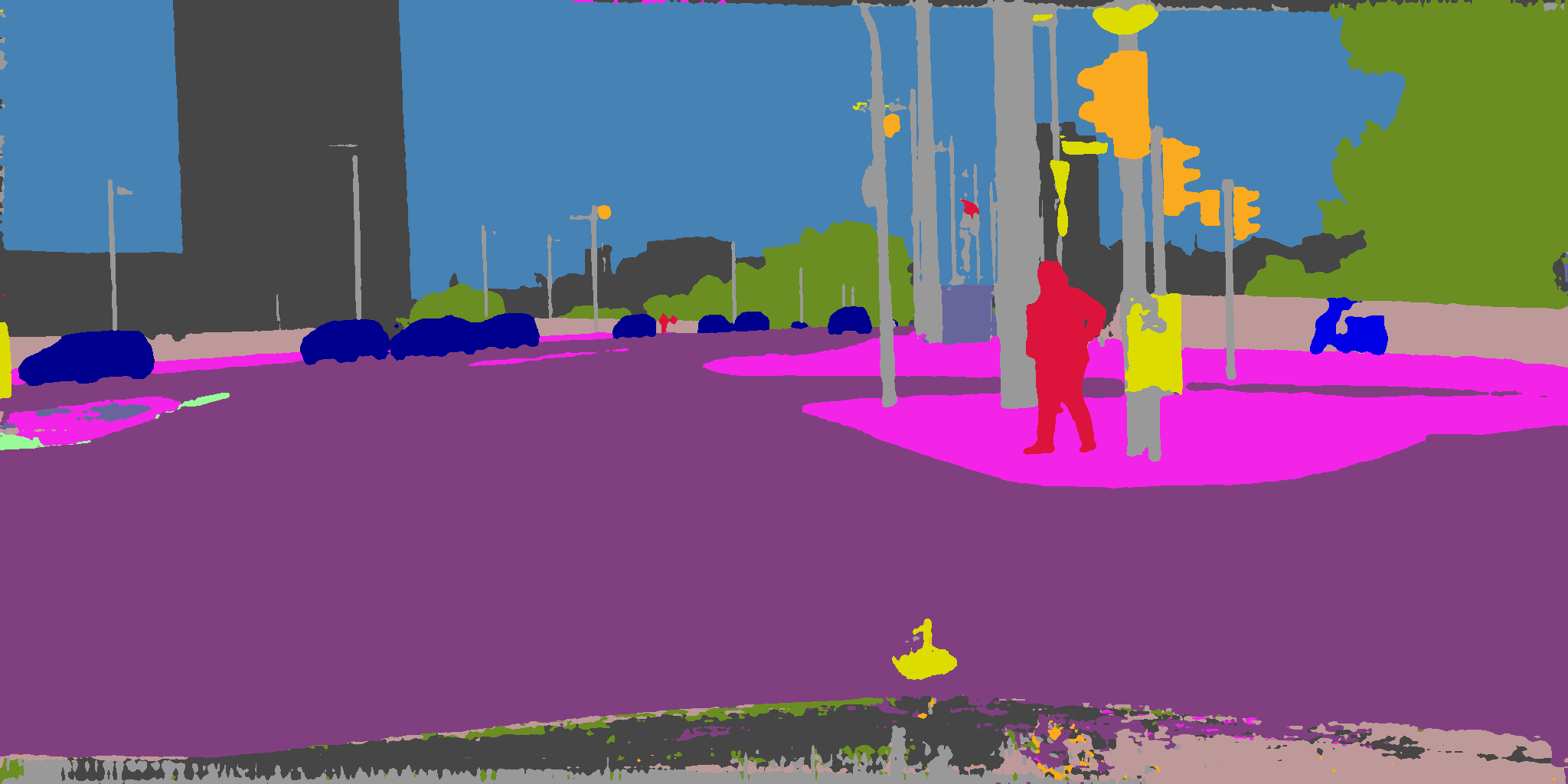} & 
        \includegraphics[width=0.25\textwidth]{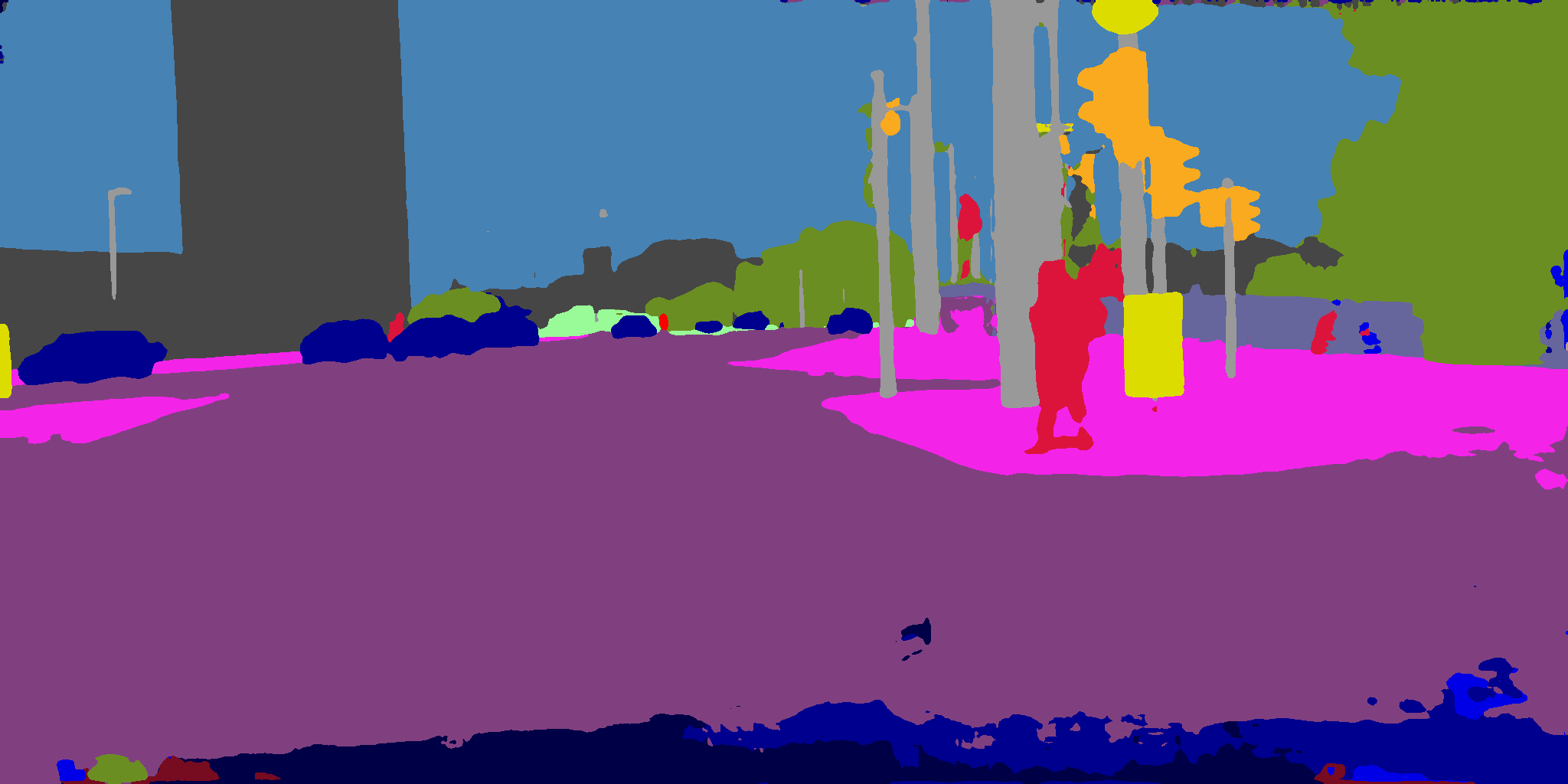} & 
        \includegraphics[width=0.25\textwidth]{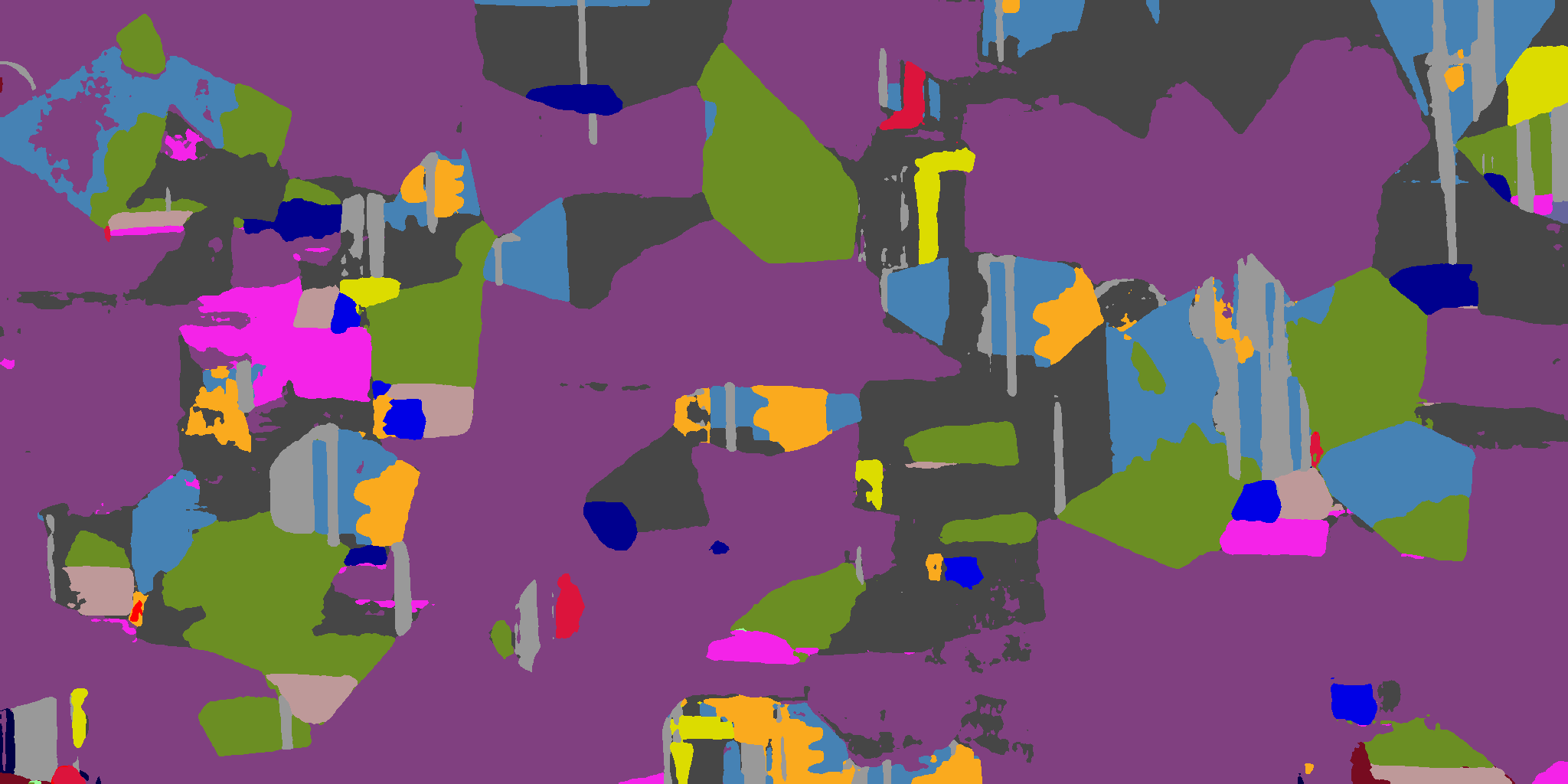} \\
        $S_\mathrm{cd} = 0.562$ & \(Q_O=0.837\) & \(Q_S=0.410\) & \(Q_T=0.447\) \\
        \hline
        \vspace{-32pt} SETR-Pup & 
        \includegraphics[width=0.25\textwidth]{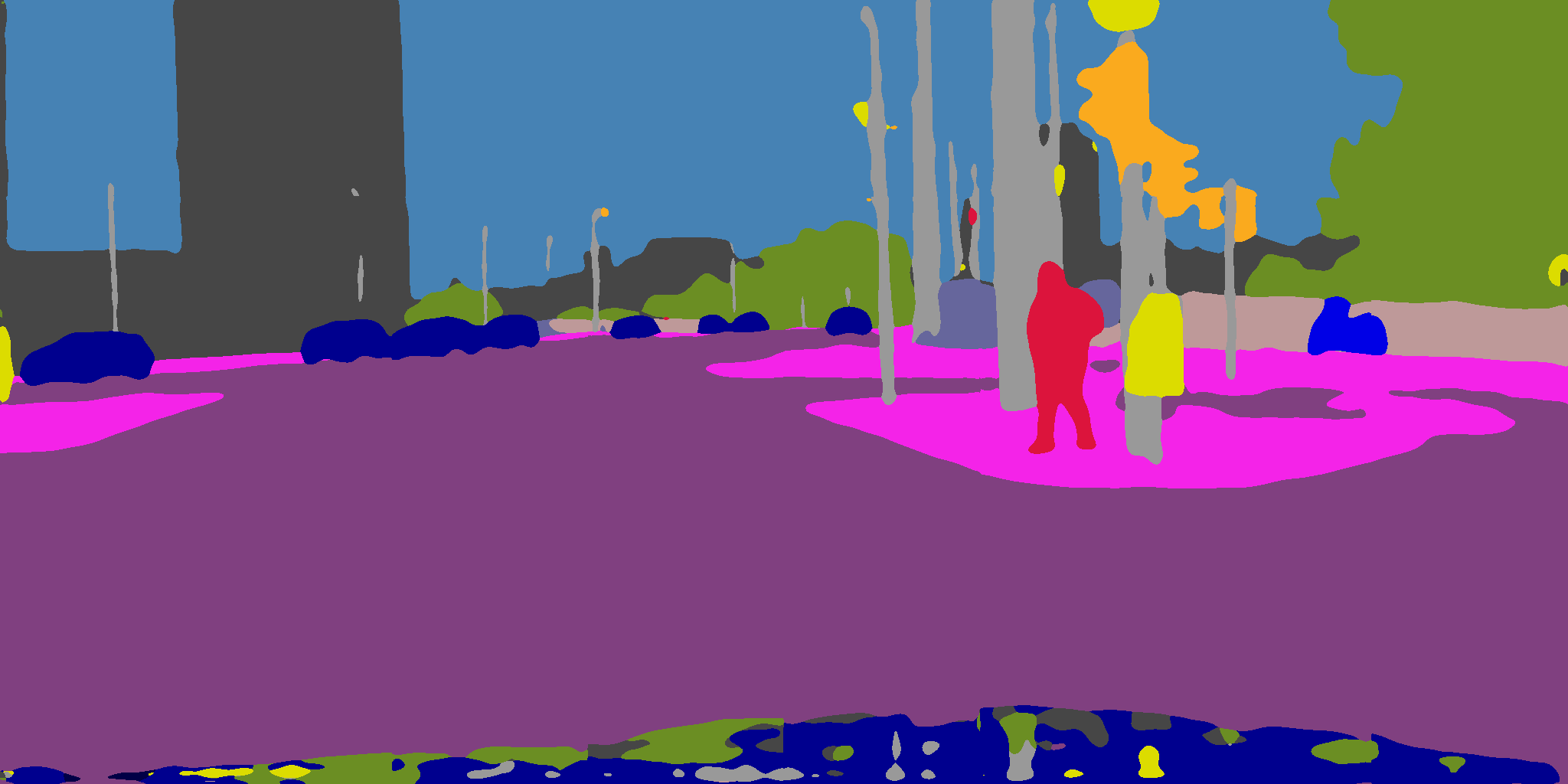} & 
        \includegraphics[width=0.25\textwidth]{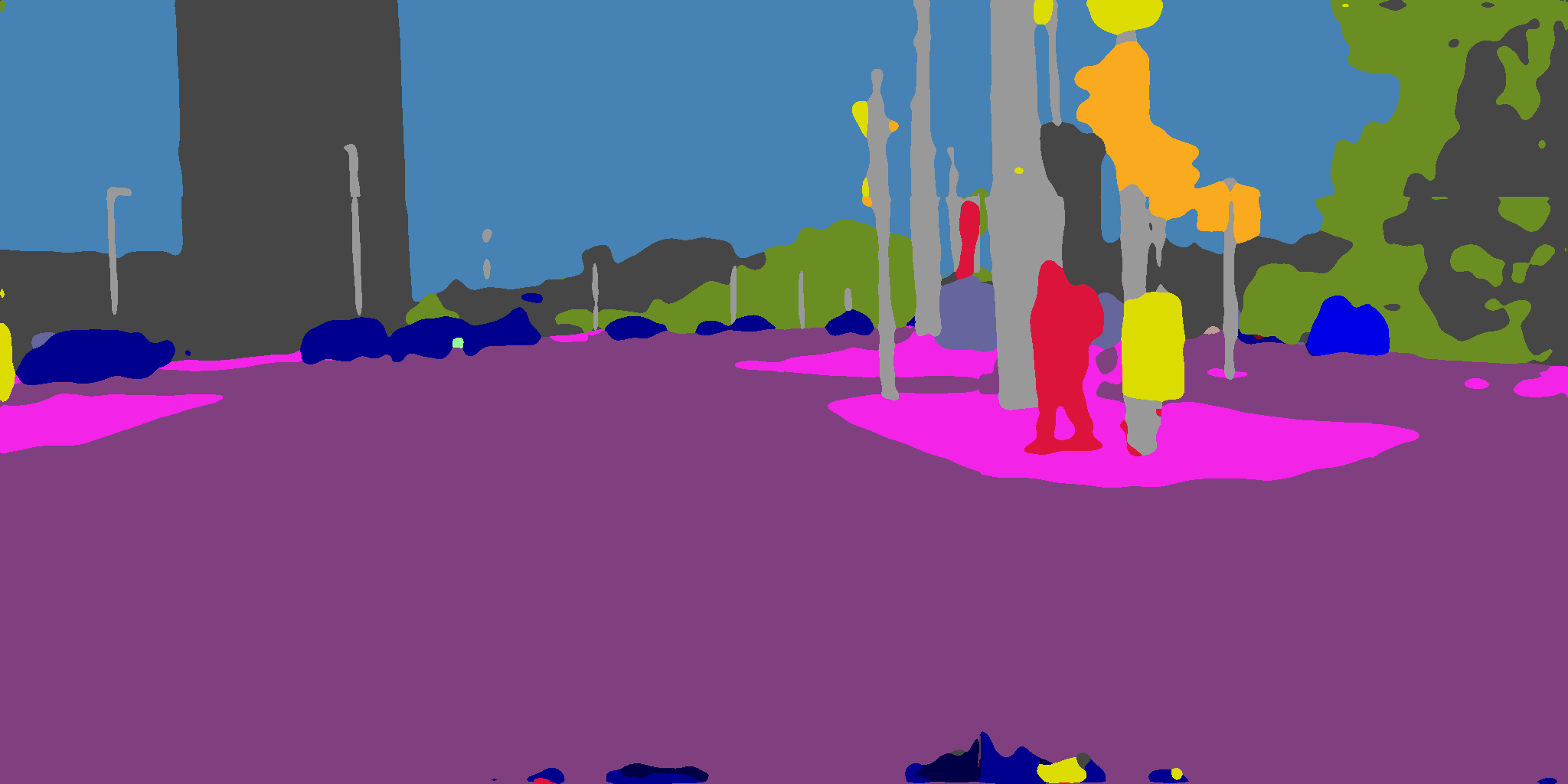} & 
        \includegraphics[width=0.25\textwidth]{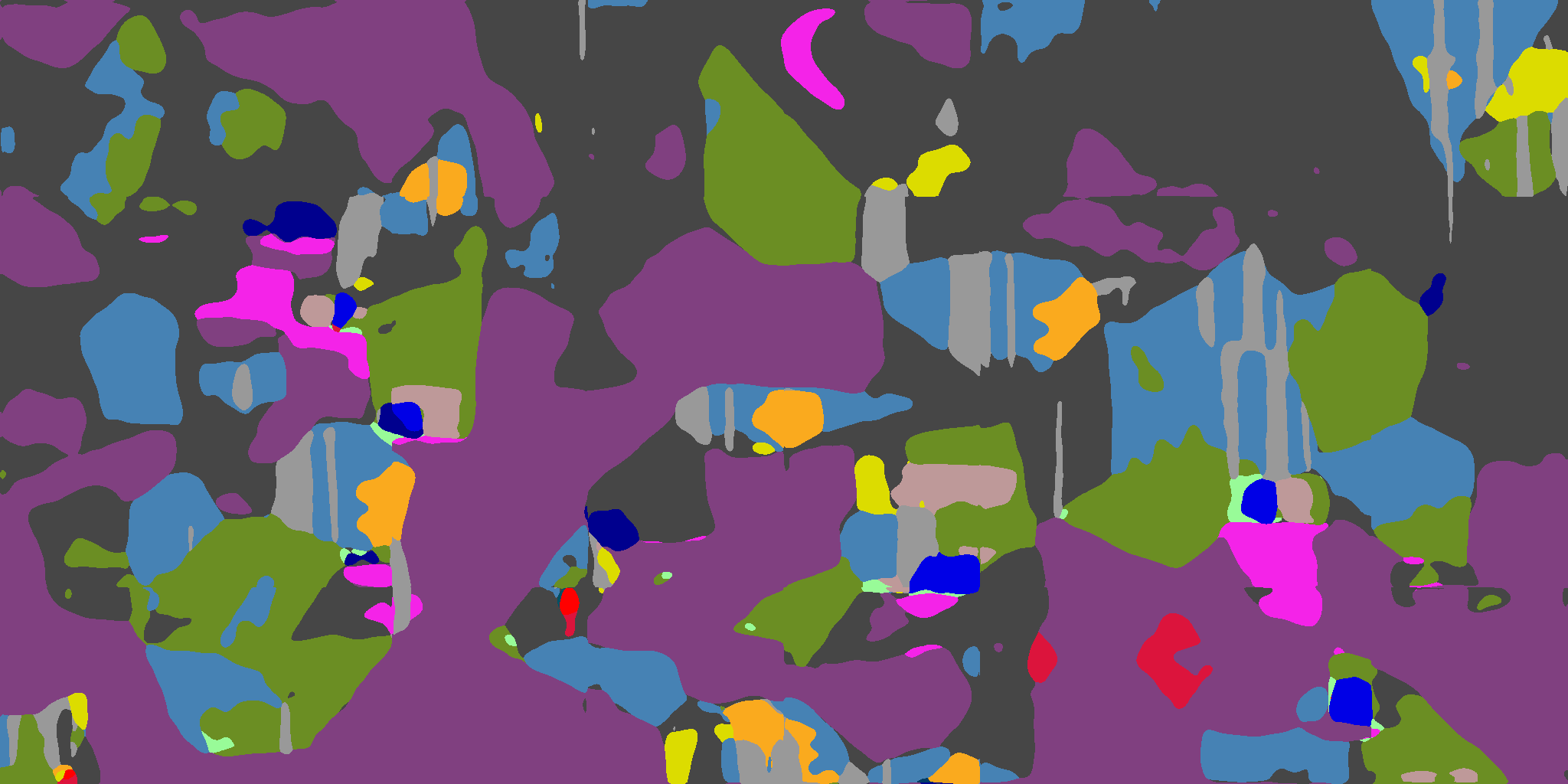} \\
        $S_\mathrm{cd} = 0.648$ & \(Q_O=0.787\) & \(Q_S=0.480\) & \(Q_T=0.366\) \\
        \hline
    \end{tabular}
    \caption{
    A visual example the prediction of three different semantic segmentation DNNs on an original Cityscapes image as well as its EED-transformed and Voronoi-shuffled counterparts.
    \(Q_O\), \(Q_S\) and \(Q_T\) are the Cityscapes mIoU for the respective image type.}
    \label{fig-app:example}
\end{figure*}

\paragraph{A qualitative example}
For a qualitative evaluation, we show prediction results of three different DNNs w.r.t.\ an original image, its EED-transformed as well as its Voronoi-shuffled counterparts in \cref{fig-app:example}. The DeepLabV3Plus with RN101 backbone is a texture biased representative of the class of CNNs. Apparently, its prediction suffers a lot when processing an EED-transformed input image since it clearly relies on texture. On the other hand, when closely inspecting the Voronoi-shuffled image or ground truth mask and comparing it with the predicted segmentation mask for the Voronoi-shuffled input, it can be seen that the network is able to recognize the content of many of the cells quite well, including details like traffic lights and traffic signs.
On the other hand, the Mask2Former and the SETR-Pup show strong capabilities to segment the EED-transformed image, except for some details which is to be expected due to the image diffusion performed by EED. Qualitatively, the predictions of both models on the Voronoi-shuffled input do not deviate too much from the prediction of DeepLabV3Plus. Considering the trade-off between shape and texture cue extraction capabilities, it can be seen that the transformer models have relatively well-balanced capabilities, extracting information well from both cues, while the DeepLabV3Plus has a clear tendency towards texture.

\end{document}